\documentclass[lettersize,journal]{IEEEtran}
\usepackage{amsmath,amsfonts}
\usepackage{algorithmic}
\usepackage{algorithm}
\usepackage{array}
\usepackage[caption=false,font=normalsize,labelfont=sf,textfont=sf]{subfig}
\usepackage{textcomp}
\usepackage{stfloats}
\usepackage{url}
\usepackage{verbatim}

\usepackage{graphicx}
\usepackage{cite}
\usepackage{amssymb}
\usepackage{mathtools}
\usepackage{amsthm}
\usepackage{bbm} %
\usepackage{pifont}
\providecommand{\cmark}{\ding{51}}  %
\providecommand{\xmark}{\ding{55}}  %

\usepackage{booktabs}   %
\usepackage{multirow}   %
\usepackage{makecell}   %
\usepackage{adjustbox}  %
\usepackage{tabularx}

\usepackage[table]{xcolor} 
\definecolor{lightblue}{rgb}{0.85,0.95,1.0}
\definecolor{lightgray}{rgb}{0.95,0.95,0.95}

\theoremstyle{plain}
\newtheorem{theorem}{Theorem}
\newtheorem{proposition}{Proposition}
\newtheorem{lemma}{Lemma}
\newtheorem{corollary}{Corollary}
\theoremstyle{definition}
\newtheorem{definition}{Definition}
\newtheorem{assumption}{Assumption}
\theoremstyle{remark}
\newtheorem{remark}{Remark}

\usepackage{hyperref} 
\usepackage[capitalize,noabbrev]{cleveref} %

\begin{document}
\pagestyle{plain}

\title{A Spectral Framework for Universal Proxy Regularization in Model Merging}

\author{Shangge Liu, Danyue Qian, Qi Fan, Yinghuan Shi, Wenbin Li\textsuperscript{*}, \\
Lei Wang, \textit{Senior Member, IEEE}, Yang Gao and Dacheng Tao, \textit{Fellow, IEEE}
\thanks{
Shangge Liu, Danyue Qian, Qi Fan, Yinghuan Shi, Wenbin Li, and Yang Gao are with the State Key Laboratory of Novel Software Technology, Nanjing University, China (e-mails: lshangge@smail.nju.edu.cn; 231880084@smail.nju.edu.cn; fanqi@nju.edu.cn; syh@nju.edu.cn; liwenbin@nju.edu.cn, gaoy@nju.edu.cn, gaoy@nju.edu.cn)
}
\thanks{
Lei Wang is with the School of Computing and Information Technology, University of Wollongong, Australia (e-mail: leiw@uow.edu.au).
}
\thanks{
Dacheng Tao is with Nanyang Technological University, Singapore (e-mail: dacheng.tao@gmail.com).
}
\thanks{\textsuperscript{*}Corresponding author: Wenbin Li.}
}

\maketitle
\thispagestyle{plain}

\begin{abstract}
Model merging constructs multi-task models without expensive joint training but often suffers from parameter interference. Some solutions mitigate this by constraining finetuning using data from the other tasks to be merged, violating the core independence and privacy assumption of model merging.  Recent work offers preliminary empirical evidence that a universal, task-agnostic proxy distribution can substitute for such data, but leaves open when and why it suffices.
To this end, we propose a \textit{Spectral Framework for Universal Proxy Regularization}, which characterizes the relationship between a proxy distribution and the distribution of other tasks to be merged through spectral coverage ($\alpha$) and leakage ($\beta$). 
Based on the framework, we derive a unified parameter interference bound: smaller $(\alpha, \beta)$ provably yields lower parameter interference through better spectral alignment. 
To exploit such a proxy, we augment the training objective with a proxy regularizer. Evaluating this regularizer during fine-tuning requires a concrete mechanism. We study three instantiations, including (J) an exact evaluation via Jacobian-vector products, (F) a frozen curvature approximation using the Fisher information matrix, and (O) an output-drift formulation that constrains the model's functional deviation on the proxy directly. 
We prove that all three coincide under the NTK linearization, where existing analyses of such regularizers stop, and go further to characterize standard non-linear fine-tuning, bounding how far each instantiation departs from its linearized reference and obtain a three-way hierarchy, zero for (J), proxy-independent for (F), and proxy-quality-scaling for (O).
Empirical validation on the commonly used vision-language model CLIP confirms these theoretical predictions. And instantiated with publicly accessible proxies on both vision and language benchmarks, our framework improves existing pre-merging baselines and approaches the oracle performance that requires access to other tasks' data.
\end{abstract}


\section{Introduction}
\IEEEPARstart{M}{odel} merging~\cite{Yang2024,Adamerging2024, tsv2025,orthoreg2026} has emerged as a powerful paradigm to construct multi-task models by integrating independently finetuned checkpoints without sharing their training data, bypassing the need for expensive joint training. Among various techniques, \textit{Task Arithmetic}~\cite{ta2023} offers a simple yet effective approach by manipulating \textit{task vectors} $\tau := \theta_{\mathrm{ft}} - \theta_{0}$, the parameter shift from the pre-trained model $\theta_{0}$ to the finetuned one $\theta_{\mathrm{ft}}$. However, the success of merging is often hindered by parameter interference, where weight updates for one task may degrade the performance of others.

To mitigate the parameter interference, some solutions~\cite{tta2023,linearatt2025,taujp2025} focus on constraining the fine-tuning process to cultivate ``merge-friendly" task vectors. These methods belong to the category of pre-merging methods, which are distinct from during-merging methods~\cite{Adamerging2024,tsv2025} attempting to resolve conflicts at the final aggregation stage. The ideal objective of these pre-merging methods is to achieve \textit{weight disentanglement}~\cite{tta2023}, a state where each task vector modifies the model's outputs only on its own input region, without affecting the outputs on the regions of other tasks.

\begin{figure}
  \centering
   \includegraphics[width=\linewidth]{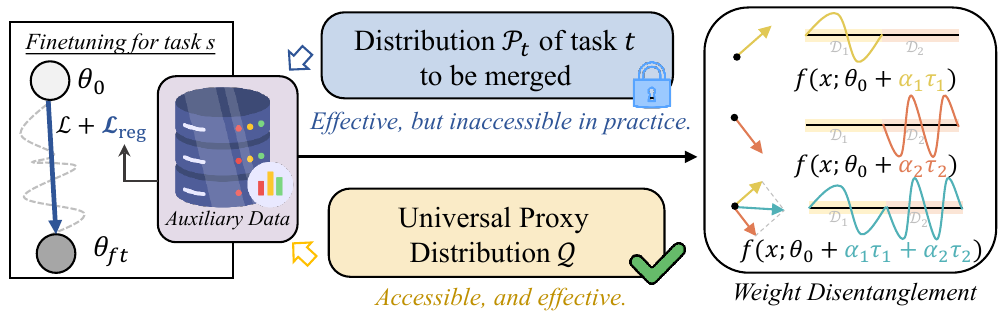}
   \caption{
   Universal proxy distribution regularization. Pre-merging methods typically mitigate interference with data or statistics from the other tasks to be merged ($\mathcal{P}_t$), which are often inaccessible. We instead regularize on an accessible proxy distribution ($\mathcal{Q}$). A suitable proxy achieves weight disentanglement (right), with task vectors ($\tau_1,\tau_2$) operating on their respective domains without cross-task parameter interference.
   }
   \label{fig:motivation}
\end{figure}

To realize this, an intuitive strategy is to impose constraints using auxiliary data. Recent pre-merging methods explore this idea in various forms. For instance, $\tau_{Jp}$~\cite{taujp2025} minimizes the projection of the current task's gradients onto the data representations of other tasks. TAK~\cite{tak2026} penalizes the task vector through a frozen curvature matrix estimated via Kronecker-factored approximations on peer-task data. And OSRM~\cite{osrm2025} constrains each expert's LoRA subspace using examples drawn from the other tasks. Despite their differences in mechanism, these methods share a common limitation: they all necessitate explicit access to peer-task data, that is, the data of the other tasks to be merged. It fundamentally contradicts the core assumption of model merging, that expert models are trained independently without knowledge of other tasks. This reliance on peer-task data undermines the scalability and privacy benefits that render model merging attractive.

A natural question is whether peer-task data can be replaced by a universal, task-agnostic proxy distribution $\mathcal{Q}$ (Figure~\ref{fig:motivation}). 
Recent work offers preliminary empirical support for this possibility~\cite{tak2026,delta2026,ri2026}. The appendix of TAK~\cite{tak2026} shows that KFAC factors estimated once on ImageNet retain most of its gains. These findings, however, are empirical and leave open when and why such a proxy suffices. In this paper, we study this substitution systematically. To this end, we augment fine-tuning with a proxy regularizer $R_\mathcal{Q}$, which measures how strongly $\tau$ perturbs the model's outputs on proxy samples. Three fundamental challenges then arise:
(1) What properties constitute a ``good" universal proxy that effectively suppresses the parameter interference on unknown targets?
(2) How should $R_{\mathcal{Q}}(\tau)$ be evaluated during fine-tuning in practice, and how do different instantiations of the same regularizer compare?
(3) Task arithmetic and its regularization-based methods are typically analyzed under the NTK linearization regime~\cite{tta2023}, while practical fine-tuning operates in the non-linear regime. Do these regularizers remain effective under standard nonlinear fine-tuning, and how far do their solutions deviate from the linearized reference?

To address these challenges, we propose a \textit{Spectral Framework for Universal Proxy Regularization}. It characterizes the relationship between a proxy distribution $\mathcal{Q}$ and the distribution $\mathcal{P}_t$ of other tasks through an $(\alpha, \beta)$-relative spectral bound: $A_{\mathcal{P}_t} \preceq \alpha_t A_{\mathcal{Q}} + \beta_t I$, where $A$ denotes the Jacobian covariance operator. The spectral parameter $\alpha$ (spectral coverage) measures how well the proxy's gradient space spans that of the target task, while $\beta$ (spectral leakage) quantifies the gradient energy in directions not covered by the proxy.
Based on this framework, for \textit{Challenge 1}, we derive a unified interference bound (Theorem~\ref{theorem:unified_bound}), proving that smaller $(\alpha, \beta)$ yields lower interference through better spectral alignment.

For \textit{Challenge~2}, we study three natural instantiations of our framework's proxy regularizer $R_{\mathcal{Q}}$: (J) the Jacobian penalty evaluated exactly via Jacobian-vector products, as in $\tau_{Jp}$~\cite{taujp2025}; (F) the Fisher Information Matrix (FIM)-based penalty using a frozen curvature estimate, as in TAK~\cite{tak2026}; and (O) an output-drift penalty that constrains the model's functional deviation on the proxy directly. We show that the first two exactly recover existing regularizers under our framework, while the third has not been systematically studied as a regularizer in the model-merging literature. 
We prove their equivalence under NTK linearization (Proposition~\ref{prop:linearized_equivalence}), the regime where task arithmetic is known to achieve strong weight disentanglement~\cite{tta2023}. 

For \textit{Challenge~3}, our framework further characterizes the three instantiations beyond the linearization regime. Since practical fine-tuning is non-linear, the equivalence in Proposition~\ref{prop:linearized_equivalence} no longer holds and the three instantiations begin to diverge. Characterizing how far each of them departs from its linearized reference is essential for understanding their practical behavior. To this end, we develop a regularizer-specific gap analysis (Theorems~\ref{theorem:gap_fisher} and \ref{theorem:gap_output}), revealing that the three instantiations exhibit different gap behaviors. Taking (J) as the reference, we show that (F) has a proxy-independent gap governed by the frozen-curvature estimation error, whereas (O) has a gap that scales with the proxy's spectral quality $(\alpha,\beta)$. 
Combined with a per-step cost analysis, we then characterize the practical trade-offs among the three instantiations.

Empirically, we validate the theoretical predictions achieved via the proposed spectral framework on standard vision benchmarks with CLIP ViT-B/32. The estimated spectral parameters across proxies that we construct in various ways correlate strongly with merged accuracy, confirming the prediction of Theorem~\ref{theorem:unified_bound} that smaller $(\alpha, \beta)$ provably suppress parameter interference and thus yield higher merged accuracy.
The three predicted gap behaviors of Theorems~\ref{theorem:gap_fisher} and \ref{theorem:gap_output} are each matched in practice: (J)'s gap is proxy-insensitive, (F)'s gap stays independent across proxies, and (O)'s gap grows as the proxy degrades. 
Paired with ImageNet as proxy,  all three instantiations substantially improve over baselines across ViT-B/32, ViT-B/16, and ViT-L/14, approaching the performance of  oracle $\tau_{Jp}$~\cite{taujp2025} that requires peer-task data.
A task-agnostic WikiText-103 proxy also improves merged performance over unregularized baselines.
Taken together, these results substantiate a key practical implication of our spectral framework. Both theoretically and empirically, effective interference suppression does not require access to peer-task data, and commonly used general-purpose datasets such as ImageNet and WikiText-103 can serve as universal proxies.

Our contributions can be summarized as follows.

\begin{itemize}
\item We develop a spectral framework that characterizes a proxy distribution through an $(\alpha, \beta)$-relative bound, and derive a unified interference bound (Theorem~\ref{theorem:unified_bound}) proving that smaller $(\alpha, \beta)$ yield lower interference.
\item We study three instantiations of the proxy regularizer, Jacobian (J), FIM-based (F), and output-drift (O), and prove their equivalence under NTK linearization.
\item Beyond the linearization regime, we characterize the regularizer-specific gaps between each instantiation's non-linear and linearized solutions (Theorems~\ref{theorem:gap_fisher} and~\ref{theorem:gap_output}). 
\item We empirically confirm both the interference bound and the predicted gap hierarchy. Furthermore, with publicly accessible proxies, our framework improves over existing pre-merging methods and approaches oracle methods.
\end{itemize}
\section{Related Work}
Model merging~\cite{Yang2024,li2026} has emerged as a powerful paradigm for constructing multi-task models by aggregating independently fine-tuned checkpoints, avoiding the cost of joint training. In particular, Task Arithmetic~\cite{ta2023} achieves this idea by manipulating \emph{task vectors}, the parameter shifts between fine-tuned experts and the pre-trained backbone, which can be summed to compose multiple capabilities in a single model. However, simple summation often leads to parameter interference.
Existing solutions generally fall into two categories. \emph{During-merging} methods~\cite{Adamerging2024,tsv2025,Tiesmerging2023,dare2024,knots2025,ri2026} resolve conflicts at the final aggregation stage, \textit{e.g.}, by pruning conflicting signs and magnitudes~\cite{Tiesmerging2023}, learning per-layer aggregation coefficients~\cite{Adamerging2024}, projecting onto shared singular subspaces of the task vectors~\cite{tsv2025}, or adapting each fine-tuned expert on unlabeled auxiliary data~\cite{ri2026}.
\emph{Pre-merging} methods~\cite{llora2024,tta2023,linearatt2025,orthoreg2026,taujp2025,tak2026,osrm2025,mergopt2026,liu2026muon} instead constrain the fine-tuning process to cultivate ``merge-friendly'' weights, \textit{e.g.}, by restricting the update to the tangent space of the pre-trained model~\cite{tta2023}, fine-tuning only the attention modules~\cite{linearatt2025}, or enforcing an internal orthogonal structure on the weight updates~\cite{orthoreg2026}. The ideal objective shared by these methods is to achieve \emph{weight disentanglement}~\cite{tta2023}, a state in which the functional subspace modified by one task vector remains decoupled from those modified by the others.
Appendix~XIII compares our framework with each of these methods in detail (Tables~XVI and~XVII).

To realize weight disentanglement, an intuitive strategy is to impose constraints during fine-tuning using data from the other tasks to be merged. Some pre-merging methods explore this idea in various forms, and their differences lie primarily in how they leverage such data to constrain fine-tuning.
$\tau_{Jp}$~\cite{taujp2025} evaluates the constraint exactly through per-step Jacobian–vector products (JVPs) on peer-task samples, incurring the full cost of repeated Jacobian evaluation. TAK~\cite{tak2026} instead replaces the per-step JVP with a frozen Kronecker-factored curvature matrix pre-computed from peer-task activation and gradient statistics. OSRM~\cite{osrm2025} operates on low-rank adapters, initializing each expert's LoRA subspace to be orthogonal to peer-task feature representations before fine-tuning begins, so that the constraint is enforced structurally rather than as an explicit penalty.
Despite these differences, all three derive their constraint from the data of the tasks to be merged. 
The appendix of TAK~\cite{tak2026} reports that KFAC factors estimated once on ImageNet retain most of its gains, an empirical indication that a task-agnostic distribution can serve the same purpose. Similarly, DELTA~\cite{delta2026} distills a linearized teacher into a non-linear student with a curvature regularizer estimated on a task-agnostic reference dataset. When and why such a proxy suffices, however, remains open.

This raises the questions of what constitutes a good proxy for suppressing parameter interference, and how such a proxy should be leveraged to regularize fine-tuning. To this end, we introduce a spectral framework that characterizes the relationship between the proxy and the tasks to be merged, thereby establishing when proxy regularization can control interference on unseen tasks.
Within our framework, $\tau_{Jp}$~\cite{taujp2025} and TAK~\cite{tak2026} reappear as two instantiations of the same proxy objective on a universal distribution, and we further introduce a third that penalizes the model's output drift on the proxy directly. And we characterize the interference-suppression behavior of the three instantiations both theoretically and empirically.

\section{Preliminaries and Problem Formulation}\label{sec:preliminaries}

\subsection{Notation and Basic Definitions}
We consider a neural network $f: \mathcal{X} \times \Theta \to \mathcal{Y}$, where $\mathcal{X} \subseteq \mathbb{R}^d$, $\mathcal{Y} \subseteq \mathbb{R}^c$, and $\Theta \subseteq \mathbb{R}^p$. Let $\theta_0 \in \Theta$ denote the pre-trained weights. For a given task $t$, we denote the fine-tuned weights by $\theta^*_t \in \Theta$. For any input $x \in \mathcal{X}$, we define the Jacobian matrix of the model with respect to its
parameters as $J(x; \theta_0) = \nabla_\theta f(x; \theta_0) \in \mathbb{R}^{p \times c}$. \footnote{Generally, for a scalar output or a specific component of a vector output, we have $J(x) \in \mathbb{R}^p$.}

\begin{figure*}[t]
  \centering
   \includegraphics[width=0.9\linewidth]{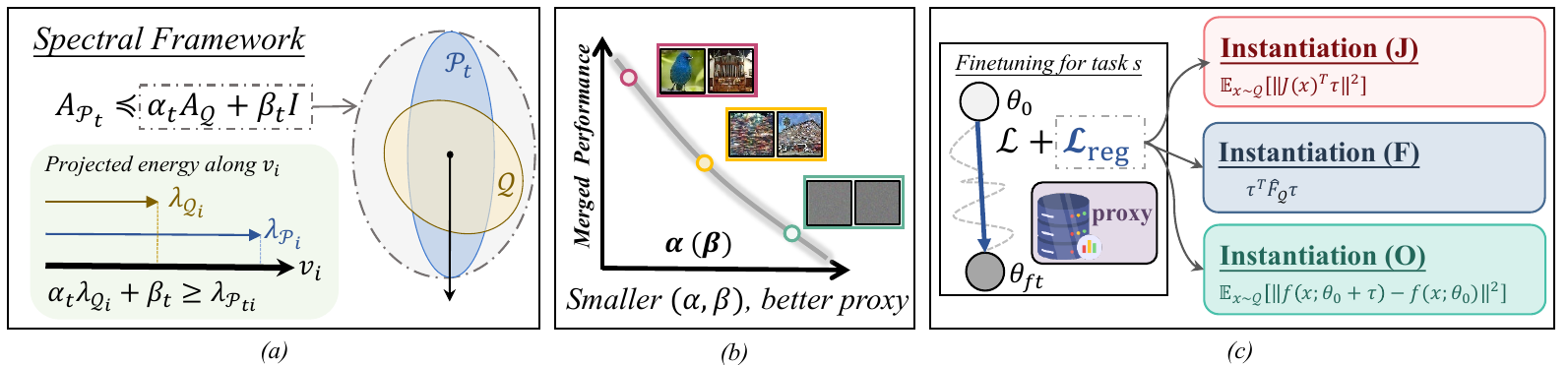}
   \caption{Overview of the \textit{Spectral Framework for Universal Proxy Distribution Regularization}. (a) Spectral Framework. We characterize the relationship between the universal proxy distribution $\mathcal{Q}$ (yellow) and the unseen target task distribution $\mathcal{P}_t$ (blue) using the $(\alpha, \beta)$-relative spectral bound. As illustrated by the projected energy along eigenvector $v_i$, the condition $\alpha_t \lambda_{\mathcal{Q}_i} + \beta_t \geq \lambda_{\mathcal{P}_{ti}}$ ensures that the proxy's gradient subspace adequately covers that of the target task. (b) Theoretical Prediction of Interference. A schematic illustration of our unified interference bound (Theorem~\ref{theorem:unified_bound}). We theoretically predict that smaller spectral parameters $(\alpha, \beta)$ lead to tighter interference bounds and thus higher merged performance.
(c) Three Instantiations of the Proxy Regularizer. We study three realizations of the spectral objective $R_{\mathcal{Q}}(\tau)=\tau^\top A_{\mathcal{Q}}\tau$, Jacobian (J), frozen Fisher (F), and output-drift (O), which coincide under NTK linearization (Proposition~\ref{prop:linearized_equivalence}) and diverge beyond it in a regularizer-specific manner (Theorems~\ref{theorem:gap_fisher} and \ref{theorem:gap_output}).}
   \label{fig:framework}
   \vspace{-10pt}
\end{figure*}

Model merging aims to construct a multi-task model by integrating parameters from independently fine-tuned checkpoints. Task Arithmetic~\cite{ta2023} achieves this by manipulating \textit{task vectors}, defined as the parameter shift from the pre-trained model to the fine-tuned one: $\tau_t := \theta^*_t - \theta_0$. To merge a set of tasks $\mathcal{T}$, the task vectors are typically aggregated and added to the pre-trained base,
\begin{equation}\small \label{eq:merge}
    \theta_\text{merged} = \theta_0 + \sum_{t \in \mathcal{T}} \gamma_t \tau_t,
\end{equation}
where $\gamma_t$ is the scaling coefficient. However, this summation often causes parameter interference, where parameter updates for one task degrade the performance of other tasks.

\subsection{Problem Formulation}
Following recent works~\cite{tta2023}, we formulate the parameter interference under the NTK linearization assumption.

\begin{assumption}[NTK linearization]\label{assumption:NTK linearization}
For a small task vector $\tau$, the model satisfies,
\begin{equation}\small
    f(x; \theta_0 + \tau) \approx f(x; \theta_0) + J(x; \theta_0)^{\top}\tau.
\end{equation}
\end{assumption}\label{assumption:ntk}

The ideal objective of task arithmetic is to achieve \textit{weight disentanglement} as follows.

\begin{definition}[Weight disentanglement]\label{def:weight disentanglement}
For a collection of tasks $\mathcal{T}$ with corresponding task vectors $\{\tau_t\}_{t \in \mathcal{T}}$, we say the task vectors achieve weight disentanglement if,
\begin{equation}\small
    \begin{aligned}
    f\left(x; \theta_0 + \sum_{t \in \mathcal{T}} \gamma_t \tau_t\right) &= \sum_{t \in \mathcal{T}} f(x; \theta_0+ \gamma_t \tau_t) \mathbbm{1}(x \in \mathcal{D}_t) \\ 
    &+ f(x; \theta_0) \mathbbm{1}\left(x \notin \bigcup_{t \in \mathcal{T}} \mathcal{D}_t\right).
\end{aligned}
\end{equation}

\end{definition}

Specifically, for $x\in\mathcal{D}_t$, the merged model behaves as if only $\tau_t$ were applied, while outside all task domains it reduces to the pre-trained model. Thus, on task (t)'s domain, the influence of all other task vectors $\tau_{s\neq t}$ vanishes.

\section{Our Theoretical Framework}\label{sec:framework}

\subsection{Parameter Interference}
We first introduce an operator to quantify the spectral properties of the model's gradients over a given distribution.

\begin{definition} [Jacobian covariance operator]\label{definition:jacobian_cov}
For data distribution $\mathcal{P}$, we define its Jacobian covariance operator as,
\begin{equation}\small
    A_{\mathcal{P}} := \mathbb{E}_{x \sim \mathcal{P}}[J(x)J(x)^{\top}] \in \mathbb{R}^{p \times p}.
\end{equation}
\end{definition}

Intuitively, $A_{\mathcal{P}}$ characterizes the \emph{gradient subspace of $\mathcal{P}$}: the directions in parameter space along which $x \sim \mathcal{P}$ excites the model output through $J(x; \theta_0)$, weighted by their energy under $\mathcal{P}$.
Based on this operator, we can reformulate the condition for weight disentanglement (Definition~\ref{def:weight disentanglement}) as follows.

\begin{lemma}\label{lemma:disentanglement2tauj}
Under Assumption~1, weight disentanglement holds if and only if, for any pair of distinct tasks $s, t \in \mathcal{T}$,
\begin{equation}\small
    J(x_t; \theta_0)^{\top}\tau_s = 0, \quad \forall s \neq t, \, x_t \sim \mathcal{P}_t,
\end{equation}
\end{lemma}

\begin{proof}
    See Appendix~X-A.%
\end{proof}
Lemma~\ref{lemma:disentanglement2tauj} turns the functional notion of weight disentanglement into an explicit condition on cross-task interference.
To quantify the deviation from this ideal orthogonality condition, we define the following interference metric.

\begin{definition}[Distribution-based interference]\label{definition:distribution_interference}
The interference of task vector $\tau$ on data distribution $\mathcal{P}$ is,
\begin{equation}\small
    \mathcal{I}(\tau | \mathcal{P}) := \mathbb{E}_{x \sim \mathcal{P}}\left[\|J(x)^{\top}\tau\|^2\right]. \label{eq:interference}
\end{equation}

This metric quantifies how much $\tau$ perturbs the model's outputs on samples from $\mathcal{P}$. Ideally, for weight disentanglement, we desire $\mathcal{I}(\tau_s | \mathcal{P}_t) \approx 0$ when $s \neq t$.
\end{definition}

\begin{proposition}\label{proposition:quadratic}
    The interference admits a quadratic form
\begin{equation}\small
    \mathcal{I}(\tau | \mathcal{P}) = \tau^{\top}A_{\mathcal{P}}\tau.
\end{equation}
\end{proposition}

\begin{proof}
    See Appendix~X-B.%
\end{proof}

\subsection{Decoupling Regularization via Proxy Distributions}
Recent work~\cite{taujp2025} proposed $\tau_{Jp}$ regularization. When fine-tuning task $s$, their loss objective is
\begin{equation}\small \label{eq:taujp_objective}
\mathcal{L}_{s} = \mathcal{L}_{\text{task}} + \lambda\sum_{t\in\mathcal{T}_{\text{orth}}}\|(\theta_s-\theta_0)^\top\nabla_\theta f(x_t; \theta_0)\|^2,
\end{equation}
where $\mathcal{T}_{\text{orth}}$ denotes other tasks to be merged. Since $\|(\theta_s - \theta_0)^\top \nabla_\theta f(x_t; \theta_0)\|^2$ is exactly the per-sample integrand $\|J(x_t)^\top \tau_s\|^2$ in the interference definition of Eq.~(\ref{eq:interference}), Eq.~(\ref{eq:taujp_objective}) is equivalent in our framework to minimizing the total interference $\sum_{t\in\mathcal{T}_{\text{orth}}} \mathcal{I}(\tau_s|\hat{\mathcal{P}}_t)$.\footnote{$\hat{\mathcal{P}}$ denotes the empirical distribution.}
$\tau_{Jp}$ suffers from three limitations: (1) it necessitates explicit access to the data $\mathcal{D}_t$ of other tasks, contradicting the premise of model merging; (2) the computational cost scales quadratically ($O(N^2)$) with the number of tasks ($N$); and (3) once the task collection changes, all task vectors must be retrained to maintain disentanglement.

To address the limitations, we propose decoupling the regularization from specific tasks. Instead of minimizing interference on the distribution $\mathcal{P}_t$ of other task $t$, we minimize it on a universal, task-agnostic proxy distribution $\mathcal{Q}$.
\textbf{This substitution serves as the starting point of our spectral framework}, which we develop from the proxy regularizer defined below.

\begin{definition}[Proxy regularizer]
    Given a proxy distribution $\mathcal{Q}$ over $\mathcal{X}$, we define,
\begin{equation}\small\label{eq:r_q}
    R_{\mathcal{Q}}(\tau) := \tau^{\top}A_{\mathcal{Q}}\tau = \mathbb{E}_{x \sim \mathcal{Q}}\left[\|J(x)^{\top}\tau\|^2\right].
\end{equation}
\end{definition} 

The key question is: Under what conditions does minimizing $R_{\mathcal{Q}}(\tau)$ guarantee control over the true interference $\mathcal{I}(\tau | \mathcal{P}_t)$ for unseen tasks $t$?

\subsection{Spectral Characterization and Unified Interference Bound}\label{subsec:unified_bound}
To characterize the relationship between proxy distribution ${\mathcal{Q}}$ and target distribution ${\mathcal{P}_t}$, we analyze two complementary perspectives: spectral coverage (quantifying how well ${\mathcal{Q}}$ spans the gradient subspace of ${\mathcal{P}_t}$) and perturbation tolerance (allowing controlled deviations between the two distributions). Detailed analyses of these two perspectives are provided in Appendix~IX. Building upon these insights, we now present a unified spectral framework.

\begin{definition}[$(\alpha, \beta)$-Relative spectral bound] \label{definition:relative_spectral_bound}
For two positive semi-definite matrices $A_{\mathcal{P}_t}$ and $A_{\mathcal{Q}}$, we say $A_{\mathcal{P}_t}$ admits an $(\alpha_t, \beta_t)$-relative spectral bound with respect to $A_{\mathcal{Q}}$ if there exist constants $\alpha_t, \beta_t \geq 0$ such that
\begin{equation}\small
    A_{\mathcal{P}_t} \preceq \alpha_t A_{\mathcal{Q}} + \beta_t I.
\end{equation}
\end{definition}

Here, $\alpha_t$ is the \textit{coverage coefficient}, quantifying how well $\mathcal{Q}$ spans the gradient subspace of task $t$. Complementarily, $\beta_t$ is the \textit{leakage coefficient}, measuring the gradient energy in blind directions not covered by $\mathcal{Q}$. Figure~\ref{fig:framework}(a) provides an intuitive visualization of the relationship.

To derive explicit bounds on interference, we analyze the finetuning process under the NTK regime with squared loss, which admits closed-form analytical solutions~\cite{Han2022}. Specifically, we design the following regularized fine-tuning objective for task $s$, which incorporates the proxy regularizer $R_Q(\tau)$ defined in Eq.~\eqref{eq:r_q},
\begin{equation}\label{formula:objective}\small
    \min_{\tau} \frac{1}{2} \mathbb{E} \left[ (y - f(x; \theta_0) - J(x)^\top \tau)^2 \right] + \frac{\lambda}{2} R_Q(\tau) + \frac{\mu}{2} \|\tau\|^2,
\end{equation}
where $\lambda$ controls the regularization strength, and $\frac{\mu}{2} \|\tau\|^2$ is essential for bounding $\tau$ in directions not covered by $\mathcal{Q}$. 
And this term ensures the well-posedness of Eq.~\eqref{formula:objective}, \textit{i.e.}, the existence of a unique minimizer, even when $\text{rank}(A_{\mathcal{Q}}) < p$~\cite{ridge1970}.

\begin{lemma}\label{lemma:tau_solution}
Under Assumption~1 (NTK linearization), the optimal task vector $\tau^*_s$ minimizing Eq.~\eqref{formula:objective} is given by,
\begin{equation}\small
\tau_s^* = (A_{\mathcal{P}s} + \lambda A_{\mathcal{Q}} + \mu I)^{-1}b_s,
\end{equation}
where $b_s := \mathbb{E}_{(x,y) \sim \mathcal{D}_s}[J(x)(y - f(x;\theta_0))]$.
\end{lemma}
\begin{proof}
See Appendix~X-E.%
\end{proof}

By combining the closed-form solution from Lemma~\ref{lemma:tau_solution} with Definition~\ref{definition:relative_spectral_bound}, we derive a unified bound on the interference.

\begin{theorem}[Unified interference bound]\label{theorem:unified_bound}
Let $\tau_s^*$ be the regularized solution for task $s$ as in Lemma~\ref{lemma:tau_solution}. Suppose task $t \neq s$ satisfies the $(\alpha_t, \beta_t)$-relative spectral bound. Then the interference of $\tau_s^*$ on task $t$ is bounded by,
\begin{equation}\small
    \mathcal{I}(\tau_s^* | \mathcal{P}_t) \leq \left(\frac{\alpha_t}{4\lambda\mu} + \frac{\beta_t}{\mu^2}\right)\|b_s\|^2.
\end{equation}
\end{theorem}

\begin{proof}[Proof sketch]
The proof proceeds in three main steps.

\textbf{Step 1.} By the $(\alpha_t, \beta_t)$-relative spectral bound, we decompose the interference as
\begin{equation}\small
    \mathcal{I}(\tau_s^* | \mathcal{P}_t) = (\tau_s^*)^{\top}A_{\mathcal{P}_t}\tau_s^* \leq \alpha_t (\tau_s^*)^{\top}A_{\mathcal{Q}}\tau_s^* + \beta_t \|\tau_s^*\|^2.
\end{equation}

\textbf{Step 2.} Left-multiplying the optimality condition $(A_{\mathcal{P}s} + \lambda A_{\mathcal{Q}} + \mu I)\tau_s^* = b_s$ of Lemma~\ref{lemma:tau_solution} by $(\tau_s^*)^{\top}$, and using $A_{\mathcal{P}s} \succeq 0$ and the Cauchy--Schwarz inequality, gives $\lambda(\tau_s^*)^{\top}A_{\mathcal{Q}}\tau_s^* + \mu\|\tau_s^*\|^2 \leq \|b_s\|\,\|\tau_s^*\|$. The first term is therefore at most $g(\|\tau_s^*\|)$ with $g(z) = \|b_s\|z - \mu z^2$, which is maximized at $g\big(\frac{\|b_s\|}{2\mu}\big) = \frac{\|b_s\|^2}{4\mu}$. This yields $(\tau_s^*)^{\top}A_{\mathcal{Q}}\tau_s^* \leq \frac{\|b_s\|^2}{4\lambda\mu}$.

\textbf{Step 3.} Since $\lambda(\tau_s^*)^{\top}A_{\mathcal{Q}}\tau_s^* \geq 0$, the same inequality also gives $\|\tau_s^*\|^2 \leq \frac{\|b_s\|^2}{\mu^2}$.

Combining yields $\mathcal{I}(\tau_s^* | \mathcal{P}_t) \leq \left(\frac{\alpha_t}{4\lambda\mu} + \frac{\beta_t}{\mu^2}\right)\|b_s\|^2$. See Appendix~X-F for details.
\end{proof}

\begin{remark}[Regularized coverage coefficient]\label{rem:coverage}
Theorem~\ref{theorem:unified_bound} holds for every pair in the admissible set $\mathcal{F}_t=\{(\alpha,\beta)\ge 0:\ A_{P_t}\preceq\alpha A_Q+\beta I\}$, so the pair is not unique. We select pairs on its lower boundary through a ridge level $\rho>0$:
\begin{equation}\label{eq:coverage}\small
\alpha_t(\rho):=\max_{v\neq 0}\frac{v^\top A_{P_t}v}{v^\top(A_Q+\rho I)v}, \qquad \beta_t(\rho):=\rho\,\alpha_t(\rho),
\end{equation}
\textit{i.e.}, $\alpha_t(\rho)$ is the smallest $\alpha$ with $A_{P_t}\preceq\alpha(A_Q+\rho I)$. Varying $\rho$ traces the entire boundary (Lemma~3). We call $\alpha_t(\rho)$ the \emph{regularized coverage coefficient} and the function $\alpha_t:(0,\infty)\to[0,\infty)$ the \emph{coverage curve}. 
\end{remark}

\begin{figure*}[t]
  \centering
   \includegraphics[width=0.8\linewidth]{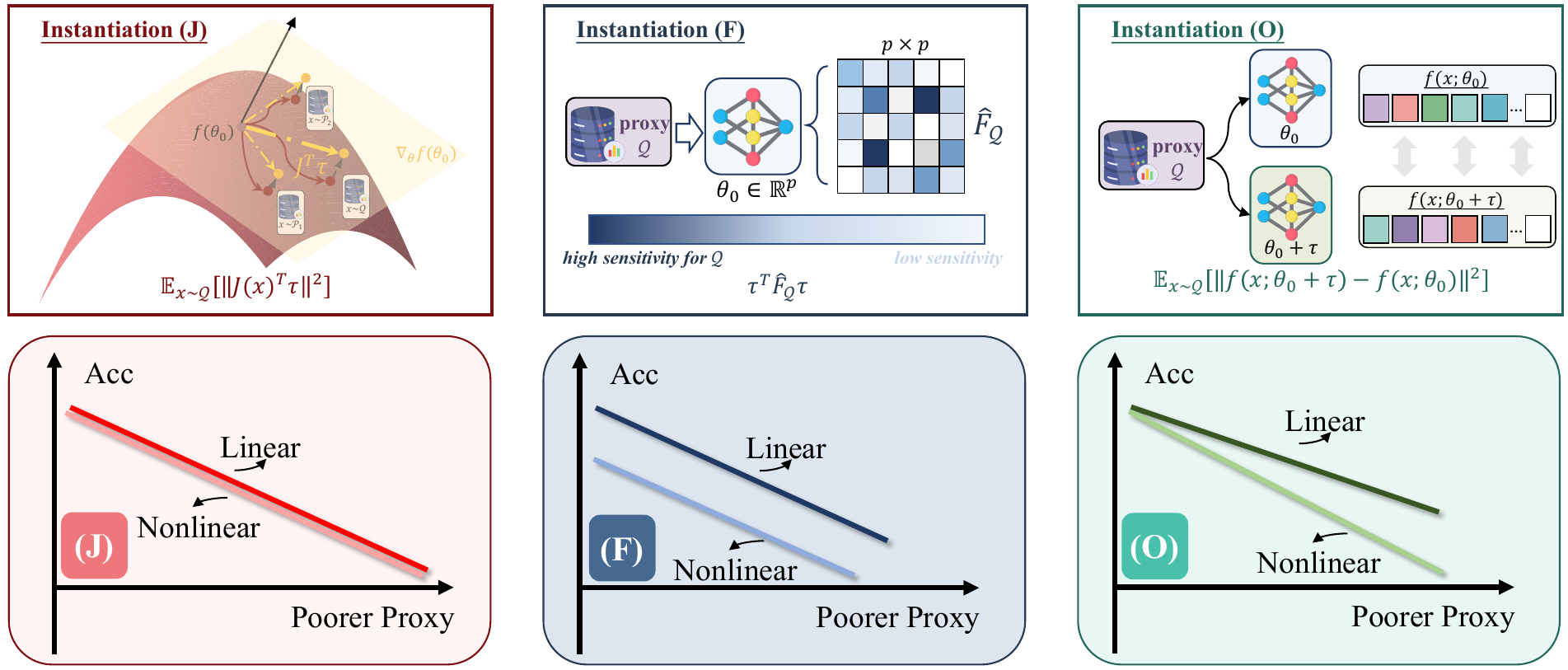}
   \caption{Three instantiations of the proxy regularizer. 
   \textbf{Top}: mechanism illustration for (J) exact evaluation via Jacobian--vector products, (F) a frozen Fisher/KFAC curvature snapshot $\widehat F_{\mathcal Q}$, and (O) direct measurement of the model's output drift on the proxy.
   \textbf{Bottom:} predicted merged accuracy as the proxy quality degrades (larger $\alpha, \beta$), illustrating the three-way regularizer-specific gap hierarchy: zero for (J), constant for (F), and proxy-quality-scaling for (O).
   }
   \label{fig:three_instantiations}
\end{figure*}

\begin{corollary}[Regularized form of Theorem~\ref{theorem:unified_bound}]\label{cor:coverage}
Under the conditions of Theorem~\ref{theorem:unified_bound}, for every $\rho>0$,
\begin{equation}\label{eq:coverage-bound}\small
I(\tau_s^*\,|\,P_t)\le \alpha_t(\rho)\Bigl(\frac{1}{4\lambda\mu}+\frac{\rho}{\mu^2}\Bigr)\|b_s\|^2 .
\end{equation}
\textup{(i)} The infimum of the right-hand side over $\rho>0$ equals the tightest bound that Theorem~\ref{theorem:unified_bound} yields over all admissible pairs $(\alpha_t,\beta_t)$.
\textup{(ii)} If two proxies satisfy $\alpha_t^{Q_1}(\rho)\le\alpha_t^{Q_2}(\rho)$ for all $\rho>0$,
then the bound for $Q_1$ is no larger than that for $Q_2$ for every $(\lambda,\mu)$.
\end{corollary}
\begin{proof}
See Appendix~X-F.
\end{proof}

In essence, a good proxy distribution needs low $\alpha_t$ (broad gradient coverage) and low $\beta_t$ (small leakage), both of which directly reduce the interference bound, given that $\|b_s\|$ is independent of the proxy distribution. 
This bound answers \textit{Challenge 1}, identifying $(\alpha_t,\beta_t)$ as the sole properties a proxy needs to control interference, and thus showing the peer-task access required by prior methods could be avoided provided that suitable $\alpha_t$ and $\beta_t$ exist.
We empirically validate this framework in Section~\ref{subsec:exp_spectral}, where the estimated regularized coverage coefficient $\widehat\alpha_t(\rho)$ (Remark~\ref{rem:coverage}) correlates strongly with task arithmetic performance.

\section{Regularization via Proxy Distributions}
\label{sec:proxy_regularization}

In Section~\ref{sec:framework}, we established that minimizing the proxy regularizer $R_{\mathcal Q}(\tau)$ provably bounds the interference on the tasks to be merged, provided the proxy $\mathcal Q$ satisfies the $(\alpha,\beta)$-relative spectral bound (Definition~\ref{definition:relative_spectral_bound}). We now turn to \textit{Challenge~2}: how should $R_{\mathcal Q}(\tau)$ be realized during fine-tuning?

Existing work offers two distinct mechanisms: Jacobian-based regularization (\textit{e.g.}, $\tau_{Jp}$~\cite{taujp2025}) and Fisher Information Matrix (FIM)-based regularization (\textit{e.g.}, TAK~\cite{tak2026}). We further introduce a third, which constrains the model's output drift on the proxy distribution directly. As Figure~\ref{fig:three_instantiations} shows, this section presents the three regularizations within a unified framework (Section~\ref{sec:three_realizations}), shows that they yield the same solution under NTK linearization (Section~\ref{sec:linearized_equivalence}), and quantifies, for each regularization, the gap between its linearized and standard (non-linear) solutions (Section~\ref{sec:method_specific_gap}).

\subsection{Three Instantiations of the Proxy Regularizer}
\label{sec:three_realizations}

Throughout, fine-tuning on task $s$ minimizes a regularized objective of
the form
\begin{equation}\small
\mathcal L_{\mathrm{task}}(\theta_0+\tau)
+\frac{\lambda}{2}\,\Phi(\tau)
+\frac{\mu}{2}\|\tau\|^2 ,
\label{eq:generic_regularized_objective}
\end{equation}
where $\Phi$ is a proxy-based penalty instantiating $R_{\mathcal Q}(\tau)=\tau^\top A_{\mathcal Q}\tau$. The three formulations differ only in \emph{the mechanism they use to evaluate} it.

\textit{\textbf{(J) Jacobian regularization.}}
The direct instantiation takes 
\begin{equation}\small
    \Phi^J(\tau)=R_{\mathcal Q}(\tau)= \mathbb{E}_{x \sim \mathcal{Q}}\left[\|J(x)^{\top}\tau\|^2\right],
\end{equation} %
evaluated exactly at every optimization step via Jacobian-vector products. This recovers the mechanism of $\tau_{Jp}$ regularization~\cite{taujp2025} under a universal proxy, whereas $\tau_{Jp}$ was originally formulated on peer-task data.
Its cost is the repeated Jacobian–vector product (JVP) evaluation, which increases peak memory and training time relative to plain fine-tuning~\cite{tta2023}.

\textit{\textbf{(F) Fisher Information Matrix (FIM)-based regularization.}}
The second instantiation freezes the curvature.  Recall that for a model with output likelihood $p(y|x;\theta)$, the FIM on the proxy $\mathcal{Q}$ is
\begin{equation}\small
F_{\mathcal Q}(\theta):=\mathbb E_{x\sim\mathcal Q}\,\mathbb E_{y\sim p(y\mid x;\theta)}\!\left[\nabla_\theta\log p(y\mid x;\theta)\,\nabla_\theta\log p(y\mid x;\theta)^\top\right].
\label{eq:fim_definition}
\end{equation}
Under a Gaussian output likelihood (squared loss), after absorbing an overall scalar into $\lambda$,  $F_{\mathcal Q}(\theta_0)=\mathbb E_{x\sim\mathcal Q}\!\left[J(x)J(x)^\top\right]=A_{\mathcal Q}$ (see Appendix~X-G), so the FIM recovers exactly the curvature matrix of $R_{\mathcal Q}$.  
The practical implementation of this 
realization replaces $A_{\mathcal Q}$ by an estimate $\widehat F_{\mathcal Q}$ of $F_{\mathcal Q}(\theta_0)$, computed \emph{once} at $\theta_0$ and kept fixed, yielding the penalty
\begin{equation}\small
\Phi^F(\tau)=\tau^\top\widehat F_{\mathcal Q}\tau .
\label{eq:fisher_realization}
\end{equation}
TAK~\cite{tak2026} is a representative instance, which approximates $\widehat F_{\mathcal Q}$ in Kronecker-factored (KFAC) form.

\textit{\textbf{(O) Output-drift regularization.}}
We introduce a third instantiation that constrains the model's functional deviation on the proxy directly,
\begin{equation}\small
\Phi^O(\tau)=\mathbb E_{x\sim\mathcal Q} \!\left[\|f(x;\theta_0+\tau)-f(x;\theta_0)\|^2\right].
\label{eq:odr_definition}
\end{equation}
Evaluating Eq.~\eqref{eq:odr_definition} requires only standard forward passes on proxy samples.  Penalizing such output drift is itself a common objective in continual learning and knowledge distillation~\cite{li2017learning,de2021continual}. Unlike (J) and (F), the penalty is not a fixed quadratic form in $\tau$; it is a \emph{trajectory} regularizer that tracks the exact output drift of the model.

We treat (J), (F), and (O) as three instantiations of the same spectral objective $R_{\mathcal Q}(\tau)$ and compare them under a common setup throughout this paper. 
Our goal is to characterize how each instantiation behaves within the proposed spectral framework, rather than to recommend any single choice as the best.

\begin{table*}[t]\small
\centering
\caption{Three instantiations of the proxy regularizer $R_{\mathcal Q}(\tau)=\tau^\top A_{\mathcal Q}\tau$. Each evaluates the proxy functional deviation differently and consequently carries a different per-step cost, extra memory, and source of the regularizer-specific gap $E_\Phi$.}
\label{tab:three_realizations}
\small
\setlength{\tabcolsep}{4pt}
\begin{tabular*}{\textwidth}{@{\extracolsep{\fill}}cllll@{}}
\toprule
Penalty $\Phi$ & Mechanism & Additional per-step cost & Extra memory & Source of $E_\Phi$ \\
\midrule
\textup{(J)} & exact quadratic form via JVPs & JVP + VJP & $\theta_0$ snapshot & $0$ \\
\textup{(F)} & frozen $\tau^\top\widehat F_{\mathcal Q}\tau$ & parameter-space matmuls & Fisher Matrix & curvature error $\overline\varepsilon_{\mathrm F}$ \\
\textup{(O)} & exact output drift & forward & $\theta_0$ snapshot & scaling as $q_t(\alpha_t,\beta_t)$ \\
\bottomrule
\end{tabular*}
\end{table*}

\subsection{Equivalence under Linearization}
\label{sec:linearized_equivalence}

The three instantiations are not merely heuristically related. We now show that under NTK linearization they induce \emph{identical} solutions.

Let $f_{\mathrm{lin}}(x;\theta_0+\tau):=f(x;\theta_0)+J(x;\theta_0)^\top\tau$ denote the linearized model according to the NTK linearization in Assumption~1. Because instantiation \textup{(J)} evaluates $R_{\mathcal Q}(\tau)$ exactly by construction, we take it as the reference and analyze how \textup{(O)} and \textup{(F)} relate to \textup{(J)} under linearization.

\begin{proposition}[Equivalence under linearization]
\label{prop:linearized_equivalence}
Fix $\lambda,\mu>0$, and consider the linearized model $f_{\mathrm{lin}}$.
\begin{itemize}
\item[(i)] \textbf{Instantiation (O).}
On the linearized model, $\Phi^O(\tau)=R_{\mathcal Q}(\tau)$ for all $\tau$; hence $\tau_{\mathrm{lin}}^{O}=\tau_{\mathrm{lin}}^{J}$.
\item[(ii)] \textbf{Instantiation (F).}
If the frozen FIM is exact, $\widehat F_{\mathcal Q}=A_{\mathcal Q}$, then $\Phi^F(\tau)=R_{\mathcal Q}(\tau)$ for all $\tau$; hence $\tau_{\mathrm{lin}}^{F}=\tau_{\mathrm{lin}}^{J}$.
\end{itemize}
\end{proposition}

\begin{proof}[Proof sketch]
On the linearized model the functional deviation is linear in $\tau$, $f_{\mathrm{lin}}(x;\theta_0+\tau)-f(x;\theta_0)=J(x)^\top\tau$, so $\Phi^O$ reduces to $R_{\mathcal Q}$ exactly. For \textup{(F)}, the Gaussian-likelihood FIM at $\theta_0$ equals $A_{\mathcal Q}$, so the exact case gives $\Phi^F=R_{\mathcal Q}$.  The three objectives are then identical and $\mu$-strongly convex due to the ridge term, and thus share a unique minimizer. See Appendix~X-G for the detailed proof.
\end{proof}

Hence Theorem~\ref{theorem:unified_bound}'s interference guarantee holds for all three instantiations in the linearized regime. They diverge once the model leaves this regime, which we quantify next.

\subsection{Beyond Linearization: Regularizer-Specific Gaps}
\label{sec:method_specific_gap}

Proposition~\ref{prop:linearized_equivalence} establishes the equivalence of the three instantiations only under NTK linearization. This is also the regime in which existing regularization-based pre-merging methods are analyzed and deployed~\cite{taujp2025,tak2026}. 
Yet standard non-linear fine-tuning remains the common practice~\cite{ta2023,linearatt2025}, and linearized training is considerably more computationally expensive~\cite{tta2023}. 
This raises \textit{Challenge~3}: do these regularizers remain effective under standard non-linear fine-tuning, and how far do their solutions then depart from the linearized reference? To our knowledge, neither question has been analyzed theoretically.
We now characterize, for each instantiation separately, how far its standard (non-linear) solution departs from its own linearized one.
To this end, we associate with each instantiation a pair of solutions: for mode $m\in\{\mathrm{lin},\mathrm{std}\}$ and $\Phi\in\{\mathrm J,\mathrm F,\mathrm O\}$,

\begin{equation}\small
\tau_m^{\Phi}:=\arg\min_\tau\;\mathcal L_{\mathrm{task}}^{m}(\tau)+\frac{\lambda}{2}\,\Phi_m(\tau)+\frac{\mu}{2}\|\tau\|^2 ,
\label{eq:method_specific_solution}
\end{equation}
where $\mathcal L_{\mathrm{task}}^{\mathrm{lin}}$ and $\mathcal L_{\mathrm{task}}^{\mathrm{std}}$ denote the task loss evaluated on $f_{\mathrm{lin}}$ and $f$, respectively.  The mode-specific penalties follow from Section~\ref{sec:three_realizations}: instantiations \textup{(J)} and \textup{(F)} are frozen quadratic forms in $\tau$ and hence identical in both modes,
\begin{equation}\small
\Phi_{\mathrm{lin}}^{\mathrm J}=\Phi_{\mathrm{std}}^{\mathrm J}=R_{\mathcal Q},
\qquad
\Phi_{\mathrm{lin}}^{\mathrm F}=\Phi_{\mathrm{std}}^{\mathrm F}=\tau^\top\widehat F_{\mathcal Q}\tau ,
\label{eq:frozen_penalty_pairs}
\end{equation}
whereas the trajectory penalty \textup{(O)} degenerates to $R_{\mathcal Q}$ only under linearization (Proposition~\ref{prop:linearized_equivalence}(i)),
\begin{equation}\small
\Phi_{\mathrm{lin}}^{\mathrm O}=R_{\mathcal Q},\qquad\Phi_{\mathrm{std}}^{\mathrm O}(\tau)=\mathbb E_{x\sim\mathcal Q}\!\left[\|f(x;\theta_0+\tau)-f(x;\theta_0)\|^2\right].
\label{eq:odr_penalty_pair}
\end{equation}
In particular, $\tau_{\mathrm{lin}}^{\mathrm O}=\tau_{\mathrm{lin}}^{\mathrm J}$.

For a target task $t\neq s$, we quantify how switching modes changes the interference of each instantiation,
\begin{equation}\small
\delta_\Phi(t):=\mathcal I(\tau_{\mathrm{std}}^\Phi\mid\mathcal P_t)
-\mathcal I(\tau_{\mathrm{lin}}^\Phi\mid\mathcal P_t).
\label{eq:signed_mode_change}
\end{equation}
Because instantiation \textup{(J)} evaluates $R_{\mathcal Q}$ exactly in both modes (Eq.~\eqref{eq:frozen_penalty_pairs}), $\delta_{\mathrm J}(t)$ receives no contribution from the regularizer. Instead, it reflects only the non-linearity of the \emph{task loss}, which is common to all three instantiations.  Taking \textup{(J)} as the reference, we define the \textbf{regularizer-specific gap}
\begin{equation}\small
E_\Phi(t):=|\delta_\Phi(t)-\delta_{\mathrm J}(t)| ,
\label{eq:excess_gap}
\end{equation}
which isolates the contribution of the \emph{regularization mechanism} itself.
Note that $E_\Phi$ is a difference of mode changes rather than a mode change itself. Since $\mathcal L_{\mathrm{task}}$ is non-linear, $\delta_{\mathrm J}(t)\neq0$ in general, so every instantiation, including \textup{(J)}, exhibits a non-zero standard-versus-linearized change. The hierarchy established below therefore concerns the \emph{deviation of each instantiation from} \textup{(J)}.

The source of the regularizer-specific gap differs between \textup{(F)} and \textup{(O)}.  For \textup{(F)}, it is the frozen curvature error $E_{\mathcal Q}:=\widehat F_{\mathcal Q}-A_{\mathcal Q}$, a fixed matrix perturbation.  For \textup{(O)}, a second-order Taylor expansion of $f$ around $\theta_0$ shows that the standard penalty deviates from its linearized counterpart through a Hessian interaction term,
\begin{equation}\small
\Phi_{\mathrm{std}}^{\mathrm O}(\tau)-\Phi_{\mathrm{lin}}^{\mathrm O}(\tau)=\mathbb E_{x\sim\mathcal Q}\!\left[\left\langle J(x)^\top\tau,\,H_x[\tau,\tau]\right\rangle\right]+O(\|\tau\|^4),
\label{eq:odr_mode_perturbation}
\end{equation}
where $H_x[u,v]:=\bigl(u^\top\nabla_\theta^2 f_k(x;\theta_0)\,v\bigr)_{k=1}^{c}$ collects the second-order directional derivatives of the $c$ outputs. A detailed derivation is given in Appendix~X-H.%

We now bound the regularizer-specific gap of each instantiation in turn. All three share the same local regularity conditions, which we state once.
\begin{assumption}[Local regularity]
\label{assumption:local_regularity}
All solutions $\{\tau_m^\Phi\}$ and the segments joining the pairs compared below lie in a bounded neighborhood of $\theta_0$. The model is three-times continuously differentiable there with bounded Jacobian. The pre-training residuals are bounded, and the standard-mode task loss is $\kappa$-strongly convex with $\kappa\ge0$ on this neighborhood.  Detailed statements are given in Appendix~X-H.
\end{assumption}

\begin{proposition}[\textup{(J)} is mode-exact]
\label{prop:gap_jacobian}
The penalty of instantiation \textup{(J)} is the \emph{same} functional $R_{\mathcal Q}$ in both modes, so its regularizer contributes no mode-dependent term.  Consequently \textup{(J)} is a valid reference in Eq.~\eqref{eq:excess_gap}, and $E_{\mathrm J}(t)=0$ for every task $t$.
\end{proposition}

\begin{theorem}[\textup{(F)}: proxy-independent gap]
\label{theorem:gap_fisher}
Let Assumption~\ref{assumption:local_regularity} hold and suppose that the frozen curvature satisfies
\begin{equation}\small
\widehat F_{\mathcal Q}=A_{\mathcal Q}+E_{\mathcal Q},
\qquad
\|E_{\mathcal Q}\|\le\overline\varepsilon_{\mathrm F}.
\label{eq:fisher_uniform_error}
\end{equation}
Then there exists a finite constant $C_{\mathrm F}\ge0$, carrying no explicit $(\alpha_t,\beta_t)$ factor, such that
\begin{equation}\small
E_{\mathrm F}(t)\le C_{\mathrm F}.
\label{eq:gap_fisher_bound}
\end{equation}
In the exact case $\widehat F_{\mathcal Q}=A_{\mathcal Q}$, instantiations \textup{(F)} and \textup{(J)} coincide as functionals in both modes, and hence $E_{\mathrm F}(t)=0$.
\end{theorem}

\begin{proof}
    See Appendix~X-H.
\end{proof}

\begin{theorem}[\textup{(O)}: proxy-amplified gap]
\label{theorem:gap_output}
Let Assumption~\ref{assumption:local_regularity} hold and suppose that for task $t\neq s$ the target matrix $A_{\mathcal P_t}$ admits an $(\alpha_t,\beta_t)$-relative spectral bound with respect to the proxy $A_{\mathcal Q}$ (Definition~\ref{definition:relative_spectral_bound}).  Define
\begin{equation}\small
q_t(\alpha_t,\beta_t):=\left(\frac{\alpha_t}{4\lambda\mu}+\frac{\beta_t}{\mu^2}\right)^{1/2}.
\label{eq:proxy_spectral_factor}
\end{equation}
Then there exists a finite constant $C_{\mathrm O}\ge0$, independent of $(\alpha_t,\beta_t)$, such that
\begin{equation}\small
E_{\mathrm O}(t)\le C_{\mathrm O}\,q_t(\alpha_t,\beta_t).
\label{eq:gap_output_bound}
\end{equation}
\end{theorem}

\begin{proof}
    See Appendix~X-H.
\end{proof}

\begin{corollary}[Three-way hierarchy]
\label{cor:three_way_hierarchy}
Under Assumption~\ref{assumption:local_regularity},
\begin{equation}\small
E_{\mathrm J}(t)=0,\qquad
E_{\mathrm F}(t)\le C_{\mathrm F},\qquad
E_{\mathrm O}(t)\le C_{\mathrm O}\,q_t(\alpha_t,\beta_t),
\label{eq:three_way_excess_summary}
\end{equation}
with $C_{\mathrm F},C_{\mathrm O}$ independent of $(\alpha_t,\beta_t)$.
\end{corollary}

Corollary~\ref{cor:three_way_hierarchy} thus establishes a three-way hierarchy: exact for \textup{(J)}, proxy-independent for \textup{(F)}, and proxy-amplified for \textup{(O)}. Empirically (Section~\ref{sec:experiments}), this yields three testable predictions, one per instantiation. Instantiation \textup{(J)}'s standard-to-linearized gap reflects only the task-loss non-linearity, and is therefore independent of proxy quality. Instantiation \textup{(F)}'s gap is likewise insensitive to proxy quality, being controlled by the frozen curvature error rather than by $(\alpha_t,\beta_t)$. Instantiation \textup{(O)}'s gap, by contrast, depends on proxy quality: it matches \textup{(J)} on high-quality proxies and widens as the proxy degrades.

\subsection{Computational Efficiency}
\label{sec:method_efficiency}

The three instantiations also differ in per-step computational cost.  Combined with Proposition~\ref{prop:linearized_equivalence} and the regularizer-specific gap of Theorems~\ref{theorem:gap_fisher} and \ref{theorem:gap_output}, this cost axis completes the practical comparison, unified in Table~\ref{tab:three_realizations}.

For \textup{(J)}, each proxy mini-batch requires a forward-mode JVP at $\theta_0$ followed by a reverse-mode VJP sweep to obtain the gradient~\cite{taujp2025}, and peak activation memory scales with these additional passes. A snapshot of $\theta_0$ must also remain in memory throughout fine-tuning.

For \textup{(F)}, the penalty $\tau^\top\widehat F_{\mathcal Q}\tau$ uses a fixed matrix computed once at $\theta_0$, so each step reduces to parameter-space matrix products, with no forward or backward pass through the network. However, storing the full $\widehat F_{\mathcal Q}\in\mathbb R^{p\times p}$ is prohibitive for modern architectures, and even in Kronecker-factored form~\cite{tak2026} the block-diagonal factors incur persistent memory that becomes material for large backbones.

For \textup{(O)}, no Jacobian products or auxiliary curvature matrices are involved. Each step performs a standard forward and backward through the output-drift loss, with the reference output $f(x;\theta_0)$ supplied either by a pre-computed tensor cache or by an additional no-grad forward through a frozen copy of the reference model.

\section{Experiments}\label{sec:experiments}

Given the theoretical foundations established in Sections~\ref{sec:framework} and~\ref{sec:proxy_regularization}, our experimental evaluation is organized around three questions. First, does a proxy's spectral quality $(\alpha,\beta)$ indeed control the merged model's performance, as predicted by Theorem~\ref{theorem:unified_bound}? Second, do the three instantiations (J), (F), and (O) exhibit the regularizer-specific gaps predicted by Theorems~\ref{theorem:gap_fisher} and \ref{theorem:gap_output}? 
Third, when combined with a practical universal proxy, what accuracy do the three instantiations of our unified framework achieve on standard model-merging benchmarks, and how does this compare to existing pre-merging baselines?

\textbf{Basic experimental setup.} Across these experiments, we utilize a consistent setup involving eight standard vision benchmarks (Cars~\cite{DBLP:conf/iccvw/Krause0DF13}, DTD~\cite{Cimpoi_2014_CVPR}, EuroSAT~\cite{Helber2017EuroSATAN}, GTSRB~\cite{Stallkamp2011TheGT}, MNIST~\cite{726791}, RESISC45~\cite{Cheng2017RemoteSI}, SUN397~\cite{5539970}, and SVHN~\cite{37648}) covering diverse domains. We employ three scales of CLIP-based Vision Transformers: ViT-B/32, ViT-B/16, and ViT-L/14. Following prior works~\cite{tta2023,taujp2025,orthoreg2026}, we report two key metrics: Absolute Accuracy of the merged model and Normalized Accuracy, which measures the performance retention relative to single-task fine-tuned specialists. 
All experts are fine-tuned with AdamW and weight decay following prior work. The explicit ridge term $\frac{\mu}{2}\|\tau\|^2$ is examined in Section~\ref{sec:ablation}. Detailed descriptions of datasets, metrics, and implementation details are provided in Appendix~XI-A.%

\begin{figure}[t]
  \centering
   \includegraphics[width=\linewidth]{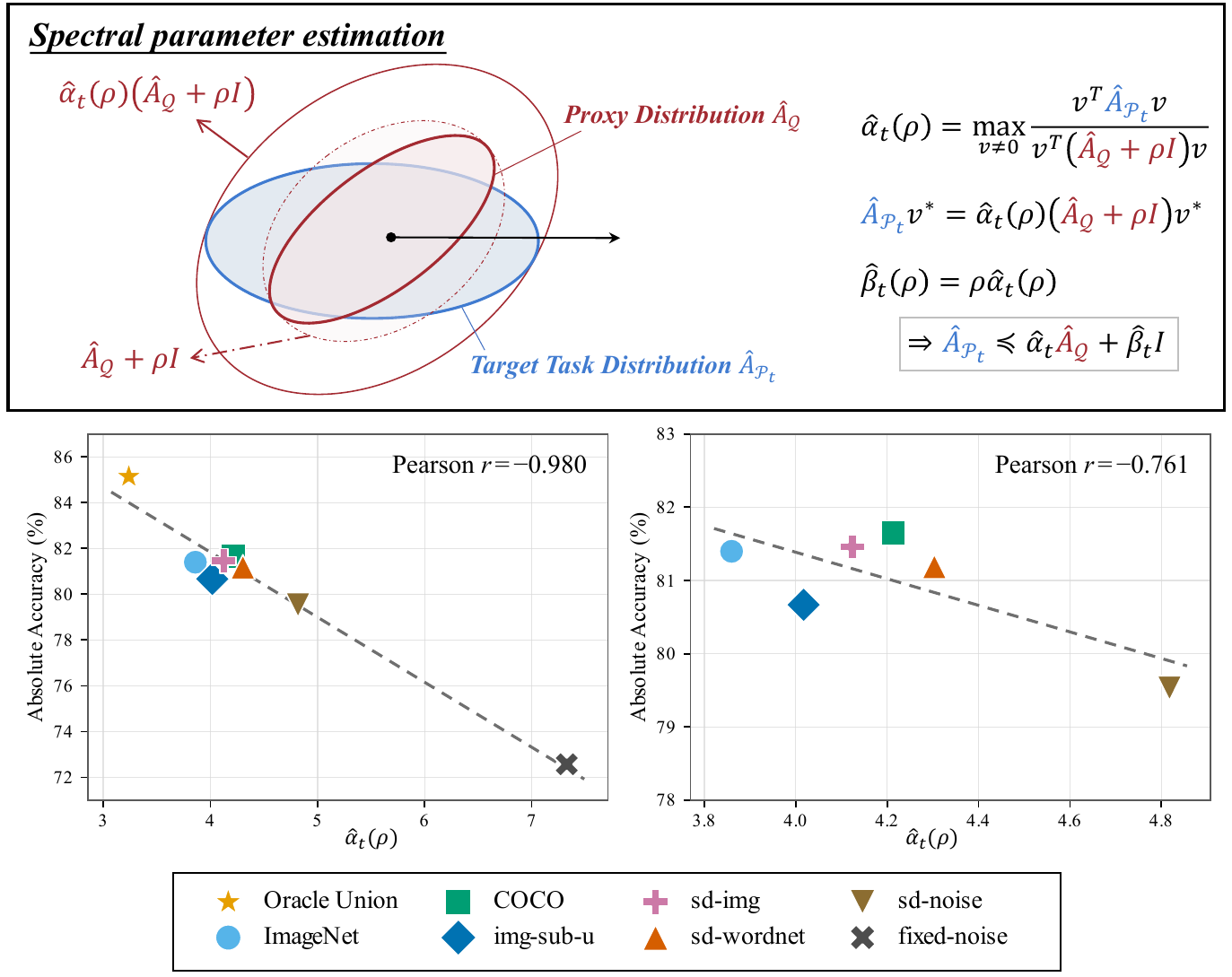}
    \caption{Verification of the spectral interference bound.
    \textbf{Top:} estimating the regularized coverage coefficient $\widehat\alpha_t(\rho)$. Each matrix is drawn as an ellipse, so that $A\preceq B$ means the ellipse of $A$ lies inside that of
    $B$. The proxy covariance $\widehat A_{\mathcal Q}$ is first regularized to $\widehat A_{\mathcal Q}+\rho I$ and then enlarged by the smallest factor $\widehat\alpha_t(\rho)$ that makes it enclose the target covariance $\widehat A_{\mathcal P_t}$. Together with $\widehat\beta_t(\rho)=\rho\,\widehat\alpha_t(\rho)$, this factor gives an admissible pair for Theorem~1, so a smaller $\widehat\alpha_t(\rho)$ means better coverage and a tighter interference bound.
    \textbf{Bottom:} task-addition accuracy of ViT-B/32 regularized with (J) on each proxy, plotted against $\widehat\alpha_t(\rho)$ averaged over the eight target tasks. Dashed lines
    are least-squares fits. Left: all proxies (Pearson $r=-0.980$). Right: the two extreme proxies, Oracle Union and fixed-noise, removed ($r=-0.761$). The negative correlation persists among the remaining six.}
   \label{fig:theory_validation}
\end{figure}

\begin{figure*}[t]
  \centering
   \includegraphics[width=\linewidth]{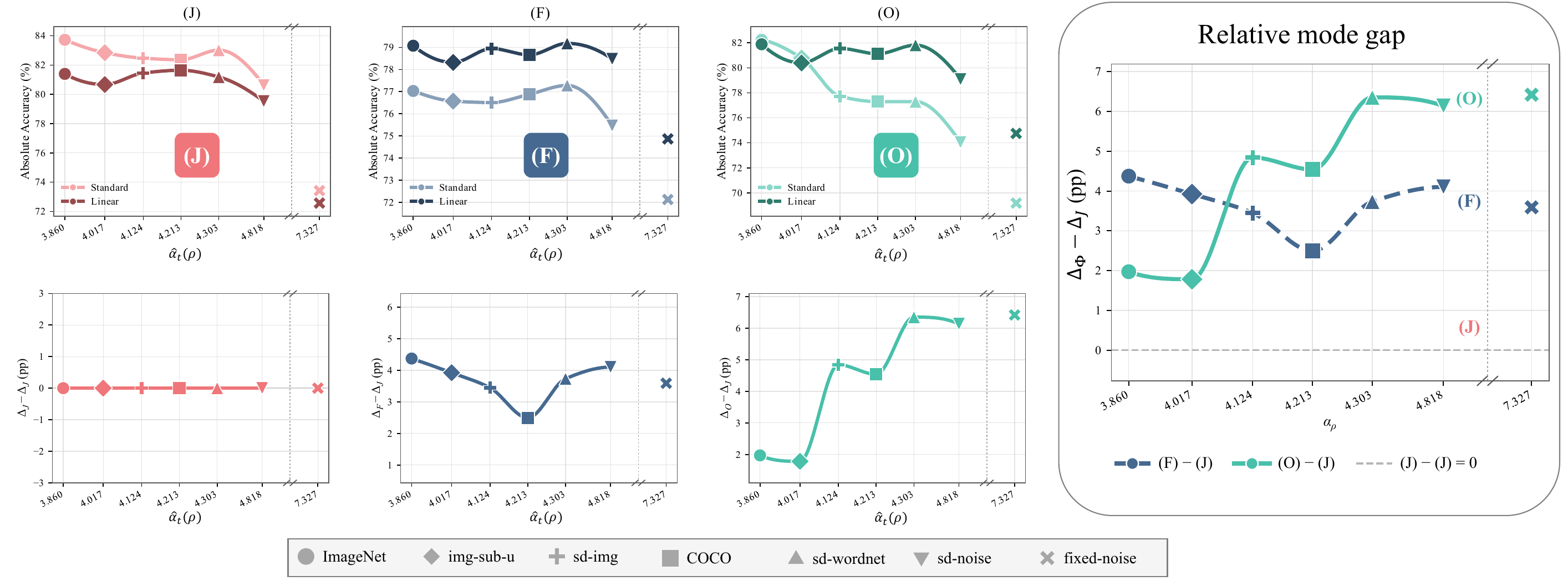}
   \caption{Empirical verification of Theorems~\ref{theorem:gap_fisher} and \ref{theorem:gap_output} on ViT-B/32. Rows show absolute accuracy (top) and the relative mode gap $\Delta_\Phi - \Delta_J$ (bottom) for each instantiation $\Phi \in \{(\mathrm{J}),(\mathrm{F}),(\mathrm{O})\}$ under standard and linearized fine-tuning, across seven proxies ordered by increasing $\widehat\alpha_t(\rho)$. 
   The rightmost panel plots the relative mode gap $\Delta_\Phi - \Delta_J$, the accuracy-level counterpart of the regularizer-specific gap $E_\Phi$ in Corollary 2; by construction (J) collapses to the zero reference line. As predicted, $(\mathrm{F})$ is proxy-independent with a nonzero offset, and $(\mathrm{O})$ grows with $\widehat\alpha_t(\rho)$.
   }
   \label{fig:reg_gap}
\end{figure*}

\subsection{Verification of the Spectral Interference Bound}\label{subsec:exp_spectral}

To validate the interference bound in Theorem~\ref{theorem:unified_bound}, we estimate spectral parameters $(\alpha_t, \beta_t)$ for various proxy distributions $\mathcal{Q}$ and correlate them with merged model performance. Since parameter interference is the primary cause of performance degradation in model merging, our framework predicts that lower $(\alpha, \beta)$ implying lower interference should theoretically yield higher accuracy. Because Theorem~\ref{theorem:unified_bound} concerns the proxy regularizer $R_\mathcal{Q}(\tau)$ itself, we verify Theorem~\ref{theorem:unified_bound} using instantiation (J), which evaluates $R_\mathcal{Q}$ exactly at every optimization step. The other two instantiations are validated separately in Section~\ref{subsec:exp_method_gap}.

\textbf{Proxy distributions.} 
We define the \textit{Oracle Union} proxy, sampling from distributions of other tasks to be merged as a theoretical performance ceiling, despite violating the independence assumption. This is the exact setting used in $\tau_{Jp}$~\cite{taujp2025}.
Meanwhile, we consider seven proxy distributions spanning different characteristics: (i) \textit{ImageNet}~\cite{DBLP:conf/cvpr/DengDSLL009}, a large-scale natural image dataset; (ii) \textit{COCO}, a small random subset sampled from COCO~\cite{coco2014}, covering everyday scenes; 
(iii) \textit{img-sub-u}, a subset of ImageNet with the lowest CLIP similarity to the labels of all eight target tasks;
(iv) \textit{sd-noise}, images generated by Stable Diffusion~\cite{DBLP:conf/cvpr/RombachBLEO22} from random noise prompts; (v) \textit{sd-img}, images generated from ImageNet class labels;
(vi) \textit{sd-wordnet}, images generated from WordNet object synsets, and (vii) \textit{fixed-noise}, fixed random Gaussian noise in pixel space. The details of the proxy distributions are in Appendix~XI-B.%

\textbf{Spectral parameter estimation.} We estimate the spectral parameters $(\alpha_t,\beta_t)$ through the boundary pair $(\alpha_t(\rho),\rho\,\alpha_t(\rho))$ of Remark~\ref{rem:coverage}. By Proposition~\ref{proposition:quadratic}, Definition~\ref{definition:relative_spectral_bound} is equivalent to $I(v\,|\,P_t)\le\alpha_t R_Q(v)+\beta_t\|v\|^2$ for every parameter direction $v$. Fixing the ratio $\beta_t/\alpha_t=\rho$, $\alpha_t(\rho)$ is the smallest $\alpha_t$ for which this holds. We estimate it by replacing $A_{P_t}$ and $A_Q$ with their empirical counterparts $\widehat A_{P_t}$ and $\widehat A_Q$ on finite samples. 
The estimate $\widehat\alpha_t(\rho)$ is the largest generalized eigenvalue of $(\widehat A_{P_t},\widehat A_Q+\rho I)$, which we compute at a fixed ridge level $\rho$, shared by all target tasks and proxies, with Jacobian--vector products and without forming any $p\times p$ matrix.
Since $\widehat\beta_t(\rho)=\rho\,\widehat\alpha_t(\rho)$, both the bound of Theorem~\ref{theorem:unified_bound} and $q_t$ in Theorem~\ref{theorem:gap_output} increase with $\widehat\alpha_t(\rho)$ at fixed $\rho$, so we summarize each proxy by $\widehat\alpha_t(\rho)$. The choice of $\rho$, implementation details, sample sizes, and convergence diagnostics are given in Appendix~XI-C.

\textbf{Results and analysis.} 
Figure~\ref{fig:theory_validation} illustrates the relationship between the estimated spectral parameters and the task-addition accuracy obtained by regularizing linearized fine-tuning with (J) on each proxy distribution. We use linearized fine-tuning because Theorem~\ref{theorem:unified_bound} is derived under NTK linearization. We observe a strong negative Pearson correlation between the task-averaged $\widehat\alpha_t(\rho)$ and merged accuracy ($r=-0.980$ over all eight proxies), and the correlation remains negative ($r=-0.761$) when the two extreme proxies, Oracle Union and fixed-noise, are excluded. This confirms our theoretical framework (Theorem~\ref{theorem:unified_bound}) that larger spectral parameters ($\alpha$ and $\beta$) correspond to looser interference bounds, suggesting greater interference and lower accuracy.

Meanwhile, the results reveal several insights about proxy selection. 
(1) Large-scale natural image datasets serve as effective universal proxies. 
Both \textit{ImageNet} and \textit{COCO}, though entirely task-agnostic, attain merged accuracy close to the \textit{Oracle Union}.
This demonstrates that broad natural image distributions suffice for regularization.
(2) Meaningful gradient structure is essential. The failure of \textit{fixed-noise} confirms that effective proxies must possess meaningful gradient structures. Mere coverage of the input domain is insufficient to activate relevant parameter subspaces. 
(3) High semantic similarity to the targets is not required. \textit{img-sub-u}, which excludes the ImageNet classes most similar to any target task, still attains spectral parameters and merged accuracy comparable to those of ImageNet.
(4) The natural image manifold is effective. \textit{Sd-noise} significantly outperforms \textit{fixed-noise}, indicating that samples on the learned natural image manifold provide valuable regularization even without specific semantic grounding. 
(5) Priors offer incremental gains. Incorporating object-centric priors (\textit{sd-wordnet}) yields better spectral alignment than unconditional generation (\textit{sd-noise}), but the gain is modest compared to the manifold effect. This suggests leveraging domain knowledge when available.

\subsection{Verification of Regularizer-Specific Gaps}\label{subsec:exp_method_gap}
We now empirically test the three-way hierarchy of regularizer-specific linear-to-standard gaps: $E_{\mathrm J}=0$, $E_{\mathrm F}$ is bounded by a proxy-independent constant, and $E_{\mathrm O}$ scales with $q_t(\alpha_t,\beta_t)$.
We examine this hierarchy both through the Standard-vs-Linear accuracy gap $\Delta_\Phi:=\mathrm{Abs}(\tau^\Phi_{\mathrm{lin}})-\mathrm{Abs}(\tau^\Phi_{\mathrm{std}})$ and by measuring $E_\Phi$ directly on the interference $\mathcal I(\tau\mid\mathcal P_t)$. Every $\Delta_\Phi$ contains the common task-loss contribution, and this is precisely what (J) alone measures: because $E_{\mathrm J}=0$, the (J) curve $\Delta_{\mathrm J}$ estimates that common term, so the deviation of $\Delta_{\mathrm F}$ and $\Delta_{\mathrm O}$ from $\Delta_{\mathrm J}$, $D_\Phi:=|\Delta_\Phi-\Delta_{\mathrm J}|$, is the accuracy-level analogue of $E_{\mathrm F}$ and $E_{\mathrm O}$. We evaluate $\Delta_\Phi$ for each instantiation $\Phi\in\{\mathrm J,\mathrm F,\mathrm O\}$ across proxies of varying spectral quality.

\begin{table}[t]\small
    \centering
    \caption{
        Absolute accuracy (\%) of the three instantiations under linearized (Lin) and standard (Std) fine-tuning on ViT-B/32. For each proxy we report the empirical raw mode gap $\Delta_\Phi = \text{Abs}(\tau_{\text{lin}}^\Phi) - \text{Abs}(\tau_{\text{std}}^\Phi)$, and its deviation from (J), $D_\Phi := |\Delta_\Phi - \Delta_J|$. Proxies are ordered by increasing $\widehat\alpha_t(\rho)$.
    }
    \label{tab:method_specific_proxy_results}
    \resizebox{\columnwidth}{!}{%
    \begin{tabular}{llccc}
        \toprule
        \multirow{2}{*}{Proxy}
        & \multirow{2}{*}{Mode}
        & \multicolumn{3}{c}{Absolute Accuracy (\%)} \\
        \cmidrule(lr){3-5}
        & & (J) & (F) & (O) \\
        \midrule
        \multirow{4}{*}{ImageNet}
        & Lin     & 81.40 & 79.07 & 81.88 \\
        & Std     & 83.73 & 77.03 & 82.24 \\
        & $\Delta_\Phi$ & $-2.33$ & $+2.04$ & $-0.36$ \\
        & $D_\Phi$ & $0$ (ref.) & $4.37$ & $1.97$ \\
        \addlinespace
        \multirow{4}{*}{img-sub-u}
        & Lin     & 80.67 & 78.32 & 80.39 \\
        & Std     & 82.85 & 76.57 & 80.78 \\
        & $\Delta_\Phi$ & $-2.18$ & $+1.75$ & $-0.39$ \\
        & $D_\Phi$ & $0$ (ref.) & $3.93$ & $1.79$ \\
        \addlinespace
        \multirow{4}{*}{sd-img}
        & Lin     & 81.46 & 78.94 & 81.55 \\
        & Std     & 82.47 & 76.51 & 77.72 \\
        & $\Delta_\Phi$ & $-1.01$ & $+2.43$ & $+3.83$ \\
        & $D_\Phi$ & $0$ (ref.) & $3.44$ & $4.84$ \\
        \addlinespace
        \multirow{4}{*}{COCO}
        & Lin     & 81.65 & 78.66 & 81.12 \\
        & Std     & 82.36 & 76.88 & 77.30 \\
        & $\Delta_\Phi$ & $-0.71$ & $+1.78$ & $+3.82$ \\
        & $D_\Phi$ & $0$ (ref.) & $2.49$ & $4.53$ \\
        \addlinespace
        \multirow{4}{*}{sd-wordnet}
        & Lin     & 81.19 & 79.18 & 81.82 \\
        & Std     & 83.03 & 77.28 & 77.30 \\
        & $\Delta_\Phi$ & $-1.84$ & $+1.90$ & $+4.52$ \\
        & $D_\Phi$ & $0$ (ref.) & $3.74$ & $6.36$ \\
        \addlinespace
        \multirow{4}{*}{sd-noise}
        & Lin     & 79.53 & 78.47 & 79.09 \\
        & Std     & 80.62 & 75.46 & 74.04 \\
        & $\Delta_\Phi$ & $-1.09$ & $+3.01$ & $+5.05$ \\
        & $D_\Phi$ & $0$ (ref.) & $4.10$ & $6.14$ \\
        \addlinespace
        \multirow{4}{*}{fixed-noise}
        & Lin     & 72.57 & 74.87 & 74.75 \\
        & Std     & 73.42 & 72.13 & 69.18 \\
        & $\Delta_\Phi$ & $-0.85$ & $+2.74$ & $+5.57$ \\
        & $D_\Phi$ & $0$ (ref.) & $3.59$ & $6.42$ \\
        \midrule
        \multicolumn{2}{l}{$|\Delta_\Phi|$ range}
                                 & $[0.71,\,2.33]$
                                 & $[1.75,\,3.01]$
                                 & $[0.36,\,5.57]$ \\
        \multicolumn{2}{l}{$D_\Phi$ range}
                                 & $0$
                                 & $[2.49,\,4.37]$
                                 & $[1.79,\,6.42]$ \\
        \bottomrule
    \end{tabular}%
    }
\end{table}

\textbf{Experimental setup.} We fine-tune the CLIP ViT-B/32 with each of the three instantiations in both standard (non-linear) and linearized modes, across the proxy distributions of Section~\ref{subsec:exp_spectral}. For each configuration, hyperparameters ($\lambda$ and, for merging, $\gamma$) are selected by grid search on a held-out split. Absolute accuracy after task addition is then compared between the two modes, holding all other factors fixed.

\textbf{Results and analysis.}
Table~\ref{tab:method_specific_proxy_results} lists absolute accuracy under both modes for each of the three instantiations, together with the empirical gap $\Delta_\Phi$, and Fig.~\ref{fig:reg_gap} visualizes how each gap behaves as the proxy quality varies, measured by the task-averaged $\widehat\alpha_t(\rho)$ of Section~\ref{subsec:exp_spectral} and abbreviated as $\hat\alpha$ below.
We now examine each instantiation against its own prediction.

\textbf{Instantiation (J)} exhibits a small, proxy-insensitive mode change. Since $E_J = 0$ holds by definition (Eq.~\eqref{eq:excess_gap}), the (J) column of Table~\ref{tab:method_specific_proxy_results} reports the mode change induced by the task-loss non-linearity alone. Across all seven proxies, $|\Delta_J|$ stays within 2.33 points and does not grow with $\hat\alpha$. The (J) mode change thus serves as a stable reference for reading (F) and (O) below.

\begin{table}[t]\small
    \centering
    \caption{Per-step cost of the three regularizer instantiations on ViT-B/32, compared against the unregularized baselines (\emph{none}, \textit{i.e.}, Nonlin.~FT~\cite{ta2023} in standard mode and TTA~\cite{tta2023} in linear mode). \emph{Step} is the wall-clock time per optimizer step. \emph{Steady peak} is peak CUDA-allocated memory during the steady-state loop. \emph{Prepare} is the one-time cost of preparing the $\theta_0$ snapshot (for (J)/(O)) or estimating the KFAC factors (for (F)), whose peak allocation is reported separately.}
    \label{tab:method_specific_efficiency}
    \resizebox{\columnwidth}{!}{%
    \begin{tabular}{llccc c}
        \toprule
        \multirow{2}{*}{Mode} & \multirow{2}{*}{Reg.} &
        \multirow{2}{*}{\makecell{Step\\(ms)}} &
        \multirow{2}{*}{\makecell{Steady peak\\(MiB)}} &
        \multicolumn{2}{c}{Prepare} \\
        \cmidrule(lr){5-6}
        & & & & Time (s) & Peak (MiB) \\
        \midrule
        \multirow{4}{*}{Standard}
            & none & $120.0$ & $2356.2$ & $0.00$  & $451.9$   \\
            & (J)  & $271.3$ & $4140.6$ & $2.99$  & $451.9$   \\
            & (F)  & $237.8$ & $7513.9$ & $18.45$ & $6349.8$  \\
            & (O)  & $139.3$ & $3371.4$ & $0.08$  & $884.7$   \\
        \midrule
        \multirow{4}{*}{Linear}
            & none & $219.1$ & $3900.8$ & $0.00$  & $884.6$   \\
            & (J)  & $237.8$ & $4769.1$ & $0.00$  & $884.6$   \\
            & (F)  & $356.4$ & $8149.1$ & $10.29$ & $6564.1$  \\
            & (O)  & $289.0$ & $5634.7$ & $0.03$  & $1750.2$  \\
        \bottomrule
    \end{tabular}%
    }
\end{table}

\textbf{Instantiation (F)}\footnote{Throughout this paper, (F) denotes a controlled KFAC instantiation evaluated inside our task-arithmetic pipeline, not a reproduction of TAK under its original training recipe~\cite{tak2026}. The two implementations optimize different parameter subsets and follow different optimization schedules, both of which materially affect merged accuracy. Appendix~XI-D provides the details.} matches $E_{\mathrm{F}}\le C_{\mathrm{F}}$. $\Delta_{\mathrm{F}}$ is positive on every proxy and confined to a $1.26$-point interval $[+1.75,\,+3.01]$.
The uniform sign indicates that the frozen KFAC surrogate underperforms its own linearized reference regardless of which proxy generates it, and the narrowness of the band confirms that this excess is insensitive to $\hat\alpha$: moving from the highest-quality proxy (ImageNet, $+2.04$) to the lowest-quality one (\emph{fixed-noise}, $+2.74$) changes $|\Delta_{\mathrm{F}}|$ by only $0.70$ points.  This is exactly the proxy-independent behaviour predicted by $E_{\mathrm{F}}(t)\le C_{\mathrm{F}}$: the (F) gap is governed by the frozen-curvature error $\overline\varepsilon_{\mathrm{F}}$, not by $(\alpha_t,\beta_t)$.

\textbf{Instantiation (O)} matches $E_{\mathrm{O}}\le C_{\mathrm{O}}\,q_t(\alpha_t,\beta_t)$.
The (O) column of Table~\ref{tab:method_specific_proxy_results} traces out an envelope several times wider than either (J) or (F). $|\Delta_{\mathrm{O}}|$ moves from $+0.36$ on ImageNet up to $+5.57$ on \textit{fixed-noise}, and the ordering aligns almost perfectly with the estimated $\hat\alpha$. 
Concretely, $|\Delta_O|$ is within $0.4$ on the two lowest-$\hat\alpha$ proxies (\textit{ImageNet}, \textit{img-sub-u}), rises to roughly $+3.8$ to $+4.5$ on the three mid-$\hat\alpha$ proxies (\textit{sd-img}, \textit{COCO}, \textit{sd-wordnet}), and reaches $+5.05$ and $+5.57$ on the two highest-$\hat\alpha$ proxies (\textit{sd-noise}, \textit{fixed-noise}). This monotone growth of $|\Delta_O|$ with $\hat\alpha$ matches the proxy-amplified envelope of Theorem~\ref{theorem:gap_output}.
Quantitatively, the deviation $D_O$, the accuracy-level counterpart of $E_O$, has a Spearman correlation of $+0.89$ with $\hat\alpha$ across the seven proxies.

\textbf{Direct measurement of $E_\Phi$.} For the linearized and non-linearized task vectors of each instantiation, we evaluate the interference $\mathcal I(\tau\mid\mathcal P_t)$ on held-out target images with Jacobian--vector products at $\theta_0$ and compute $\widehat E_\Phi=|\hat\delta_\Phi-\hat\delta_J|$ (see Appendix~XI-E. Fig.~\ref{fig:gap_direct} shows the result. $\widehat E_F$ stays between 12.1 and 21.4 with no trend in $\hat\alpha$ (Spearman $-0.07$): ImageNet (18.1) and fixed-noise (18.7) give nearly the same gap. $\widehat E_O$ instead grows with $\hat\alpha$, from below 0.6 on ImageNet and img-sub-u to 7.2 on sd-noise and 30.4 on fixed-noise (Spearman $+0.96$). The measured interference thus exhibits the hierarchy of Theorems~\ref{theorem:gap_fisher} and~\ref{theorem:gap_output} directly.

\begin{figure}[t]
  \centering
  \includegraphics[width=\linewidth]{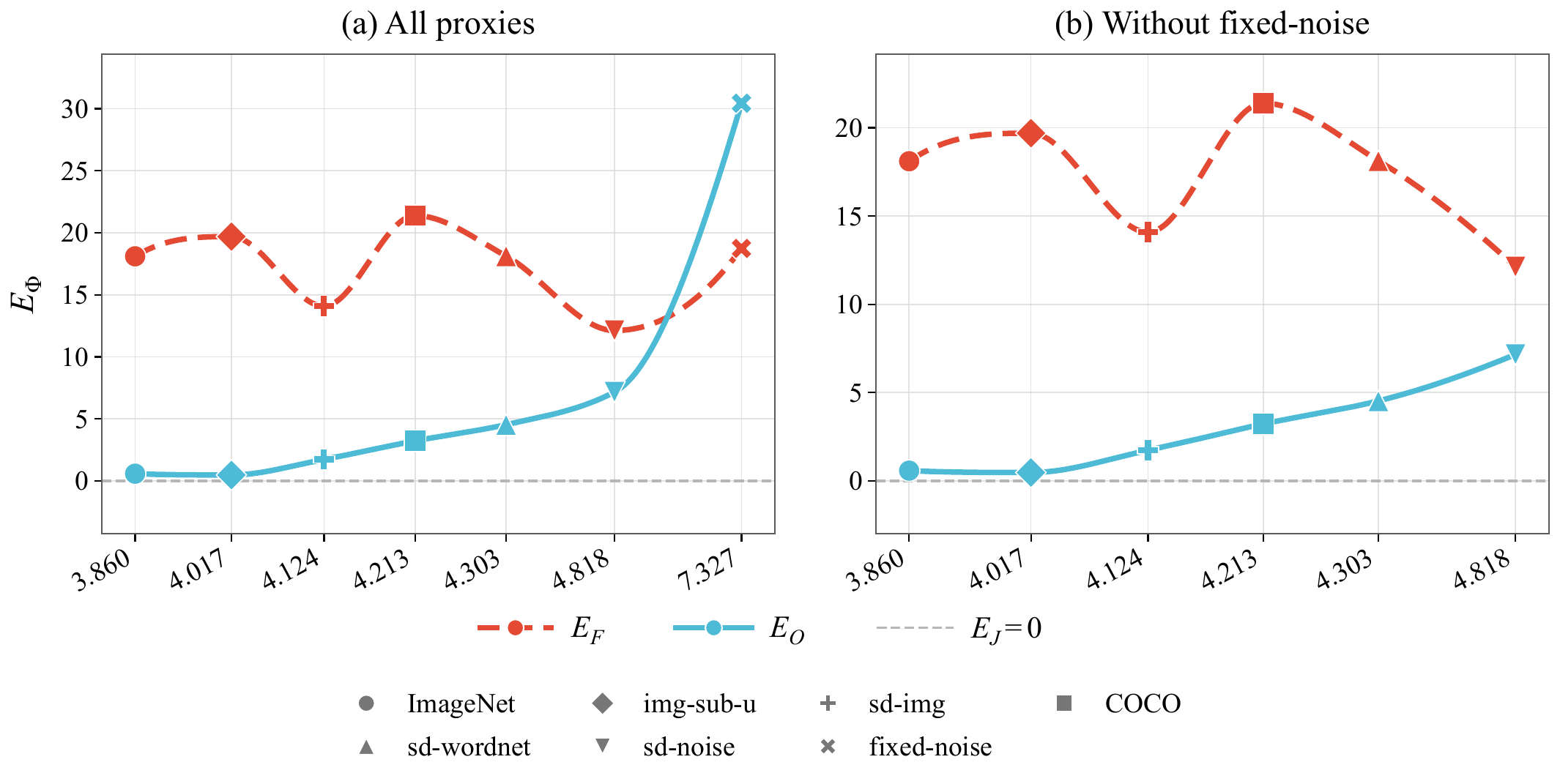}
  \caption{Directly measured regularizer-specific gaps $\widehat E_\Phi$ on ViT-B/32. Proxies are ordered by increasing
  $\hat\alpha_t(\rho)$. (a) All seven proxies. (b) Without fixed-noise. $E_J=0$ by definition (dashed line). $\widehat E_F$ shows no trend in $\hat\alpha$, whereas $\widehat E_O$ grows with it.}
  \label{fig:gap_direct}
\end{figure}

Two practical implications follow. First, on a high-quality universal proxy such as ImageNet, the instantiation (O) reproduces the Jacobian reference (J) to within $1.97$ absolute points without paying the computational cost of Jacobian-vector products (Table~\ref{tab:three_realizations}). 
Second, on a poor proxy the (O) gap widens sharply, so the choice of proxy and fine-tuning mode jointly matters for (O).

\textbf{Computational cost.} Table~\ref{tab:method_specific_efficiency} reports the per-step cost of the three instantiations on ViT-B/32.
Under standard fine-tuning, instantiation (O) is the cheapest at 139.3~ms and 3371.4~MiB, only 16\% slower and 43\% heavier than the unregularized baseline, since it only runs a forward-backward on the output-drift loss with $f(x;\theta_0)$ pre-cached. Instantiation (J) more than doubles the per-step time to 271.3~ms and reaches 4140.6~MiB from the JVP plus VJP sweep, while instantiation (F) needs 237.8~ms but has the heaviest steady footprint of 7513.9~MiB from the resident KFAC factors and an additional 18.45~s of preparation at a 6349.8~MiB peak. Under linear fine-tuning, where the task loss itself already uses JVPs, (J) approaches the unregularized baseline (237.8~ms against 219.1~ms), (O) rises to 289.0~ms, and (F) stays the highest at 356.4~ms. Combined with Table~\ref{tab:clip_scaling_complete}, this leaves (O) as the most favorable option on high-quality proxies. On ImageNet under standard fine-tuning, it trails (J) by 1.49 absolute points while nearly halving the per-step wall-clock and saving 769~MiB of steady memory, whereas (F) is both less accurate than (J) and the heaviest of the three.

\begin{table*}[t]\small
\centering
\caption{Task addition results. We report absolute and normalized accuracies for merging eight tasks across ViT-B/32, ViT-B/16, and ViT-L/14, using both grid-searched and fixed ($\gamma=1$) coefficients. All nine variants of our unified framework use ImageNet as the universal proxy. Across every backbone and fine-tuning strategy, our variants substantially improve over the corresponding unregularized baselines and approach the oracle $\tau_{Jp}$ upper bound that requires access to peer-task data. At $\gamma=1$, only (O) stays close to its grid-searched accuracy across all settings.}
\label{tab:clip_scaling_complete}
\adjustbox{max width=\textwidth}{
\footnotesize
\setlength{\tabcolsep}{4pt}
\begin{tabular}{@{}>{\centering\arraybackslash}m{4.6cm}cccccccccccc@{}}
\toprule
\multirow{3}{4.2cm}{\centering Method} &
\multicolumn{4}{c}{ViT-B/32, 8 tasks} &
\multicolumn{4}{c}{ViT-B/16, 8 tasks} &
\multicolumn{4}{c}{ViT-L/14, 8 tasks} \\
\cmidrule(lr){2-5}\cmidrule(lr){6-9}\cmidrule(lr){10-13}
& \multicolumn{2}{c}{grid search} & \multicolumn{2}{c}{$\gamma=1$} &
\multicolumn{2}{c}{grid search} & \multicolumn{2}{c}{$\gamma=1$} &
\multicolumn{2}{c}{grid search} & \multicolumn{2}{c}{$\gamma=1$} \\
\cmidrule(lr){2-3}\cmidrule(lr){4-5}\cmidrule(lr){6-7}\cmidrule(lr){8-9}\cmidrule(lr){10-11}\cmidrule(lr){12-13}
& \textbf{Abs.} ($\uparrow$) & \textbf{Norm.} ($\uparrow$) & \textbf{Abs.} ($\uparrow$) & \textbf{Norm.} ($\uparrow$) & 
\textbf{Abs.} ($\uparrow$) & \textbf{Norm.} ($\uparrow$) & \textbf{Abs.} ($\uparrow$) & \textbf{Norm.} ($\uparrow$) & 
\textbf{Abs.} ($\uparrow$) & \textbf{Norm.} ($\uparrow$) & \textbf{Abs.} ($\uparrow$) & \textbf{Norm.} ($\uparrow$) \\
\midrule
zero-shot & 47.74 & -- & -- & -- & 54.22 & -- & -- & -- & 64.54 & -- & -- & -- \\
\rowcolor{lightgray}
$\tau_{Jp}$ (oracle peer data)~\cite{taujp2025} & 86.74 & 99.09 & 87.43 & 98.96 & 87.60 & 98.50 & 87.50 & 98.40 & 90.80 & 99.00 & 90.80 & 99.00 \\
\midrule
Nonlin.~FT~\cite{ta2023}  & 70.32 & 77.56 & 20.07 & 20.68 & 75.39 & 81.02 & 17.87 & 18.39 & 84.07 & 89.19 & 29.47 & 30.60 \\
TTA~\cite{tta2023}        & 74.86 & 85.35 & 56.09 & 61.95 & 78.97 & 87.48 & 57.49 & 62.70 & 86.19 & 93.14 & 81.12 & 87.55 \\
ATT-FT~\cite{linearatt2025} & 78.07 & 86.99 & 24.44 & 25.31 & 80.71 & 87.64 & 24.41 & 25.14 & 87.81 & 93.59 & 60.71 & 63.93 \\
\midrule
Nonlin.~FT + (J) & 83.73 & 92.53 & 48.72 & 52.08 & 85.92 & 92.82 & 55.13 & 58.35 & 91.64 & 97.92 & 90.51 & 96.72 \\
Nonlin.~FT + (F) & 77.03 & 84.68 & 34.16 & 35.67 & 80.30 & 86.39 & 37.01 & 38.16 & 88.64 & 93.87 & 72.83 & 76.84 \\
Nonlin.~FT + (O) & 82.24 & 90.85 & 75.76 & 83.83 & 86.00 & 92.86 & 81.12 & 88.25 & 90.88 & 96.74 & 89.78 & 95.67 \\
\midrule
TTA + (J) & 81.40 & 93.31 & 77.46 & 88.56 & 85.01 & 94.61 & 81.07 & 90.83 & 90.25 & 98.10 & 90.20 & 97.93 \\
TTA + (F) & 79.07 & 90.47 & 68.55 & 78.02 & 82.39 & 91.38 & 62.67 & 69.08 & 88.59 & 95.73 & 84.11 & 90.68 \\
TTA + (O) & 81.88 & 93.30 & 77.91 & 88.51 & 84.63 & 93.74 & 79.73 & 88.55 & 90.19 & 97.53 & 90.37 & 97.76 \\
\midrule
ATT-FT + (J) & 82.79 & 92.81 & 39.04 & 43.44 & 85.20 & 93.23 & 37.04 & 40.48 & 90.55 & 97.48 & 82.11 & 88.36 \\
ATT-FT + (F) & 78.74 & 88.00 & 23.96 & 25.03 & 81.67 & 88.76 & 27.47 & 28.47 & 88.52 & 94.37 & 58.44 & 61.51 \\
ATT-FT + (O) & \textbf{84.64} & \textbf{94.39} & \textbf{78.01} & \textbf{87.94} & \textbf{86.99} & \textbf{95.39} & \textbf{82.67} & \textbf{91.83} & \textbf{91.81} & \textbf{98.25} & \textbf{91.64} & \textbf{97.86} \\
\bottomrule
\end{tabular}
}
\end{table*}

\subsection{Practical Configurations and Overall Comparison}
We now turn to the practical question: given a task-agnostic proxy such as ImageNet, how do the three instantiations of our framework compare against existing pre-merging baselines? Throughout this subsection, we adopt ImageNet as the default universal proxy, motivated by its near-oracle spectral quality established in Section~\ref{subsec:exp_spectral}.

\textbf{Baselines.} We compare against the following: (1) standard fine-tuning (Nonlin.~FT)~\cite{ta2023}; (2) Tangent Task Arithmetic (TTA)~\cite{tta2023}, \textit{i.e.}, linearized fine-tuning; (3) ATT-FT, attention-only fine-tuning~\cite{linearatt2025}; and (4) $\tau_{Jp}$~\cite{taujp2025} instantiated with the true peer-task data, serving as an oracle upper bound that violates the independence assumption of model merging. Because our unified framework is orthogonal to the fine-tuning strategy, each instantiation $\Phi\in\{(J),(F),(O)\}$ can be paired with any strategy $\mathcal{S}\in\{\text{Nonlin.~FT},\text{TTA},\text{ATT-FT}\}$, yielding nine universal-proxy variants of the form $\mathcal{S}+\Phi+\text{ImageNet}$, all evaluated on the same eight-task benchmark.

\textbf{Overall performance comparison.} Table~\ref{tab:clip_scaling_complete} shows that our unified framework, when paired with ImageNet as the universal proxy, consistently closes the gap to the oracle $\tau_{Jp}$ across all three fine-tuning strategies and all three backbone scales. The improvement over unregularized baselines is substantial: on ViT-B/32, adding any of the three instantiations to Nonlin.~FT lifts absolute accuracy by $6.7$--$13.4$ percentage points, with (J) reaching $83.73\%$ and (O) reaching $82.24\%$. As the backbone scales, all three instantiations converge to near-oracle performance. 
On ViT-L/14, Nonlin.~FT+(J)+ImageNet reaches $91.64\%/97.92\%$, slightly exceeding on absolute accuracy the oracle $\tau_{Jp}$ ($90.80\%/99.00\%$), which requires access to peer-task data that our method never sees. In the analyses that follow, we report results for instantiation (O) as a representative, and (J) and (F) yield qualitatively similar patterns.

\textbf{$\gamma$-robustness is instantiation-specific.}
Here $\gamma$ is the merging coefficient of Eq.~\eqref{eq:merge}, and fixing $\gamma = 1$ removes the grid search over it, following the protocol also reported in~\cite{taujp2025}.
Table~\ref{tab:clip_scaling_complete} additionally reports every variant at the fixed coefficient $\gamma=1$. All three unregularized baselines collapse (\emph{e.g.}, Nonlin.~FT drops from $70.32\%$ to $20.07\%$ on ViT-B/32), but the three instantiations behave very differently. Instantiation (O) is uniformly robust, staying within $6.63$ points of its grid-searched accuracy across all nine configurations. Instantiation (J) retains its accuracy under linearized fine-tuning (within $3.94$ points under TTA) but loses $35.01$ and $43.75$ points under Nonlin.~FT and ATT-FT on ViT-B/32. Instantiation (F) degrades under every strategy.
This pattern follows from Theorems~\ref{theorem:gap_fisher} and \ref{theorem:gap_output}. Instantiations (J) and (F) are frozen quadratic forms and thus track the true functional deviation only within the linearization neighborhood of $\theta_0$. At $\gamma=1$ the aggregate shift $\sum_t\tau_t$ exits this neighborhood and the surrogate ceases to constrain what it is meant to control. Instantiation (O), by contrast, penalizes the exact non-linear output drift (Eq.~\eqref{eq:odr_definition}) and continues to constrain the actual deviation. 

\begin{figure}[t]
  \centering
   \includegraphics[width=0.9\linewidth]{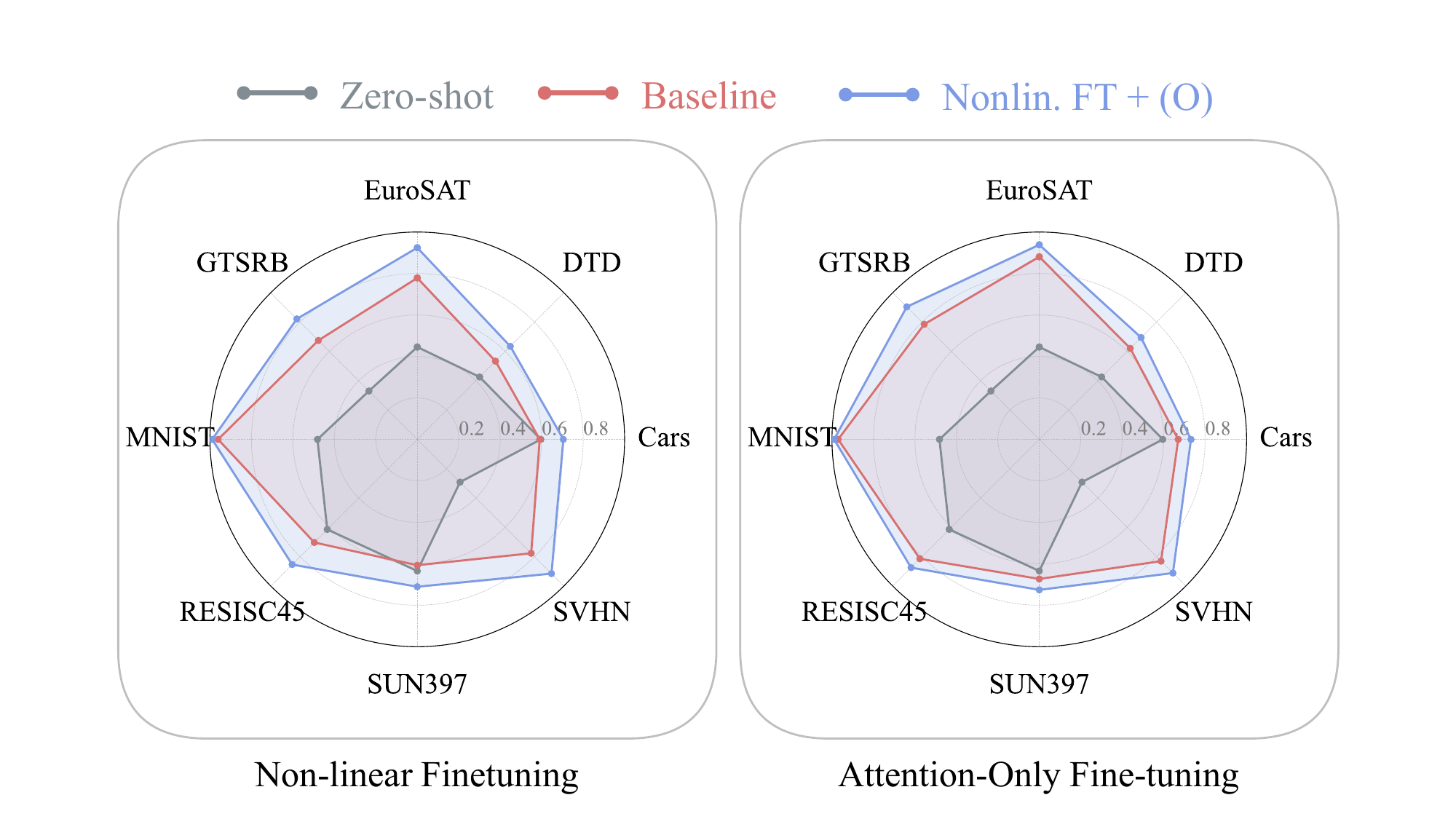}
   \caption{Accuracy of merged models (ViT-B/32) across the eight tasks. Each subplot compares zero-shot (gray), a baseline merging method (red), and the baseline enhanced with instantiation (O) on ImageNet (blue). The performance polygon of the regularized variant consistently encompasses the baseline.
   }
   \label{fig:radar_ViT-B-32}
\end{figure}

\textbf{Per-task performance analysis.} As Figure~\ref{fig:radar_ViT-B-32} shown, the performance polygon of Nonlin.~FT$+$(O) consistently encompasses that of Nonlin.~FT, with no task showing degradation. This indicates that ImageNet-based proxy regularization suppresses cross-task interference uniformly rather than trading off one task against another, preserving specialist knowledge on every task while improving merged performance.

\begin{figure}[t]
  \centering
   \includegraphics[width=\linewidth]{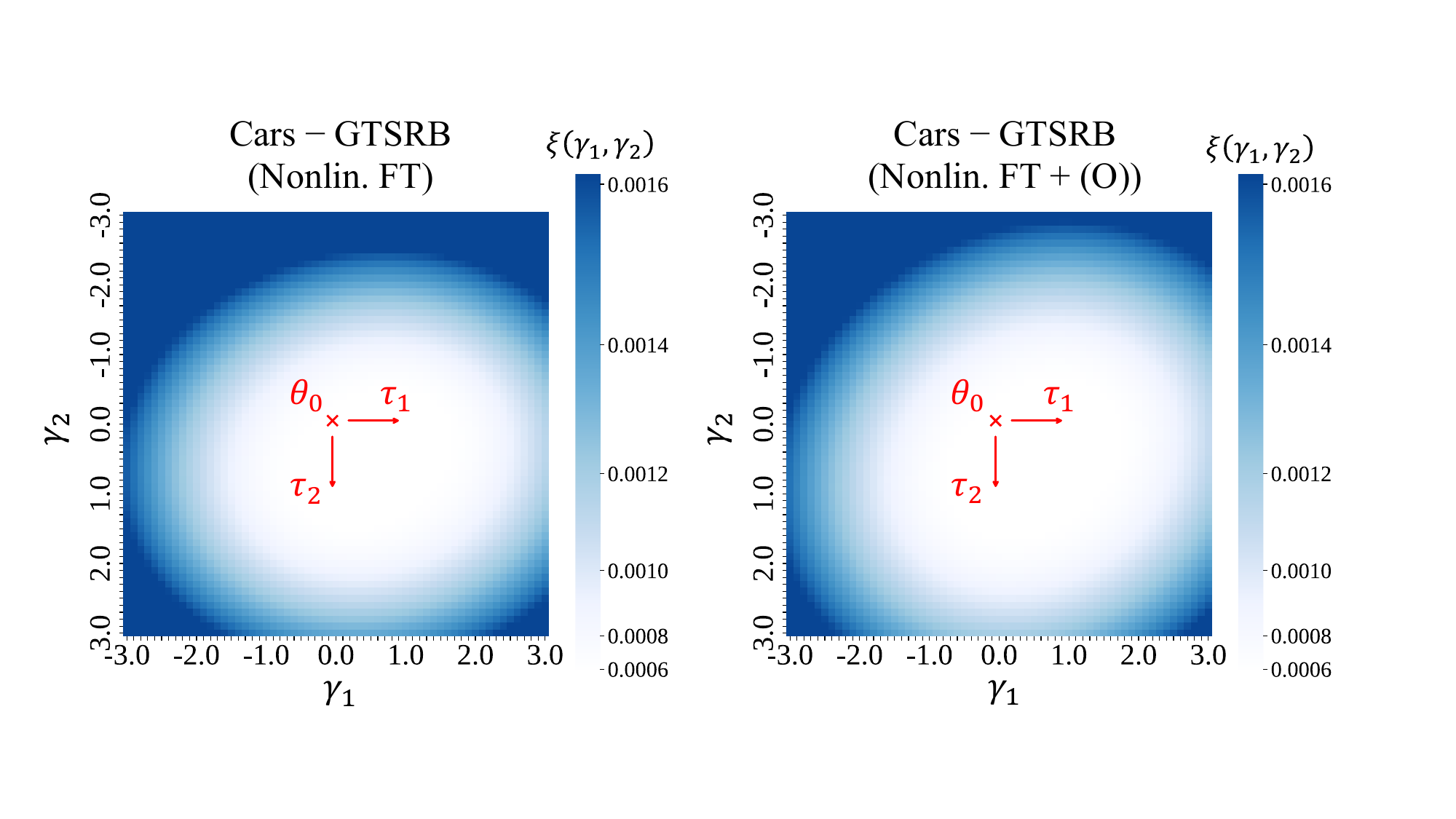}
   \caption{   Visualization of weight disentanglement. The heatmaps compare the disentanglement error $\xi(\gamma_1, \gamma_2)$ of Nonlin.~FT (left) and our method Nonlin.~FT$+$(O) (right) on the Cars--GTSRB task pair using ViT-B/32. Lighter regions denote areas of the weight space where weight disentanglement is stronger (lower functional interference). Our method maintains a significantly broader region of low interference, indicating superior composability.
   }
   \label{fig:weight_disentanglement_nonlin_EuroSAT_RESISC45_2}
\end{figure}

\begin{figure}[t]
    \centering
    \subfloat[$\lambda$ Sensitivity\label{fig:lambda_sensitivity_1}]{
        \includegraphics[width=0.45\columnwidth]{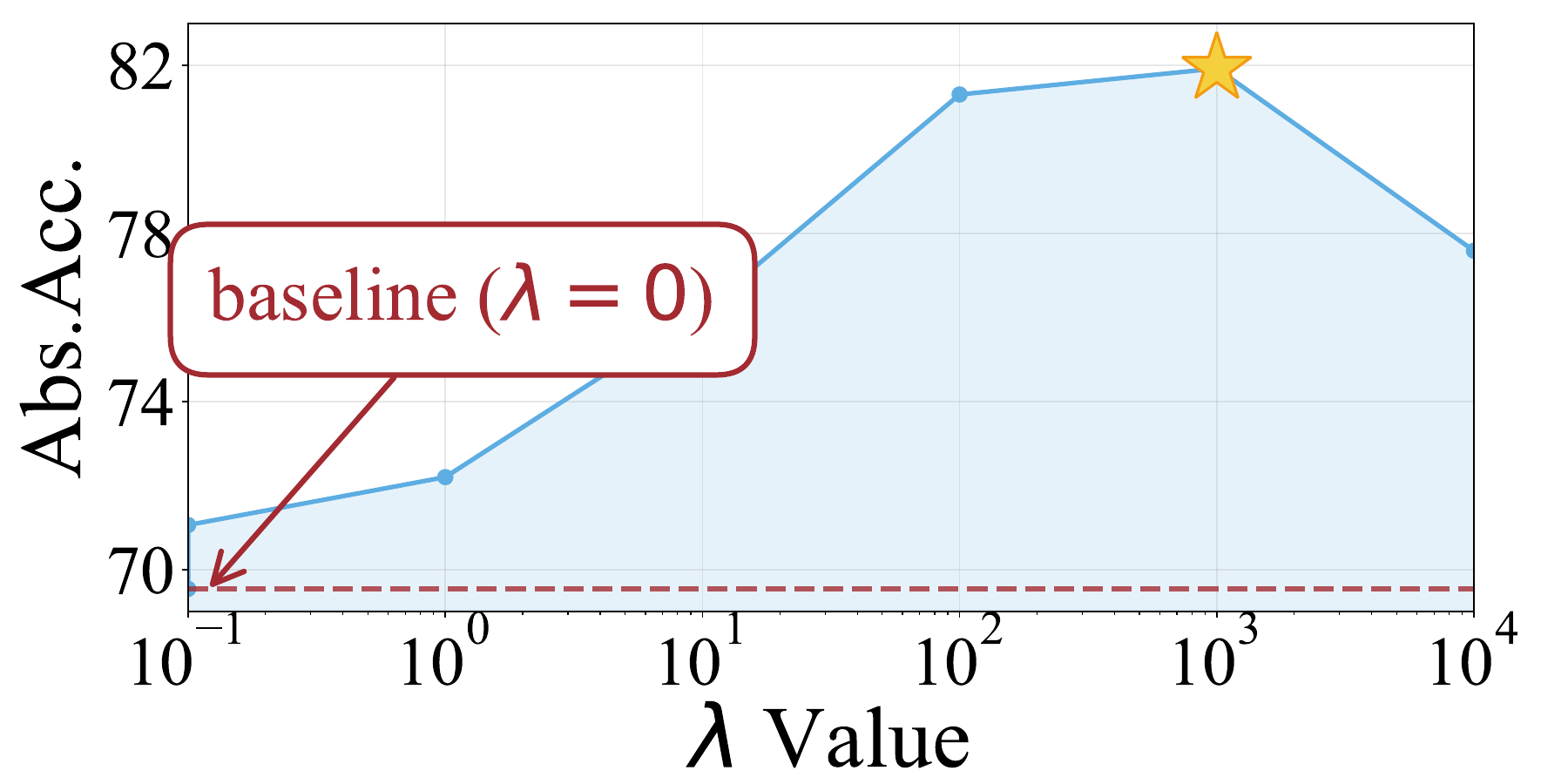}
    }
    \hfil %
    \subfloat[$\gamma$ Sensitivity\label{fig:alpha_sensitivity_nonlin_ft}]{
        \includegraphics[width=0.45\columnwidth]{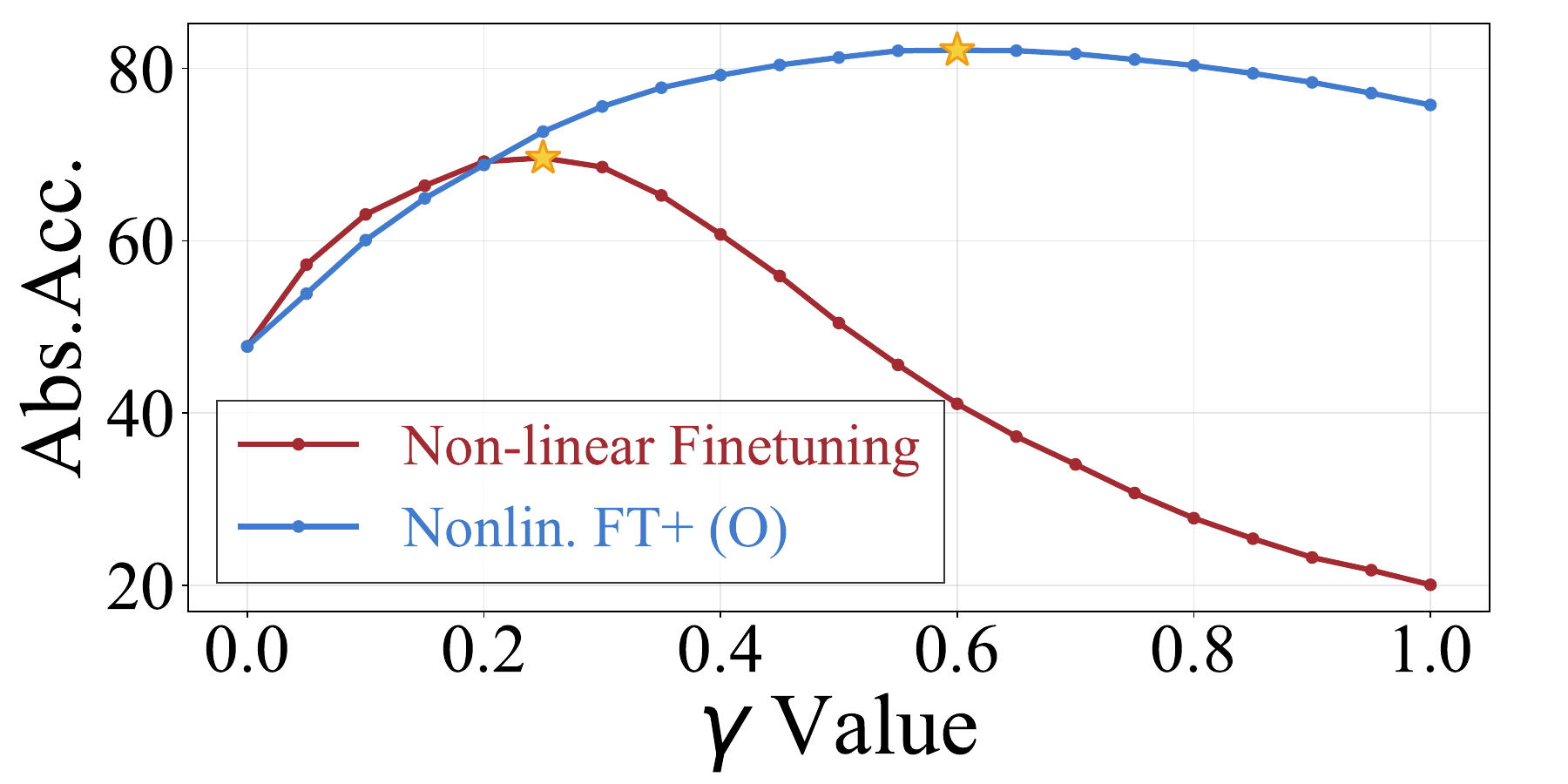}
    }
    \caption{    Analysis of hyperparameter sensitivity on ViT-B/32, using instantiation (O) with ImageNet. (a) Impact of the regularization strength $\lambda$. (b) Influence of the merging coefficient $\gamma$: our method (blue) maintains superior performance across a broad range of $\gamma$ values compared with baseline (red).}
    \label{fig:Analysis_of_hyperparameter_sensitivity_on_ViT-B-32}
\end{figure}

\textbf{Parameter sensitivity analysis.} 
We examine the sensitivity of our method to two key hyperparameters: the regularization strength $\lambda$ and the merging coefficient $\gamma$. 
Figure~\ref{fig:lambda_sensitivity_1} shows that performance improves monotonically with increasing $\lambda$ up to $\lambda=10^3$, confirming that interference suppression directly benefits from the universal proxy rather than from a narrow range of favorable hyperparameter values.
Figure~\ref{fig:alpha_sensitivity_nonlin_ft} demonstrates that Nonlin.~FT$+$(O) maintains substantially higher accuracy than the baseline across a broad range of $\gamma$. This robustness indicates that Nonlin.~FT$+$(O) produces inherently higher-quality task vectors, eliminating the need for careful coefficient tuning and making the merging procedure more stable in practice.

\textbf{Analysis of weight disentanglement.} To investigate the mechanism underlying the performance gains, we visualize the weight disentanglement landscape following~\cite{tta2023}, where a broader region of low cross-task functional interference indicates a larger weight-disentangled basin. Figure~\ref{fig:weight_disentanglement_nonlin_EuroSAT_RESISC45_2} reveals that our method (right) maintains a substantially broader weight-disentangled basin compared to the baseline (left). This expanded basin confirms that ImageNet-based proxy regularization effectively promotes weight disentanglement by orthogonalizing task vectors in functional space, directly explaining the superior task-addition performance.

\textbf{Compatibility with during-merging methods.} Although our framework operates at the pre-merging stage, we evaluate whether the resulting checkpoints can enhance during-merging algorithms that resolve conflicts at the aggregation stage. We test the task vectors produced by Nonlin.~FT$+$(O)$+$ImageNet with four representative during-merging methods: Ties-Merging~\cite{Tiesmerging2023}, AdaMerging~\cite{Adamerging2024}, Task Singular Vectors~\cite{tsv2025}, and Representation Surgery~\cite{Representation_surgery}. As shown in Table~\ref{tab:during_merging}, our regularized checkpoints consistently improve all four algorithms. Combining with TSV, for instance, boosts performance from $85.64\%$ to $90.33\%$. This demonstrates that higher-quality pre-merging task vectors directly yield better final multi-task models. The pre-merging and during-merging axes compose rather than compete.

\begin{table}[t]\small
\small
\centering
\caption{Compatibility with during-merging methods. We compare the performance of advanced merging techniques when applied to task vectors derived from standard fine-tuning versus fine-tuning regularized with our unified framework (instantiation (O), ImageNet proxy).}
\label{tab:during_merging}
\begin{tabular}{ccc}
\toprule
\textbf{Merging Method} & \textbf{Nonlin. FT} & \textbf{Nonlin.~FT + (O)} \\
\midrule
Ties-Merging~\cite{Tiesmerging2023} & 72.73 & \textbf{81.20} \\
Task Singular Vectors~\cite{tsv2025} & 85.64 & \textbf{90.33} \\
Representation Surgery~\cite{Representation_surgery} & 80.70 & \textbf{85.35} \\
AdaMerging\cite{Adamerging2024} (task-wise) & 70.71 & \textbf{82.74} \\
\bottomrule
\end{tabular}
\end{table}

\subsection{Ablation Studies}
\label{sec:ablation}

\textbf{Proxy sample size.} The proxies in Section~\ref{subsec:exp_spectral} differ widely in size, from a single fixed batch for fixed-noise to the 1.28M images of ImageNet. To verify that the differences among proxies do not come from their size, we restrict every proxy to the same number $n\in\{128,256,512,1000\}$ of images, drawn with a fixed seed, and repeat the experiment of Section~~\ref{subsec:exp_spectral} with all other settings unchanged. 
Table~\ref{tab:proxy_size} shows that the ranking of the proxies is nearly identical across all $n$ and the full proxies. 
In every column, Oracle Union ranks first and img-sub-u, sd-noise and fixed-noise rank last in this order. Only the order among ImageNet, sd-img, COCO and sd-wordnet changes, and their accuracies lie within one point of each other.
Moreover, with $n=1000$ every proxy is within 0.7 points of its full counterpart, so enlarging the proxy beyond 1,000 images brings little further gain. Even with 128 images, every proxy stays within 1.8 points of the full proxy. This indicates that the proxy does not need to be large, and a few hundred images are already sufficient for effective regularization.

\begin{table}[t]
\centering
\caption{Absolute accuracy (\%) of linearized fine-tuning with (J) on ViT-B/32 when every proxy is restricted to $n$ images. Full denotes the complete proxy.}
\label{tab:proxy_size}
\setlength{\tabcolsep}{5pt}
\begin{tabular}{lccccc}
\toprule
Proxy & $n{=}128$ & $256$ & $512$ & $1000$ & Full \\
\midrule
Oracle Union & 84.43 & 84.87 & 85.06 & 85.18 & 85.14 \\
ImageNet     & 80.79 & 81.23 & 81.37 & 81.64 & 81.40 \\
img-sub-u    & 78.91 & 79.32 & 79.89 & 80.11 & 80.67 \\
sd-img       & 80.35 & 80.95 & 81.04 & 81.49 & 81.46 \\
COCO         & 79.87 & 80.54 & 80.71 & 81.05 & 81.65 \\
sd-wordnet   & 80.72 & 81.12 & 81.25 & 81.40 & 81.19 \\
sd-noise     & 78.77 & 79.06 & 79.29 & 79.17 & 79.53 \\
fixed-noise  & 73.19 & 73.20 & 73.18 & 73.26 & 72.57 \\
\bottomrule
\end{tabular}
\end{table}

\begin{table}[t]
\centering
\caption{Merged absolute accuracy (\%) of linearized fine-tuning on ViT-B/32 when an explicit ridge term $\frac{\mu}{2}\|\theta-\theta_0\|^2$ is added to the default fine-tuning with weight
decay. The proxy rows use (J), and none denotes fine-tuning without the proxy penalty. The last row gives the Pearson correlation between the task-averaged $\widehat\alpha_t(\rho)$ and the merged accuracy over the seven proxies.}
\label{tab:mu}
\setlength{\tabcolsep}{5pt}
\begin{tabular}{lccccc}
\toprule
Proxy & $\mu{=}0$ & $10^{-6}$ & $10^{-5}$ & $10^{-4}$ & $1$ \\
\midrule
none        & 74.86 & 75.26 & 75.09 & 75.12 & 68.23 \\
\midrule
ImageNet    & 81.40 & 82.35 & 82.19 & 82.17 & 76.37 \\
img-sub-u   & 80.67 & 80.71 & 80.74 & 80.72 & 73.96 \\
sd-img      & 81.46 & 81.28 & 81.29 & 81.23 & 75.94 \\
COCO        & 81.65 & 81.31 & 81.28 & 81.28 & 74.95 \\
sd-wordnet  & 81.19 & 81.65 & 81.63 & 81.52 & 76.63 \\
sd-noise    & 79.53 & 79.22 & 79.18 & 79.17 & 74.27 \\
fixed-noise & 72.57 & 72.79 & 72.78 & 72.82 & 69.48 \\
\midrule
Pearson $r$ & $-0.987$ & $-0.987$ & $-0.988$ & $-0.989$ & $-0.918$ \\
\bottomrule
\end{tabular}
\end{table}

\textbf{Effect of $\mu$.} The ridge term $\frac{\mu}{2}\|\tau\|^2$ in Eqs.~\eqref{formula:objective} and~\eqref{eq:generic_regularized_objective} keeps the task vector bounded in directions not covered by $\mathcal{Q}$, a role that weight decay plays approximately in our fine-tuning. To examine the ridge strength in the exact form of the analysis, we add an explicit ridge $\frac{\mu}{2}\|\theta-\theta_0\|^2$ to the setting of Section~\ref{subsec:exp_spectral} with $\mu\in\{10^{-6},10^{-5},10^{-4},1\}$ (Table~\ref{tab:mu}).
For $\mu\le10^{-4}$, the accuracy of every universal proxy stays within one point of that at $\mu=0$, and the Pearson correlation with $\widehat\alpha_t(\rho)$ stays between $-0.987$ and $-0.989$.
At $\mu=1$, the ridge restricts the task vector in all directions, including those the task itself needs: the single-task accuracy decreases by about 2 points, and the merged accuracy by 4.6--6.7 points on the natural-image and synthetic proxies and by 3.1 points on fixed-noise. Without the proxy penalty, the same ridge lowers the accuracy from 74.86\% to 68.23\%, while (J) stays 5.7--8.4 points above it on the natural-image and synthetic proxies, so the gain of the proxy penalty persists under a strong ridge. The correlation with $\widehat\alpha_t(\rho)$ remains strong ($r=-0.92$), consistent with Corollary~1(ii), whose ordering holds for every $\mu$.

\section{Extension to Large Language Models}
\label{sec:llm}
The spectral framework of Sections~\ref{sec:framework}--\ref{sec:proxy_regularization} is not restricted to the visual modality. The interference bound depends only on how well the proxy $\mathcal Q$ covers the directions of the inaccessible targets $\{\mathcal P_t\}$. We validate this on large language models (LLMs) through two experiments that cover different backbones, fine-tuning regimes, and downstream benchmarks, and compare against representative pre-merging baselines that were not designed for a universal proxy.

\subsection{Representation-Space Instantiation}
\label{sec:llm_method}
 
We take the hidden representation as the LLM's model output for the framework. Let $H(x;\theta)\in\mathbb{R}^{L\times d}$ denote the final-layer hidden states of the LLM and $p(x)$ the position of the last non-padding token. We set
\begin{equation}\small
    h(x;\theta) \;:=\; H(x;\theta)_{p(x)} \;\in\;\mathbb{R}^{d}
    \label{eq:llm_representation}
\end{equation}
and treat it as the model output $f$ from Section~\ref{sec:preliminaries}. With this definition, the Jacobian covariance operator $A_{\mathcal{Q}}=\mathbb{E}_{x\sim\mathcal{Q}}[J_h(x;\theta_0)J_h(x;\theta_0)^{\top}]$ and the three instantiations (J), (F), (O) of Section~\ref{sec:proxy_regularization} apply directly.

We evaluate instantiations (J) and (O). We do not consider (F) because its frozen curvature $\widehat A_{\mathcal Q}$ must be estimated at $\theta_0$ and held throughout fine-tuning, which is prohibitive at LLM scale even in Kronecker-factored form. We focus primarily on (O).
Its reference $h(x;\theta_0)$ is a single vector per proxy sequence and, being independent of $\theta$, can be pre-computed once at $\theta_0$ and cached in bfloat16 on CPU. Each fine-tuning step then adds only one auxiliary forward-backward pass through the proxy mini-batch, without keeping a second LLM resident on GPU. For (J), both experiments use the quadratic penalty $\mathbb{E}_{x\sim\mathcal Q}\|J_h(x;\theta_0)^{\top}\tau_s\|_2^2=\tau_s^{\top}A_{\mathcal Q}\tau_s$ of Section~\ref{sec:proxy_regularization}.

\begin{table}[t]\small
\centering
\caption{Macro-averaged score (\%) of Task Arithmetic on the seven TRACE tasks. WikiText-103 serves as the universal proxy. Each method uses the merging coefficient $\gamma$ that maximizes its seven-task average. Within each optimizer module, \textbf{bold} denotes the best result on each backbone.}
\label{tab:trace_1b}
\begin{tabular}{lcc}
\toprule
Method & Llama-3.2-1B & Llama-3.2-3B \\
\midrule
AdamW            & 41.07          & 49.39 \\
AdamW $+$ (J)    & 42.69          & \textbf{51.69} \\  %
AdamW $+$ (O)    & \textbf{43.02} & 51.20 \\
\midrule
MergOPT          & 42.75          & 50.26 \\
MergOPT $+$ (J)  & 42.86          & \textbf{52.13} \\
MergOPT $+$ (O)  & \textbf{44.72} & 51.46 \\
\bottomrule
\end{tabular}
\end{table}

\subsection{Experiment I: Full-Parameter Fine-Tuning on TRACE}\label{sec:llm_trace}
\textbf{Experimental setup.} We follow the fine-tuning and merging protocol of MergOPT~\cite{mergopt2026} on the seven tasks of TRACE~\cite{trace2023}, training one expert per task. Two backbones are considered, Llama-3.2-1B-Instruct and Llama-3.2-3B-Instruct~\cite{llama3}. 
We augment this pipeline with instantiations~(O) and~(J), both using $256$ sequences from WikiText-103~\cite{wikitext103-2017}, a standard, publicly available general-domain Wikipedia corpus, as the universal proxy $\mathcal Q$. It is entirely unrelated to TRACE and contains none of its labels or samples.
Full hyperparameters, evaluation metrics, and implementation details are deferred to Appendix~XII-A.
The seven experts are merged with Task Arithmetic, and each method is reported at the merging coefficient $\gamma$ that maximizes its seven-task average.

\textbf{Baselines.} We compare against two representative pre-merging baselines. Standard AdamW fine-tuning provides the unregularized reference. MergOPT~\cite{mergopt2026} is a merge-aware optimizer that samples synthesized weight-space perturbations during fine-tuning.
Because instantiations~(O) and~(J) are orthogonal to the choice of optimizer, we pair each of them with both baselines, yielding configurations of the form $\mathcal S + (O)$ and $\mathcal S + (J)$ with $\mathcal S \in \{\text{AdamW},\text{MergOPT}\}$.

\textbf{Results and analysis.} Table~\ref{tab:trace_1b} reports the merged performance with the task-agnostic WikiText-103 proxy.
Instantiation~(O) improves the seven-task average in all four configurations, by $1.20$ to $1.97$ points, in line with the vision-domain results of Section~\ref{sec:experiments}.
The gain holds not only over AdamW but also on top of MergOPT, so the proxy penalty remains effective when combined with a merge-aware optimizer.
Instantiation~(J) improves the average in three of the four configurations, by $1.62$ to $2.30$ points, and leaves it nearly unchanged with MergOPT on Llama-3.2-1B.
Per-task results are given in Appendix~XII-A.

\textbf{Computational efficiency.}
Table~\ref{tab:llm_efficiency} reports a 100-step controlled benchmark on Llama-3.2-1B. Instantiation~(O) requires $0.86$s per step and $15.89$ GiB peak GPU memory, adding only $0.48$ GiB ($3.1\%$) over plain AdamW because the pre-cached reference representations remove the need to keep a second LLM resident on GPU. The direct Jacobian implementation instead uses $0.96$s per step and $32.75$ GiB, more than double the peak memory of plain AdamW.

\begin{table}[t]\small 
\centering
\caption{Per-step cost of instantiations~(O) and~(J) on Llama-3.2-1B. Percentages in parentheses are relative to AdamW.}
\label{tab:llm_efficiency}
\setlength{\tabcolsep}{6pt}
\begin{tabular}{lcc}
\toprule
Method & Sec./step & Peak alloc.\ (GiB) \\
\midrule
AdamW           & 0.649          & 15.41 \\
AdamW $+ (O)$   & 0.862 (+32.8\%) & 15.89 (+3.1\%) \\
AdamW $+ (J)$   & 0.962 (+48.2\%) & 32.75 (+112.5\%) \\
\bottomrule
\end{tabular}
\end{table}

\subsection{Experiment II: LoRA Fine-Tuning on GLUE}\label{sec:llm_glue}
In Experiment~II, we follow the setup of OSRM~\cite{osrm2025}, which fine-tunes a Llama-3-8B backbone with rank-$8$ LoRA adapters on the eight GLUE tasks~\cite{glue2019} and merges the resulting specialists. OSRM constrains each expert's LoRA subspace prior to fine-tuning using examples drawn from the other GLUE tasks to be merged, and thus stands as a LoRA-specific pre-merging baseline that has undesired explicit access to peer-task data. We augment this pipeline with our universal-proxy regularizer and ask two questions: whether the proxy still suppresses interference once the task vector is a low-rank, bilinearly parameterized update, and, if so, how far a task-agnostic proxy closes the gap to this peer-data baseline, which is expected to serve as an upper bound.

\textbf{Experimental setup.} We follow the LoRA fine-tuning and evaluation protocol of OSRM~\cite{osrm2025} on the eight GLUE tasks (CoLA, SST-2, MRPC, STS-B, QQP, MNLI, QNLI, RTE). Rank-$8$ LoRA~\cite{lora2022} adapters are applied to the query and value projections of every self-attention block, both LoRA factors are trainable. We augment this pipeline with two instantiations of our framework, (O) and (J), using $2{,}048$ task-agnostic WikiText-103~\cite{wikitext103-2017} sequences as the universal proxy $\mathcal Q$. And the proxy contains no GLUE labels or samples. Full details are deferred to Appendix~XII-B.

\textbf{Baselines.} We compare against: (i) standard LoRA fine-tuning; and (ii) OSRM~\cite{osrm2025}, a state-of-the-art LoRA-specific pre-merging baseline that constrains each expert's LoRA subspace prior to fine-tuning by using some examples per peer GLUE task. 
OSRM therefore has explicit access to information that our universal-proxy methods do not use, and serves as an oracle-style upper reference like $\tau_{Jp}$~\cite{taujp2025} in Section~\ref{sec:experiments}. 

\begin{table*}[t]\small
\centering
\caption{Eight-task merging of Llama-3-8B GLUE specialists fine-tuned with LoRA, under Task Arithmetic and TIES-Merging. All values are percentages. ``Peer data'' indicates whether the fine-tuning stage accesses training examples from the other tasks to be merged. Norm.\ is the mean of the eight per-task retention ratios with respect to the corresponding specialists. \textbf{Bold} and \underline{underline} denote the best and second-best result in each column, excluding the oracle OSRM (shaded).}
\label{tab:llm_glue}\label{tab:osrm}
\setlength{\tabcolsep}{7pt}
\begin{tabular}{lcccccccc}
\toprule
\multirow{2}{*}{Method} & \multirow{2}{*}{Inst.} & \multirow{2}{*}{Peer data} & \multirow{2}{*}{Specialist Avg.}
 & \multicolumn{2}{c}{Task Arithmetic} & \multicolumn{2}{c}{TIES-Merging} \\
\cmidrule(lr){5-6}\cmidrule(lr){7-8}
 & & & & Abs.\ ($\uparrow$) & Norm.\ ($\uparrow$) & Abs.\ ($\uparrow$) & Norm.\ ($\uparrow$) \\
\midrule
Standard LoRA & -- & No & \textbf{87.30} & 62.81 & 70.76 & 59.46 & 66.63 \\
\rowcolor{lightgray}
OSRM~\cite{osrm2025} & -- & Yes & 88.53 & 75.01 & 83.98 & 67.66 & 75.52 \\
\midrule[\heavyrulewidth]
Standard LoRA +(O) & (O) & No & \underline{86.27} & \textbf{74.48} & \textbf{85.93} & \underline{67.88} & \underline{78.20} \\
Standard LoRA +(J) & (J) & No & 86.08 & \underline{74.18} & \underline{85.77} & \textbf{70.24} & \textbf{80.86} \\
\bottomrule
\end{tabular}
\end{table*}

\begin{figure}[t]
  \centering
   \includegraphics[width=\linewidth]{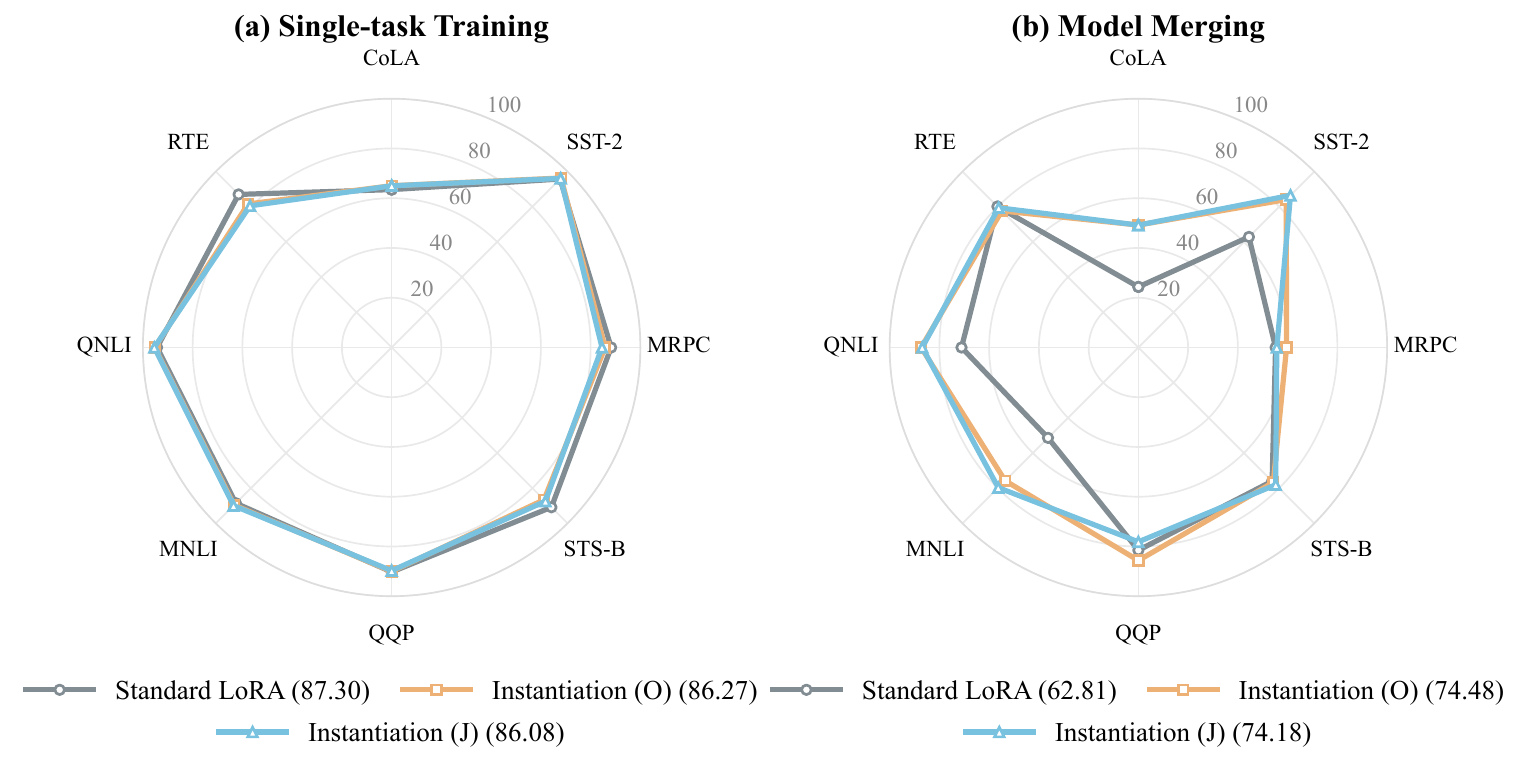}
   \caption{Per-task GLUE scores of standard LoRA, (O), and (J) on Llama-3-8B, with WikiText-103 as the proxy. (a) Specialists before merging; (b) merged models under Task Arithmetic. Legend values are eight-task averages.}
   \label{fig:radar_glue}
\end{figure}

\textbf{Results and analysis.}
Table~\ref{tab:llm_glue} summarizes the eight-task merging results under both Task Arithmetic~\cite{ta2023} and TIES-Merging~\cite{Tiesmerging2023}. Per-task results are deferred to Appendix~XII-B. The results are consistent with the vision-domain findings of Section~\ref{sec:experiments}.

\textit{A task-agnostic proxy approaches the peer-data oracle.} 
Under Task Arithmetic, standard LoRA drops to an Abs.\ of $62.81\%$ when the eight specialists are aggregated. Constraining the LoRA subspace with OSRM using peer-task data raises this to $75.01\%$. Replacing the peer-task constraint with a task-agnostic WikiText-103 proxy closes most of this gap: (O) reaches $74.48\%$ and (J) reaches $74.18\%$ (Figure~\ref{fig:radar_glue}). The pattern repeats under TIES-Merging, where (O) and (J) lift the standard-LoRA baseline from $59.46\%$ to $67.88\%$ and $70.24\%$, respectively. Instantiation (J) exceeds the $67.66\%$ attained by OSRM, and (O) matches it. Since the proxy contains no examples from any GLUE task, these gains are consistent with the interference-suppression mechanism of Sections~\ref{sec:framework}--\ref{sec:proxy_regularization} extending from vision transformers to LoRA-adapted LLMs: a broad task-agnostic text distribution already provides enough gradient coverage to reduce the interference on unseen targets.

\textbf{Computational efficiency.} As shown in Table~\ref{tab:llm_glue_efficiency}, Instantiation (O) needs only one additional forward-backward through the proxy mini-batch and materializes no Jacobian, leaving peak memory unchanged from standard LoRA ($14.57$~GiB) at $1.60\times$ the per-step time. Instantiation (J) requires forward-mode differentiation, raising peak memory to $16.73$~GiB and per-step time to $2.85\times$. Thus (O) matches the merged accuracy of (J) under Task Arithmetic and recovers most of its gain under TIES-Merging at a markedly lower cost, matching the trade-off of Table~\ref{tab:three_realizations}.

\begin{table}[t]\small
\centering
\caption{Per-step cost of instantiations (O) and (J) on Llama-3-8B with rank-$8$ LoRA, measured on SST-2 under an effective task batch of $64$ and a proxy batch of one. Multipliers in parentheses are relative to standard LoRA.}
\label{tab:llm_glue_efficiency}
\setlength{\tabcolsep}{5pt}
\begin{tabular}{lcc}
\toprule
Method & Sec./step & Peak alloc. (GiB) \\
\midrule
Standard LoRA        & 2.461 & 14.57 \\
Standard LoRA $+(O)$ & 3.938 ($1.60\times$) & 14.57 ($1.00\times$) \\
Standard LoRA $+(J)$ & 7.021 ($2.85\times$) & 16.73 ($1.15\times$) \\
\bottomrule
\end{tabular}
\end{table}
\section{Conclusion}
In this work, we addressed the challenge of parameter interference in task arithmetic by introducing a Spectral Framework for Universal Proxy Regularization. 
We theoretically characterized the proxy distributions through the $(\alpha, \beta)$-relative spectral bound, proving that a universal proxy with smaller $(\alpha, \beta)$ effectively suppresses interference (Theorem~\ref{theorem:unified_bound}). 
Within this framework, we studied three instantiations of the same spectral objective, Jacobian~(J), frozen Fisher~(F), and output-drift~(O), proved their equivalence under NTK linearization (Proposition~\ref{prop:linearized_equivalence}), and established a 
three-way hierarchy of their regularizer-specific gaps beyond linearization (Theorems~\ref{theorem:gap_fisher} and \ref{theorem:gap_output}): zero for (J), proxy-independent for (F), and proxy-quality-scaling for (O). Empirically, using ImageNet as a universal proxy approaches oracle performance on CLIP across vision benchmarks, and a task-agnostic WikiText-103 proxy improves merged performance on large language models. Future work will explore actively constructing proxies that optimize the $(\alpha,\beta)$ criterion and extend the spectral analysis to broader model composition paradigms such as continual learning and cross-modal merging.

\section*{Acknowledgments}
This work is supported in part by the National Natural Science Foundation of China (62576160, 62192783), and the Australian Research Council’s Discovery Project (DP220101784).

\bibliographystyle{IEEEtran}
\bibliography{reference.bib}

\newpage
\onecolumn
\begin{center}
    {\LARGE \bfseries A Spectral Framework for Universal Proxy 

        \vspace{0.5em}
    Regularization in Model Merging}
    
    \vspace{1em}
    {\large Supplementary Material}
\end{center}

\section{Theoretical Perspectives on Proxy Distribution}\label{sec:Theoretical Perspectives on Proxy Distribution}
In Section~IV-C of the main text, we introduced the $(\alpha, \beta)$-relative spectral bound as a unified framework to characterize the relationship between a proxy distribution $\mathcal{Q}$ and a target task distribution $\mathcal{P}_t$. In this appendix, we provide the detailed theoretical motivations underpinning this framework. We analyze the interference suppression mechanism from two complementary perspectives: Spectral Coverage, which requires the proxy to span the gradient subspace of the target, and Perturbation Tolerance, which analyzes the stability under spectral deviations.

\subsection{Spectral Coverage}
The first perspective focuses on the ideal scenario where the proxy distribution's gradient space fully covers that of the target. This ensures that minimizing the proxy regularizer strictly bounds the interference.

\begin{theorem}[Spectral coverage]\label{theorem:spectral_coverage}
Let $\mathcal{Q}$ be a proxy distribution and $\mathcal{P}_t$ be the distribution of task $t$. If there exists a constant $\kappa_t \geq 0$ such that $A_{\mathcal{P}_t} \preceq \kappa_t A_{\mathcal{Q}}$, then for any $\tau \in \mathbb{R}^p$:
\begin{equation}
\mathcal{I}(\tau | \mathcal{P}_t) \leq \kappa_t R_{\mathcal{Q}}(\tau).
\end{equation}
\end{theorem}

\begin{proof}
See Appendix~\ref{subsec:proof_theorem:spectral_coverage}.
\end{proof}

The coverage constant $\kappa_t$ measures how well $\mathcal{Q}$ spans the gradient directions relevant to task $t$. Smaller $\kappa_t$ indicates better coverage, with $\kappa_t = 1$ representing perfect alignment of gradient distributions. This perspective motivates the $\alpha$ parameter in our unified bound.

\subsection{Perturbation Tolerance}
In practice, perfect spectral coverage is often unattainable. The second perspective relaxes this requirement by allowing controlled deviations between the covariance operators.

\begin{theorem}[Perturbation tolerance]\label{theorem:perturbation}
Define the deviation matrix $\Delta_t := A_{\mathcal{P}_t} - A_{\mathcal{Q}}$ and its numerical radius $r(\Delta_t) := \sup_{\|\tau\|=1} |\tau^{\top}\Delta_t\tau|.$ If $r(\Delta_t) \leq \varepsilon_t$ for some $\varepsilon_t \geq 0$, then:
\begin{equation}
    \mathcal{I}(\tau | \mathcal{P}_t) \leq R_{\mathcal{Q}}(\tau) + \varepsilon_t\|\tau\|^2.
\end{equation}
\end{theorem}

\begin{proof}
    See Appendix~\ref{subsec:proof_perturbation}.
\end{proof}

This theorem suggests that even when $\mathcal{Q}$ does not perfectly cover $\mathcal{P}_t$, the proxy regularizer remains effective as long as the spectral discrepancy (measured by the numerical radius) is bounded. The residual interference $\varepsilon_t \|\tau\|^2$ can be controlled via weight decay. This perspective motivates the $\beta$ parameter (leakage) in our unified bound.

\subsection{Connection to the Unified Bound}
The $(\alpha, \beta)$-relative spectral bound presented in the main text unifies these two perspectives. $\alpha_t$ generalizes the coverage constant $\kappa_t$, while $\beta_t$ captures the leakage energy corresponding to the perturbation magnitude $\varepsilon_t$. More precisely, both perspectives lie on the lower boundary of the admissible set $\mathcal{F}_t$ of Remark~\ref{rem:coverage}. As $\rho\to 0$, the regularized coverage coefficient $\alpha_t(\rho)$ increases to the smallest coverage constant $\kappa_t$ of Theorem~\ref{theorem:spectral_coverage}, with $\kappa_t=\infty$ if the range of $A_{\mathcal{P}_t}$ is not contained in that of $A_{\mathcal{Q}}$ (Lemma~\ref{lem:coverage}(c)). At $\alpha=1$, the smallest admissible $\beta$ is $\max\{\lambda_{\max}(A_{\mathcal{P}_t}-A_{\mathcal{Q}}),0\}\le r(\Delta_t)\le\varepsilon_t$, so the condition of Theorem~\ref{theorem:perturbation} yields the admissible pair $(1,\varepsilon_t)$. The coverage curve $\alpha_t(\rho)$ interpolates between these two regimes, and by Corollary~\ref{cor:coverage} it carries all the proxy-dependent information used by Theorem~\ref{theorem:unified_bound}.
\section{Detailed Proofs}
\subsection{Proof of Lemma~\ref{lemma:disentanglement2tauj}}\label{subsec:proof_lemma1}
\begin{quote}
\textbf{Lemma~\ref{lemma:disentanglement2tauj}.}
\textit{Under Assumption 1 (NTK linearization), weight disentanglement (Definition~1) holds if and only if, for any pair of distinct tasks $s, t \in \mathcal{T}$,}
\begin{equation}
    J(x_t; \theta_0)^{\top}\tau_s = 0, \quad \forall s \neq t, \, x_t \sim \mathcal{P}_t.
\end{equation}
\end{quote}

\begin{proof}
For clarity, we consider the case of two tasks $s$ and $t$ with task vectors $\tau_s$ and $\tau_t$. The generalization to multiple tasks follows by induction.

\textbf{($\Rightarrow$) Necessity.}

Assume weight disentanglement holds. Consider any two distinct tasks $s, t \in \mathcal{T}$ with $s \neq t$, and any input $x_t$ from task $t$'s distribution $\mathcal{P}_t$. 

By the definition of weight disentanglement with $\mathcal{T} = \{s, t\}$, we have,
\begin{equation}
    f(x_t; \theta_0 + \alpha_s\tau_s + \alpha_t\tau_t) = f(x_t; \theta_0 + \alpha_t\tau_t), \quad \forall \alpha_s, \alpha_t \in \mathbb{R}.
\end{equation}

Applying the NTK linearization assumption to both sides,
\begin{equation}
    f(x_t; \theta_0) + J(x_t)^{\top}(\alpha_s\tau_s + \alpha_t\tau_t) = f(x_t; \theta_0) + J(x_t)^{\top}(\alpha_t\tau_t).
\end{equation}

Simplifying
\begin{equation}
    \alpha_s J(x_t)^{\top}\tau_s + \alpha_t J(x_t)^{\top}\tau_t = \alpha_t J(x_t)^{\top}\tau_t.
\end{equation}

This yields
\begin{equation}
    \alpha_s J(x_t)^{\top}\tau_s = 0, \quad \forall \alpha_s \in \mathbb{R}.
\end{equation}

Since this must hold for arbitrary $\alpha_s$, we conclude,
\begin{equation}
    J(x_t)^{\top}\tau_s = 0.
\end{equation}

By symmetry, for input $x_s$ from $\mathcal{P}_s$ we can get $J(x_s)^{\top}\tau_t = 0$ for all $x_s \sim \mathcal{P}_s$. Since $s, t$ were arbitrary distinct tasks, the condition holds for all pairs.

\textbf{($\Leftarrow$) Sufficiency.}

Conversely, assume the condition $J(x_t)^{\top}\tau_s = 0$ holds for all $s \neq t$ and $x_t \sim \mathcal{P}_t$.

For any input $x_t$ from task $t$ and any collection of coefficients $\{\alpha_s\}_{s \in \mathcal{T}}$, the NTK linearization assumption gives,
\begin{equation}
    f\left(x_t; \theta_0 + \sum_{s \in \mathcal{T}} \alpha_s\tau_s\right) \approx f(x_t; \theta_0) + \sum_{s \in \mathcal{T}} \alpha_s J(x_t)^{\top}\tau_s.
\end{equation}

Since $J(x_t)^{\top}\tau_s = 0$ for all $s \neq t$,
\begin{equation}
    \sum_{s \in \mathcal{T}} \alpha_s J(x_t)^{\top}\tau_s = \alpha_t J(x_t)^{\top}\tau_t.
\end{equation}

Therefore,
\begin{equation}
    f\left(x_t; \theta_0 + \sum_{s \in \mathcal{T}} \alpha_s\tau_s\right) \approx f(x_t; \theta_0) + \alpha_t J(x_t)^{\top}\tau_t = f(x_t; \theta_0 + \alpha_t\tau_t).
\end{equation}

Thus, the condition is sufficient for achieving weight disentanglement.
\end{proof}

\subsection{Proof of Proposition~\ref{proposition:quadratic}}\label{subsec:proof_proposition:quadratic}
\begin{quote}
\textbf{Proposition~\ref{proposition:quadratic}.}
\textit{The interference admits the following quadratic form}
\begin{equation}
   \mathcal{I}(\tau | \mathcal{P}) = \tau^{\top}A_{\mathcal{P}}\tau.
\end{equation}
\end{quote}

\begin{proof}
By Definition~\ref{definition:distribution_interference} and expanding the squared norm,
\begin{equation}
    \mathcal{I}(\tau | \mathcal{P}) = \mathbb{E}_{x \sim \mathcal{P}}\left[\|J(x)^{\top}\tau\|^2\right] = \mathbb{E}_{x \sim \mathcal{P}}\left[\tau^{\top}J(x)J(x)^{\top}\tau\right].
\end{equation}

By linearity of expectation and Definition~\ref{definition:jacobian_cov} ($A_{\mathcal{P}} = \mathbb{E}_{x \sim \mathcal{P}}[J(x)J(x)^{\top}]$),
\begin{equation}
    \mathcal{I}(\tau | \mathcal{P}) = \tau^{\top}\mathbb{E}_{x \sim \mathcal{P}}\left[J(x)J(x)^{\top}\right]\tau = \tau^{\top}A_{\mathcal{P}}\tau.
\end{equation}
\end{proof}

\subsection{Proof of Theorem~\ref{theorem:spectral_coverage}}\label{subsec:proof_theorem:spectral_coverage}
\begin{quote}
\textbf{Theorem~\ref{theorem:spectral_coverage}} (Spectral coverage). 
\textit{Let $\mathcal{Q}$ be a proxy distribution and $\mathcal{P}_t$ be the distribution of task $t$. If there exists a constant $\kappa_t \geq 0$ such that $A_{\mathcal{P}_t} \preceq \kappa_t A_{\mathcal{Q}}$, then for any $\tau \in \mathbb{R}^p$:}
\begin{equation}
   \mathcal{I}(\tau | \mathcal{P}_t) \leq \kappa_t R_{\mathcal{Q}}(\tau).
\end{equation}
\end{quote}

\begin{proof}
Recall from Proposition~\ref{proposition:quadratic} that the interference on task $t$ is given by $\mathcal{I}(\tau | \mathcal{P}_t) = \tau^{\top}A_{\mathcal{P}_t}\tau$, and the proxy regularization term is $R_{\mathcal{Q}}(\tau) = \tau^{\top}A_{\mathcal{Q}}\tau$.

The condition $A_{\mathcal{P}_t} \preceq \kappa_t A_{\mathcal{Q}}$ implies that the matrix $\kappa_t A_{\mathcal{Q}} - A_{\mathcal{P}_t}$ is positive semi-definite. By definition of positive semi-definiteness, for any vector $\tau \in \mathbb{R}^p$, we have,
\begin{equation}
\tau^{\top}(\kappa_t A_{\mathcal{Q}} - A_{\mathcal{P}_t})\tau \geq 0.
\end{equation}

Rearranging the terms yields,
\begin{equation}
\tau^{\top} A_{\mathcal{P}_t}\tau \leq \kappa_t \tau^{\top}A_{\mathcal{Q}}\tau.
\end{equation}

Substituting the definitions of interference and the proxy regularizer, we obtain:
\begin{equation}
\mathcal{I}(\tau | \mathcal{P}_t) \leq \kappa_t R_{\mathcal{Q}}(\tau).
\end{equation}
\end{proof}

\subsection{Proof of Theorem~\ref{theorem:perturbation}}\label{subsec:proof_perturbation}
\begin{quote}
\textbf{Theorem~\ref{theorem:perturbation}} (Perturbation tolerance). 
\textit{Define the deviation matrix $\Delta_t := A_{\mathcal{P}_t} - A_{\mathcal{Q}}$ and its numerical radius $r(\Delta_t) := \sup_{\|\tau\|=1} |\tau^{\top}\Delta_t\tau|.$ If $r(\Delta_t) \leq \varepsilon_t$ for some $\varepsilon_t \geq 0$, then:}
\begin{equation}
   \mathcal{I}(\tau | \mathcal{P}_t) \leq R_{\mathcal{Q}}(\tau) + \varepsilon_t\|\tau\|^2.
\end{equation}
\end{quote}

\begin{proof}
Recall that $\mathcal{I}(\tau | \mathcal{P}_t) = \tau^{\top}A_{\mathcal{P}_t}\tau$ and $R{\mathcal{Q}}(\tau) = \tau^{\top}A_{\mathcal{Q}}\tau$.
By the definition of the deviation matrix $\Delta_t = A_{\mathcal{P}_t} - A_{\mathcal{Q}}$, we can rewrite the interference term as,
\begin{equation}
\mathcal{I}(\tau | \mathcal{P}_t) = \tau^{\top}(A_{\mathcal{Q}} + \Delta_t)\tau = \tau^{\top}A_{\mathcal{Q}}\tau + \tau^{\top}\Delta_t\tau.
\end{equation}

Substituting $R_{\mathcal{Q}}(\tau)$, we have,
\begin{equation}
\mathcal{I}(\tau | \mathcal{P}_t) = R_{\mathcal{Q}}(\tau) + \tau^{\top}\Delta_t\tau.
\end{equation}

By the definition of the numerical radius $r(\Delta_t) = \sup_{\|\tau\|=1} |\tau^{\top}\Delta_t\tau|$, for any vector $\tau$, the following inequality holds,
\begin{equation}
|\tau^{\top}\Delta_t\tau| \leq r(\Delta_t)\|\tau\|^2.
\end{equation}

Given the condition $r(\Delta_t) \leq \varepsilon_t$, we have:
\begin{equation}
\tau^{\top}\Delta_t\tau \leq |\tau^{\top}\Delta_t\tau| \leq \varepsilon_t\|\tau\|^2.
\end{equation}

Therefore:
\begin{equation}
\mathcal{I}(\tau | \mathcal{P}_t) \leq R_{\mathcal{Q}}(\tau) + \varepsilon_t\|\tau\|^2.
\end{equation}
\end{proof}

\subsection{Proof of Lemma~\ref{lemma:tau_solution}}\label{subsec:proof_lemma:tau_solution}
\begin{quote}
\textbf{Lemma~\ref{lemma:tau_solution}.} 
\textit{Under Assumption~\ref{assumption:NTK linearization} (NTK linearization), the optimal task vector $\tau^*_s$ minimizing Eq.~\eqref{formula:objective} is given by,}
\begin{equation}
\tau_s^* = (A_{\mathcal{P}s} + \lambda A_{\mathcal{Q}} + \mu I)^{-1}b_s,
\end{equation}
where $b_s := \mathbb{E}_{(x,y) \sim \mathcal{D}_s}[J(x)(y - f(x;\theta_0))]$.
\end{quote}

\begin{proof}
Under the NTK linearization (Assumption~\ref{assumption:NTK linearization}), we have $f(x; \theta_0 + \tau) \approx f(x; \theta_0) + J(x)^{\top}\tau$. Define the residual $r(x) := y - f(x; \theta_0)$. The objective function can be rewritten as,
\begin{equation}
\mathcal{L}(\tau) = \frac{1}{2}\mathbb{E}_{(x,y) \sim \mathcal{D}_s}\left[\left(r(x) - J(x)^{\top}\tau\right)^2\right] + \frac{\lambda}{2} \tau^{\top}A_{\mathcal{Q}}\tau + \frac{\mu}{2}\|\tau\|^2.
\end{equation}

Expanding the squared term
\begin{align}
\mathcal{L}(\tau) &= \frac{1}{2}\mathbb{E}\left[r(x)^2 - 2r(x)J(x)^{\top}\tau + \tau^{\top}J(x)J(x)^{\top}\tau\right] + \frac{\lambda}{2} \tau^{\top}A_{\mathcal{Q}}\tau + \frac{\mu}{2}\|\tau\|^2 \\ \nonumber
&= \frac{1}{2}\mathbb{E}[r(x)^2] - \tau^{\top}\mathbb{E}[J(x)r(x)] + \frac{1}{2}\tau^{\top}\mathbb{E}[J(x)J(x)^{\top}]\tau + \frac{\lambda}{2} \tau^{\top}A_{\mathcal{Q}}\tau + \frac{\mu}{2}\|\tau\|^2.
\end{align}

By Definition~\ref{definition:jacobian_cov}, $A_{\mathcal{P}s} = \mathbb{E}_{x \sim \mathcal{D}_s}[J(x)J(x)^{\top}]$, and let $b_s = \mathbb{E}_{(x,y) \sim \mathcal{D}_s}[J(x)r(x)]$. Thus,
\begin{equation}
\mathcal{L}(\tau) = \text{const.} - \tau^{\top}b_s + \frac{1}{2}\tau^{\top}A_{\mathcal{P}_s}\tau + \frac{\lambda}{2} \tau^{\top}A_{\mathcal{Q}}\tau + \frac{\mu}{2}\|\tau\|^2.
\end{equation}

Taking the gradient with respect to $\tau$ and setting it to zero,
\begin{equation}
\nabla_{\tau}\mathcal{L}(\tau) = -b_s + A_{\mathcal{P}_s}\tau + \lambda A_{\mathcal{Q}}\tau + \mu\tau = 0.
\end{equation}

Rearranging,
\begin{equation}
(A_{\mathcal{P}s} + \lambda A_{\mathcal{Q}} + \mu I)\tau = b_s.
\end{equation}

Solving for $\tau$ yields,
\begin{equation}
\tau_s^* = (A_{\mathcal{P}_s} + \lambda A_{\mathcal{Q}} + \mu I)^{-1}b_s.
\end{equation}
\end{proof}

\subsection{Proof of Theorem~\ref{theorem:unified_bound} and Corollary~\ref{cor:coverage}}\label{subsec:proof_the_unified_bound}

\begin{quote}
\textbf{Theorem~\ref{theorem:unified_bound}} (Unified interference bound).
\textit{Let $\tau_s^*$ be the regularized solution for task $s$ as in Lemma~\ref{lemma:tau_solution}. Suppose task $t \neq s$ satisfies the $(\alpha_t, \beta_t)$-relative spectral bound. Then the interference of $\tau_s^*$ on task $t$ is bounded by,}
\begin{equation}
    \mathcal{I}(\tau_s^* | \mathcal{P}_t) \leq \left(\frac{\alpha_t}{4\lambda\mu} + \frac{\beta_t}{\mu^2}\right)\|b_s\|^2.
\end{equation}
\end{quote}

\begin{proof}
From Proposition~\ref{proposition:quadratic}, the interference on task $t$ is given by,
\begin{equation}
\mathcal{I}(\tau_s^* | \mathcal{P}_t) = (\tau_s^*)^{\top}A_{\mathcal{P}_t}\tau_s^*.
\end{equation}

By the $(\alpha_t, \beta_t)$-relative spectral bound (Definition~\ref{definition:relative_spectral_bound}), we have $A_{\mathcal{P}_t} \preceq \alpha_t A_{\mathcal{Q}} + \beta_t I$. This implies,
\begin{equation}
(\tau_s^*)^{\top}A_{\mathcal{P}_t}\tau_s^* \leq \alpha_t (\tau_s^*)^{\top}A_{\mathcal{Q}}\tau_s^* + \beta_t \|\tau_s^*\|^2.
\end{equation}

We now bound each term separately.

\textbf{Step 1: Bounding $(\tau_s^*)^{\top}A_{\mathcal{Q}}\tau_s^*$.}

From Lemma~\ref{lemma:tau_solution}, we have $\tau_s^* = (A_{\mathcal{P}_s} + \lambda A_{\mathcal{Q}} + \mu I)^{-1}b_s$, \textit{i.e.}, $\tau_s^*$ satisfies the first-order optimality condition,
\begin{equation}
(A_{\mathcal{P}_s} + \lambda A_{\mathcal{Q}} + \mu I)\tau_s^* = b_s.
\end{equation}

Left-multiplying both sides by $(\tau_s^*)^{\top}$ gives,
\begin{equation}
(\tau_s^*)^{\top}A_{\mathcal{P}_s}\tau_s^* + \lambda (\tau_s^*)^{\top}A_{\mathcal{Q}}\tau_s^* + \mu\|\tau_s^*\|^2 = b_s^{\top}\tau_s^*.
\end{equation}

By the Cauchy--Schwarz inequality, the right-hand side satisfies,
\begin{equation}
b_s^{\top}\tau_s^* \leq \|b_s\|\,\|\tau_s^*\|.
\end{equation}

Since $A_{\mathcal{P}_s} \succeq 0$ (positive semi-definite), we have $(\tau_s^*)^{\top}A_{\mathcal{P}_s}\tau_s^* \geq 0$. Dropping this term and combining the two relations above,
\begin{equation}\label{eq:energy_ineq}
\lambda (\tau_s^*)^{\top}A_{\mathcal{Q}}\tau_s^* + \mu\|\tau_s^*\|^2 \leq \|b_s\|\,\|\tau_s^*\|.
\end{equation}

Rearranging,
\begin{equation}
\lambda (\tau_s^*)^{\top}A_{\mathcal{Q}}\tau_s^* \leq \|b_s\|\,\|\tau_s^*\| - \mu\|\tau_s^*\|^2.
\end{equation}

The right-hand side depends on $\tau_s^*$ only through its norm $z = \|\tau_s^*\|$. Define the function $g: [0, \infty) \to \mathbb{R}$ by,
\begin{equation}
g(z) = \|b_s\|\,z - \mu z^2.
\end{equation}

To find the maximum of $g(z)$, we compute its derivative,
\begin{equation}
g'(z) = \|b_s\| - 2\mu z.
\end{equation}

Setting $g'(z) = 0$, we find that the critical point occurs at,
\begin{equation}
z^* = \frac{\|b_s\|}{2\mu}.
\end{equation}

Since $g'(z) > 0$ for $z < z^*$ and $g'(z) < 0$ for $z > z^*$, this is a global maximum. Evaluating $g$ at $z^*$,
\begin{equation}
g\left(\frac{\|b_s\|}{2\mu}\right) = \frac{\|b_s\|^2}{2\mu} - \mu \cdot \frac{\|b_s\|^2}{4\mu^2} = \frac{\|b_s\|^2}{4\mu}.
\end{equation}

Therefore, $\lambda (\tau_s^*)^{\top}A_{\mathcal{Q}}\tau_s^* \leq g(\|\tau_s^*\|) \leq \frac{\|b_s\|^2}{4\mu}$. Dividing both sides by $\lambda > 0$,
\begin{equation}\label{eq:aq_bound}
(\tau_s^*)^{\top}A_{\mathcal{Q}}\tau_s^* \leq \frac{\|b_s\|^2}{4\lambda\mu}.
\end{equation}

\textbf{Step 2: Bounding $\|\tau_s^*\|^2$.}

We return to inequality~\eqref{eq:energy_ineq}. Since $A_{\mathcal{Q}} \succeq 0$, we have $\lambda(\tau_s^*)^{\top}A_{\mathcal{Q}}\tau_s^* \geq 0$, and dropping this term gives,
\begin{equation}
\mu\|\tau_s^*\|^2 \leq \|b_s\|\,\|\tau_s^*\|.
\end{equation}

If $\tau_s^* = 0$, the desired bound holds trivially. Otherwise, dividing both sides by $\mu\|\tau_s^*\| > 0$,
\begin{equation}
\|\tau_s^*\| \leq \frac{\|b_s\|}{\mu}.
\end{equation}

Therefore,
\begin{equation}\label{eq:norm_bound}
\|\tau_s^*\|^2 \leq \frac{\|b_s\|^2}{\mu^2}.
\end{equation}

\textbf{Step 3: Combining the bounds.}

Substituting inequalities~\eqref{eq:aq_bound} and~\eqref{eq:norm_bound} into the initial decomposition,
\begin{align}
\mathcal{I}(\tau_s^* | \mathcal{P}_t) &\leq \alpha_t (\tau_s^*)^{\top}A_{\mathcal{Q}}\tau_s^* + \beta_t \|\tau_s^*\|^2 \\ \nonumber
&\leq \alpha_t \cdot \frac{\|b_s\|^2}{4\lambda\mu} + \beta_t \cdot \frac{\|b_s\|^2}{\mu^2} \\ \nonumber
&= \left(\frac{\alpha_t}{4\lambda\mu} + \frac{\beta_t}{\mu^2}\right)\|b_s\|^2.
\end{align}

This completes the proof.
\end{proof}

We next prove Corollary~\ref{cor:coverage}. Throughout, let $\mathcal{F}_t:=\{(\alpha,\beta)\geq 0:\ A_{\mathcal{P}_t}\preceq\alpha A_{\mathcal{Q}}+\beta I\}$ be the admissible set of Remark~\ref{rem:coverage}, and let $\beta_t^\star(\alpha):=\max\{\lambda_{\max}(A_{\mathcal{P}_t}-\alpha A_{\mathcal{Q}}),0\}$ for $\alpha\geq 0$ denote its lower boundary, \textit{i.e.}, the smallest $\beta$ with $(\alpha,\beta)\in\mathcal{F}_t$. We assume $A_{\mathcal{P}_t}\neq 0$; otherwise $\alpha_t(\rho)=0$ for every $\rho>0$ and all statements below are trivial. We first collect the properties of the coverage curve $\rho\mapsto\alpha_t(\rho)$ on which the proof relies.

\begin{lemma}[Coverage curve]\label{lem:coverage}
For every $\rho>0$, the following hold.
\begin{itemize}
\item[(a)] $(\alpha_t(\rho),\beta_t(\rho))\in\mathcal{F}_t$ and $\beta_t^\star(\alpha_t(\rho))=\beta_t(\rho)$, \textit{i.e.}, the pair lies on the lower boundary of $\mathcal{F}_t$.
\item[(b)] $\alpha_t(\rho)$ is non-increasing and $\beta_t(\rho)$ is non-decreasing in $\rho$, with $\beta_t(\rho)\leq\lambda_{\max}(A_{\mathcal{P}_t})$.
\item[(c)] $(\alpha_t(\rho),\beta_t(\rho))\to(0,\lambda_{\max}(A_{\mathcal{P}_t}))$ as $\rho\to\infty$, and $\alpha_t(\rho)\uparrow\kappa_t$ as $\rho\to 0$, where $\kappa_t\in(0,\infty]$ is the smallest coverage constant in Theorem~\ref{theorem:spectral_coverage} ($\kappa_t=\infty$ if no such constant exists). If $\kappa_t<\infty$, then $\beta_t(\rho)\to 0$ as $\rho\to 0$.
\item[(d)] Every boundary point $(\alpha,\beta_t^\star(\alpha))$ with $\alpha>0$ and $\beta_t^\star(\alpha)>0$ equals $(\alpha_t(\rho),\beta_t(\rho))$ with $\rho=\beta_t^\star(\alpha)/\alpha$.
\end{itemize}
Hence, as $\rho$ ranges over $(0,\infty)$, $(\alpha_t(\rho),\beta_t(\rho))$ traces the lower boundary $\{(\alpha,\beta_t^\star(\alpha)):0<\alpha<\kappa_t\}$, whose end points $(0,\lambda_{\max}(A_{\mathcal{P}_t}))$ and, if $\kappa_t<\infty$, $(\kappa_t,0)$ are the limits $\rho\to\infty$ and $\rho\to 0$. Moreover, the maximizer in the definition of $\alpha_t(\rho)$ is a generalized eigenvector of $(A_{\mathcal{P}_t},A_{\mathcal{Q}}+\rho I)$ and in general differs from the principal eigenvector of $A_{\mathcal{P}_t}$.
\end{lemma}

\begin{proof}
(a) For $\alpha\geq 0$, $A_{\mathcal{P}_t}\preceq\alpha(A_{\mathcal{Q}}+\rho I)$ holds if and only if $A_{\mathcal{P}_t}-\alpha A_{\mathcal{Q}}\preceq\rho\alpha I$, \textit{i.e.}, if and only if $g(\alpha):=\lambda_{\max}(A_{\mathcal{P}_t}-\alpha A_{\mathcal{Q}})-\rho\alpha\leq 0$. Since $A_{\mathcal{Q}}\succeq 0$, $\lambda_{\max}(A_{\mathcal{P}_t}-\alpha A_{\mathcal{Q}})$ is continuous and non-increasing in $\alpha$, so $g$ is continuous and strictly decreasing, with $g(0)=\lambda_{\max}(A_{\mathcal{P}_t})>0$ and $g(\alpha)\to-\infty$ as $\alpha\to\infty$. As $\alpha_t(\rho)$ is the smallest $\alpha$ with $A_{\mathcal{P}_t}\preceq\alpha(A_{\mathcal{Q}}+\rho I)$, we have $g(\alpha_t(\rho))=0$, \textit{i.e.}, $\lambda_{\max}(A_{\mathcal{P}_t}-\alpha_t(\rho)A_{\mathcal{Q}})=\rho\,\alpha_t(\rho)>0$. This gives both $(\alpha_t(\rho),\beta_t(\rho))\in\mathcal{F}_t$ and $\beta_t^\star(\alpha_t(\rho))=\beta_t(\rho)$.

(b) For $0<\rho_1<\rho_2$, we have $A_{\mathcal{Q}}+\rho_1 I\preceq A_{\mathcal{Q}}+\rho_2 I\preceq\frac{\rho_2}{\rho_1}(A_{\mathcal{Q}}+\rho_1 I)$, hence
\begin{equation}
\alpha_t(\rho_2)\leq\alpha_t(\rho_1)\leq\frac{\rho_2}{\rho_1}\,\alpha_t(\rho_2).
\end{equation}
The first inequality shows that $\alpha_t$ is non-increasing; multiplying the second by $\rho_1$ gives $\beta_t(\rho_1)\leq\beta_t(\rho_2)$. Moreover, $A_{\mathcal{P}_t}\preceq\lambda_{\max}(A_{\mathcal{P}_t})I\preceq\frac{\lambda_{\max}(A_{\mathcal{P}_t})}{\rho}(A_{\mathcal{Q}}+\rho I)$, so $\alpha_t(\rho)\leq\lambda_{\max}(A_{\mathcal{P}_t})/\rho$, \textit{i.e.}, $\beta_t(\rho)\leq\lambda_{\max}(A_{\mathcal{P}_t})$.

(c) We can write $\beta_t(\rho)=\max_{v\neq 0}v^\top A_{\mathcal{P}_t}v/(\rho^{-1}v^\top A_{\mathcal{Q}}v+\|v\|^2)$. Taking $v=u$, the unit principal eigenvector of $A_{\mathcal{P}_t}$, gives $\beta_t(\rho)\geq\lambda_{\max}(A_{\mathcal{P}_t})/(1+\rho^{-1}u^\top A_{\mathcal{Q}}u)\to\lambda_{\max}(A_{\mathcal{P}_t})$ as $\rho\to\infty$. Together with (b), $\beta_t(\rho)\to\lambda_{\max}(A_{\mathcal{P}_t})$ and $\alpha_t(\rho)=\beta_t(\rho)/\rho\to 0$. As $\rho\to 0$, $\alpha_t(\rho)$ is non-decreasing by (b) and therefore converges to $\sup_{\rho>0}\alpha_t(\rho)$. If $A_{\mathcal{P}_t}\preceq\kappa A_{\mathcal{Q}}$, then $A_{\mathcal{P}_t}\preceq\kappa(A_{\mathcal{Q}}+\rho I)$ and $\alpha_t(\rho)\leq\kappa$ for all $\rho>0$, so $\sup_{\rho>0}\alpha_t(\rho)\leq\kappa_t$. Conversely, if $\alpha_t(\rho)\leq a$ for all $\rho>0$, then $v^\top A_{\mathcal{P}_t}v\leq a\,v^\top A_{\mathcal{Q}}v+a\rho\|v\|^2$ for all $v$ and $\rho>0$; letting $\rho\to 0$ gives $A_{\mathcal{P}_t}\preceq aA_{\mathcal{Q}}$, so $\kappa_t\leq a$. Hence $\sup_{\rho>0}\alpha_t(\rho)=\kappa_t$, and $\kappa_t>0$ because $A_{\mathcal{P}_t}\neq 0$. If $\kappa_t<\infty$, then $\beta_t(\rho)=\rho\,\alpha_t(\rho)\leq\rho\,\kappa_t\to 0$.

(d) Let $\alpha>0$ with $\beta:=\beta_t^\star(\alpha)>0$, and set $\rho:=\beta/\alpha$. Since $(\alpha,\beta)\in\mathcal{F}_t$, we have $A_{\mathcal{P}_t}\preceq\alpha(A_{\mathcal{Q}}+\rho I)$, so $\alpha\geq\alpha_t(\rho)$. If $\alpha>\alpha_t(\rho)$, then $A_{\mathcal{P}_t}\preceq\alpha_t(\rho)A_{\mathcal{Q}}+\rho\,\alpha_t(\rho)I\preceq\alpha A_{\mathcal{Q}}+\rho\,\alpha_t(\rho)I$, so $\beta_t^\star(\alpha)\leq\rho\,\alpha_t(\rho)<\rho\alpha=\beta$, a contradiction. Hence $\alpha=\alpha_t(\rho)$ and $\beta=\rho\,\alpha_t(\rho)=\beta_t(\rho)$.

For the concluding statement, note that $\beta_t^\star(\alpha)=0$ if and only if $A_{\mathcal{P}_t}\preceq\alpha A_{\mathcal{Q}}$, \textit{i.e.}, if and only if $\alpha\geq\kappa_t$. By (a), every $\rho>0$ yields a boundary point with $\alpha_t(\rho)>0$ and $\beta_t^\star(\alpha_t(\rho))=\beta_t(\rho)>0$, hence with $0<\alpha_t(\rho)<\kappa_t$; by (d), every boundary point with $0<\alpha<\kappa_t$ is obtained in this way; and the end points follow from (c). Finally, $\alpha_t(\rho)$ is the largest generalized Rayleigh quotient of $(A_{\mathcal{P}_t},A_{\mathcal{Q}}+\rho I)$ with $A_{\mathcal{Q}}+\rho I\succ 0$, so its maximizer $v$ satisfies $A_{\mathcal{P}_t}v=\alpha_t(\rho)(A_{\mathcal{Q}}+\rho I)v$. If $v$ were also the principal eigenvector of $A_{\mathcal{P}_t}$, this relation would force $v$ to be an eigenvector of $A_{\mathcal{Q}}$, which does not hold in general.
\end{proof}

We now prove Corollary~\ref{cor:coverage}.
\begin{quote}
\textbf{Corollary~\ref{cor:coverage}} (Regularized form of Theorem~\ref{theorem:unified_bound}).
\textit{Under the conditions of Theorem~\ref{theorem:unified_bound}, for every $\rho>0$,}
\begin{equation}\label{eq:app_coverage_bound}
    \mathcal{I}(\tau_s^* | \mathcal{P}_t) \leq \alpha_t(\rho)\left(\frac{1}{4\lambda\mu} + \frac{\rho}{\mu^2}\right)\|b_s\|^2.
\end{equation}
\textit{(i) The infimum of the right-hand side over $\rho>0$ equals the tightest bound that Theorem~\ref{theorem:unified_bound} yields over all admissible pairs $(\alpha_t,\beta_t)$. (ii) If two proxies satisfy $\alpha_t^{\mathcal{Q}_1}(\rho)\leq\alpha_t^{\mathcal{Q}_2}(\rho)$ for all $\rho>0$, then the bound for $\mathcal{Q}_1$ is no larger than that for $\mathcal{Q}_2$ for every $(\lambda,\mu)$.}
\end{quote}

\begin{proof}
Write $c_1:=1/(4\lambda\mu)$ and $c_2:=1/\mu^2$, so that, for an admissible pair $(\alpha,\beta)\in\mathcal{F}_t$, the bound of Theorem~\ref{theorem:unified_bound} reads $(c_1\alpha+c_2\beta)\|b_s\|^2$. By Lemma~\ref{lem:coverage}(a), $(\alpha_t(\rho),\rho\,\alpha_t(\rho))\in\mathcal{F}_t$, and substituting it into Theorem~\ref{theorem:unified_bound} gives Eq.~\eqref{eq:app_coverage_bound}, since $c_1\alpha_t(\rho)+c_2\rho\,\alpha_t(\rho)=\alpha_t(\rho)(c_1+c_2\rho)$.

(i) Let $B^\star:=\inf_{(\alpha,\beta)\in\mathcal{F}_t}(c_1\alpha+c_2\beta)$ and $B_\rho:=\alpha_t(\rho)(c_1+c_2\rho)$. Since every $(\alpha_t(\rho),\beta_t(\rho))$ is admissible, $\inf_{\rho>0}B_\rho\geq B^\star$. Conversely, $\mathcal{F}_t$ is upward closed, so for fixed $\alpha$ the objective is minimized at $\beta=\beta_t^\star(\alpha)$, and $B^\star=\inf_{\alpha\geq 0}[c_1\alpha+c_2\beta_t^\star(\alpha)]$. We distinguish three cases. If $0<\alpha<\kappa_t$, then $\beta_t^\star(\alpha)>0$, and Lemma~\ref{lem:coverage}(d) gives $c_1\alpha+c_2\beta_t^\star(\alpha)=B_\rho$ with $\rho=\beta_t^\star(\alpha)/\alpha$. If $\alpha=0$, then $\beta_t^\star(0)=\lambda_{\max}(A_{\mathcal{P}_t})$, and $c_2\lambda_{\max}(A_{\mathcal{P}_t})=\lim_{\rho\to\infty}B_\rho$ by Lemma~\ref{lem:coverage}(c). If $\alpha\geq\kappa_t$ (possible only when $\kappa_t<\infty$), then $\beta_t^\star(\alpha)=0$, and $c_1\alpha\geq c_1\kappa_t=\lim_{\rho\to 0}B_\rho$ by Lemma~\ref{lem:coverage}(c). In every case $c_1\alpha+c_2\beta_t^\star(\alpha)\geq\inf_{\rho>0}B_\rho$, hence $B^\star=\inf_{\rho>0}B_\rho$.

(ii) Let $\mathcal{F}_t^{(i)}$ denote the admissible set for $\mathcal{Q}_i$, $i=1,2$. We show $\mathcal{F}_t^{(2)}\subseteq\mathcal{F}_t^{(1)}$. Let $(\alpha,\beta)\in\mathcal{F}_t^{(2)}$. If $\alpha,\beta>0$, set $\rho:=\beta/\alpha$; then $A_{\mathcal{P}_t}\preceq\alpha(A_{\mathcal{Q}_2}+\rho I)$ gives $\alpha\geq\alpha_t^{\mathcal{Q}_2}(\rho)\geq\alpha_t^{\mathcal{Q}_1}(\rho)$, so
\begin{equation}
A_{\mathcal{P}_t}\preceq\alpha_t^{\mathcal{Q}_1}(\rho)(A_{\mathcal{Q}_1}+\rho I)\preceq\alpha A_{\mathcal{Q}_1}+\beta I.
\end{equation}
If $\alpha=0$, the condition $A_{\mathcal{P}_t}\preceq\beta I$ does not involve the proxy. If $\beta=0$, then $A_{\mathcal{P}_t}\preceq\alpha A_{\mathcal{Q}_2}$, and Lemma~\ref{lem:coverage}(c) gives $\alpha\geq\kappa_t^{\mathcal{Q}_2}=\sup_{\rho>0}\alpha_t^{\mathcal{Q}_2}(\rho)\geq\sup_{\rho>0}\alpha_t^{\mathcal{Q}_1}(\rho)=\kappa_t^{\mathcal{Q}_1}$, so $A_{\mathcal{P}_t}\preceq\kappa_t^{\mathcal{Q}_1}A_{\mathcal{Q}_1}\preceq\alpha A_{\mathcal{Q}_1}$. Hence $\mathcal{F}_t^{(2)}\subseteq\mathcal{F}_t^{(1)}$, and the infimum of $c_1\alpha+c_2\beta$ over $\mathcal{F}_t^{(1)}$ is no larger than that over $\mathcal{F}_t^{(2)}$. Since this holds for all $c_1,c_2>0$, it holds for every $(\lambda,\mu)$.
\end{proof}

\subsection{Proof of Proposition~\ref{prop:linearized_equivalence}}
\label{app:proof_linearized_equivalence}

\begin{quote}
\textbf{Proposition~\ref{prop:linearized_equivalence}} (Equivalence under linearization)
\textit{
Fix $\lambda,\mu>0$, and consider the linearized model $f_{\mathrm{lin}}$.
\begin{itemize}
\item[(i)] \textbf{Instantiation (O).}
On the linearized model, $\Phi^O(\tau)=R_{\mathcal Q}(\tau)$ for all $\tau$; hence $\tau_{\mathrm{lin}}^{O}=\tau_{\mathrm{lin}}^{J}$.
\item[(ii)] \textbf{Instantiation (F).}
If the frozen FIM is exact, $\widehat F_{\mathcal Q}=A_{\mathcal Q}$, then $\Phi^F(\tau)=R_{\mathcal Q}(\tau)$ for all $\tau$; hence $\tau_{\mathrm{lin}}^{F}=\tau_{\mathrm{lin}}^{J}$.
\end{itemize}
}
\end{quote}

\begin{proof}
Throughout, fix $\lambda,\mu>0$ and let
\begin{equation}
F_{\mathrm{lin}}^{\Phi}(\tau) :=\mathcal L_{\mathrm{task}}^{\mathrm{lin}}(\tau) +\frac{\lambda}{2}\,\Phi(\tau) +\frac{\mu}{2}\|\tau\|^2, \qquad \Phi\in\{\Phi^J,\Phi^F,\Phi^O\},
\label{eq:app_lin_objective}
\end{equation}
where $\mathcal L_{\mathrm{task}}^{\mathrm{lin}}$ denotes the task loss evaluated on the linearized model $f_{\mathrm{lin}}$, and the penalties are evaluated on $f_{\mathrm{lin}}$ as well.

\textbf{Step 1: $\Phi^O=R_{\mathcal Q}$ on the linearized model.}
By definition of $f_{\mathrm{lin}}$,
\begin{equation}
f_{\mathrm{lin}}(x;\theta_0+\tau)-f(x;\theta_0)=J(x)^\top\tau \qquad\text{for every }x\text{ and }\tau .
\end{equation}
Substituting into Eq.~\eqref{eq:odr_definition},
\begin{equation}
\Phi^O(\tau)=\mathbb E_{x\sim\mathcal Q}\!\left[\|J(x)^\top\tau\|^2\right]=\tau^\top A_{\mathcal Q}\tau=R_{\mathcal Q}(\tau).
\label{eq:app_odr_identity}
\end{equation}

\textbf{Step 2: $\Phi^F=R_{\mathcal Q}$ under exact curvature.}
We first
justify the claim, used after Eq.~(19) in the main text, that the FIM at
$\theta_0$ equals $A_{\mathcal Q}$ up to a global scalar under a
Gaussian output likelihood.

Consider the Gaussian output model
\begin{equation}
p(y\mid x;\theta) \;=\; \mathcal{N}\!\bigl(f(x;\theta),\,\sigma^2 I_c\bigr),
\qquad \sigma>0.
\label{eq:gauss-lik}
\end{equation}
The log-likelihood is, up to constants independent of $\theta$,
\begin{equation}
\log p(y\mid x;\theta) \;=\; -\tfrac{1}{2\sigma^2}\,\bigl\|y - f(x;\theta)\bigr\|^2 + \mathrm{const.}
\end{equation}
Differentiating with respect to $\theta$ gives
\begin{equation}
\nabla_\theta \log p(y\mid x;\theta) \;=\; \tfrac{1}{\sigma^2}\,J(x;\theta)\bigl(y - f(x;\theta)\bigr).
\label{eq:score}
\end{equation}
Evaluate at $\theta_0$. Because $y\sim p(y\mid x;\theta_0)$ means
$y - f(x;\theta_0) = \varepsilon$ with $\varepsilon\sim\mathcal N(0,\sigma^2 I_c)$,
substituting Eq.~\eqref{eq:score} into the definition of the FIM
(Eq.~(19)) yields
\begin{align}
F_{\mathcal Q}(\theta_0)
&= \mathbb{E}_{x\sim\mathcal Q}\,\mathbb{E}_{\varepsilon\sim\mathcal N(0,\sigma^2 I_c)}
 \!\left[\tfrac{1}{\sigma^4}\,J(x;\theta_0)\,\varepsilon\varepsilon^\top J(x;\theta_0)^\top\right] \nonumber\\
&= \tfrac{1}{\sigma^2}\,\mathbb{E}_{x\sim\mathcal Q}\!\left[J(x;\theta_0)J(x;\theta_0)^\top\right]
\;=\; \tfrac{1}{\sigma^2}\,A_{\mathcal Q},
\label{eq:fim-eq-aq}
\end{align}
where the second equality uses $\mathbb E[\varepsilon\varepsilon^\top]=\sigma^2 I_c$
and Definition~2. The multiplicative constant $1/\sigma^2$ in
Eq.~\eqref{eq:fim-eq-aq} enters the penalty $\Phi^F(\tau) =
\tau^\top \hat F_{\mathcal Q}\tau$ as a global scalar; absorbing it into
the regularization strength via $\lambda' := \lambda/\sigma^2$ preserves
the optimization problem in Eq.~(17) and lets us identify $\hat
F_{\mathcal Q}$ with $A_{\mathcal Q}$ without loss of generality. This
realizes the ``exact FIM'' scenario in Proposition~2(ii), i.e.
$\hat F_{\mathcal Q} = A_{\mathcal Q}$. Then, directly from Eq.~(20),
\begin{equation}
\Phi^F(\tau) \;=\; \tau^\top \hat F_{\mathcal Q}\,\tau
\;=\; \tau^\top A_{\mathcal Q}\,\tau \;=\; R_{\mathcal Q}(\tau).
\end{equation}

Steps 1--2 show that the three objectives coincide as functions of $\tau$.  The common objective is $\mu$-strongly convex: the linearized task loss is a convex quadratic in $\tau$ (as $f_{\mathrm{lin}}$ is affine in $\tau$), the penalty $R_{\mathcal Q}(\tau)=\tau^\top A_{\mathcal Q}\tau$ is convex since $A_{\mathcal Q}\succeq0$, and the ridge term is $\mu$-strongly convex.  Hence it admits a unique
minimizer, so that $\tau_{\mathrm{lin}}^{J}=\tau_{\mathrm{lin}}^{F}
=\tau_{\mathrm{lin}}^{O}$.

\end{proof}

\subsection{Method-Specific Gaps: Regularity Conditions and Proofs of Theorems~\ref{theorem:gap_fisher} and~\ref{theorem:gap_output}}
\label{app:method_specific_gap}

This appendix spells out the regularity conditions collected in Assumption~\ref{assumption:local_regularity}, derives the mode perturbation (Eq.~\eqref{eq:odr_mode_perturbation}), gives the explicit forms of the constants $C_{\mathrm F}$ and $C_{\mathrm O}$, and provides the complete proofs, followed by several remarks on boundary cases.
\begin{quote}
\textbf{Theorem~\ref{theorem:gap_fisher}} (\textup{(F)}: proxy-independent gap)
\textit{Let Assumption~\ref{assumption:local_regularity} hold and suppose that the frozen curvature satisfies Eq.~\eqref{eq:fisher_uniform_error}. Then there exists a finite constant $C_{\mathrm F}\ge0$, carrying no explicit $(\alpha_t,\beta_t)$ factor, such that $E_{\mathrm F}(t)\le C_{\mathrm F}$.}

\smallskip
\textbf{Theorem~\ref{theorem:gap_output}} (\textup{(O)}: proxy-amplified gap)
\textit{Let Assumption~\ref{assumption:local_regularity} hold and suppose that for task $t\neq s$ the target matrix $A_{\mathcal P_t}$ admits an $(\alpha_t,\beta_t)$-relative spectral bound with respect to the proxy $A_{\mathcal Q}$ (Definition~\ref{definition:relative_spectral_bound}). Then, with $q_t(\alpha_t,\beta_t)$ as in Eq.~\eqref{eq:proxy_spectral_factor}, there exists a finite constant $C_{\mathrm O}\ge0$, independent of $(\alpha_t,\beta_t)$, such that $E_{\mathrm O}(t)\le C_{\mathrm O}\,q_t(\alpha_t,\beta_t)$.}
\end{quote}

\subsubsection{Setup and additional notation}\label{app:gap_setup}
We retain the notation of Section~\ref{sec:proxy_regularization} and abbreviate
\begin{equation}
J_x:=J(x;\theta_0), \qquad A_t:=A_{\mathcal P_t},\qquad \Lambda_t:=\|A_t\|,
\label{eq:app_gap_short_notation}
\end{equation}
so that
\begin{equation}
\mathcal I(\tau\mid\mathcal P_t)=\tau^\top A_t\tau,\qquad R_{\mathcal Q}(\tau)=\tau^\top A_{\mathcal Q}\tau .
\label{eq:app_gap_interference_proxy}
\end{equation}
As in Eq.~\eqref{eq:odr_mode_perturbation}, $H_x$ denotes the symmetric bilinear map collecting the second-order directional derivatives of the $c$ outputs,
\begin{equation}
H_x[u,v]:=\bigl(u^\top\nabla_\theta^2 f_k(x;\theta_0)\,v\bigr)_{k=1}^{c}\in\mathbb R^{c},\qquad \|H_x\|:=\sup_{\|u\|=\|v\|=1}\|H_x[u,v]\| .
\label{eq:app_gap_hessian_map}
\end{equation}
For a scalar output, $H_x[\tau,\tau]$ reduces to the familiar quadratic form $\tau^\top H_x\tau$.

Consistent with the squared-loss NTK setting of Section~\ref{sec:framework}, the two task losses are
\begin{align}
\mathcal L_{\mathrm{task}}^{\mathrm{lin}}(\tau)&:=\frac12\,\mathbb E_{(x,y)\sim\mathcal D_s}\!\left[\bigl\|y-f(x;\theta_0)-J_x^\top\tau\bigr\|^2\right],\label{eq:app_gap_linear_task_loss}\\
\mathcal L_{\mathrm{task}}^{\mathrm{std}}(\tau)&:=\frac12\,\mathbb E_{(x,y)\sim\mathcal D_s}\!\left[\bigl\|y-f(x;\theta_0+\tau)\bigr\|^2\right].\label{eq:app_gap_standard_task_loss}
\end{align}
For $m\in\{\mathrm{lin},\mathrm{std}\}$ and
$\Phi\in\{\mathrm J,\mathrm F,\mathrm O\}$, write the objectives of Eq.~\eqref{eq:method_specific_solution} as
\begin{equation}
F_m^{\Phi}(\tau)
:=\mathcal L_{\mathrm{task}}^{m}(\tau)
+\frac{\lambda}{2}\,\Phi_m(\tau)
+\frac{\mu}{2}\|\tau\|^2 ,
\qquad
\tau_m^{\Phi}\in\arg\min_\tau F_m^{\Phi}(\tau),
\label{eq:app_gap_objectives}
\end{equation}
with the mode-specific penalties given in Eq.~\eqref{eq:frozen_penalty_pairs}--\eqref{eq:odr_penalty_pair}.  By
Proposition~\ref{prop:linearized_equivalence}(i),
\begin{equation}
F_{\mathrm{lin}}^{\mathrm O}=F_{\mathrm{lin}}^{\mathrm J},
\qquad
\tau_{\mathrm{lin}}^{\mathrm O}=\tau_{\mathrm{lin}}^{\mathrm J},
\label{eq:app_gap_linear_odr_equals_jacobian}
\end{equation}
where uniqueness of the linearized minimizers follows from the ridge term
$\frac{\mu}{2}\|\tau\|^2$.

The proof strategy rests on decomposing the standard-mode objectives into their linearized counterparts plus explicit perturbation terms. We introduce two such perturbations.

The \emph{task-loss mode perturbation} captures the difference between the standard and linearized task losses,
\begin{equation}
\Psi(\tau):=\mathcal L_{\mathrm{task}}^{\mathrm{std}}(\tau)-\mathcal L_{\mathrm{task}}^{\mathrm{lin}}(\tau).
\label{eq:app_gap_psi_definition}
\end{equation}
This term reflects the non-linearity of the model $f$ itself and is common to all three regularizations. It appears whenever the linearized task loss is replaced by the standard one, regardless of the regularization mechanism.

The \emph{output-drift mode perturbation} captures the difference between the standard and linearized output-drift penalties:
\begin{equation}
\Delta_{\mathrm O}(\tau):=\Phi_{\mathrm{std}}^{\mathrm O}(\tau)-\Phi_{\mathrm{lin}}^{\mathrm O}(\tau).
\label{eq:app_gap_delta_o_definition}
\end{equation}
This term is specific to regularization \textup{(O)}: because the trajectory penalty $\Phi_{\mathrm{std}}^{\mathrm O}$ tracks the exact (non-linear) output drift while its linearized counterpart reduces to $R_{\mathcal Q}$ (Proposition~\ref{prop:linearized_equivalence}~(i)), $\Delta_{\mathrm O}$ measures precisely the higher-order contribution that the non-linearity of $f$ injects into the penalty. No analogous term arises for \textup{(J)} or \textup{(F)}, because their penalties are frozen quadratic forms identical in both modes. Both perturbations vanish identically in the exact linear regime.

From these definitions, the standard-mode objectives decompose as follows.  For instantiation \textup{(J)}, the penalty $R_{\mathcal Q}(\tau)$ is the same frozen quadratic form in both modes (Eq.~\eqref{eq:frozen_penalty_pairs}), so the only source of difference between $F_{\mathrm{std}}^{\mathrm J}$ and
$F_{\mathrm{lin}}^{\mathrm J}$ is the task loss:
\begin{equation}
F_{\mathrm{std}}^{\mathrm J}=F_{\mathrm{lin}}^{\mathrm J}+\Psi .
\label{eq:app_gap_jacobian_mode_relation}
\end{equation}
For instantiation \textup{(O)}, the standard-mode penalty additionally departs from the Jacobian penalty by $\Delta_{\mathrm O}$. Since $\Phi_{\mathrm{std}}^{\mathrm O}=R_{\mathcal Q}+\Delta_{\mathrm O}$ (by Eq.~\eqref{eq:app_gap_delta_o_definition} and $\Phi_{\mathrm{lin}}^{\mathrm O}=R_{\mathcal Q}$) and the task loss and ridge terms are shared with \textup{(J)},
\begin{equation}
F_{\mathrm{std}}^{\mathrm O}=F_{\mathrm{std}}^{\mathrm J}+\frac{\lambda}{2}\,\Delta_{\mathrm O} .
\label{eq:app_gap_odr_standard_relation}
\end{equation}
For instantiation \textup{(F)}, the frozen curvature decomposition $\widehat F_{\mathcal Q}=A_{\mathcal Q}+E_{\mathcal Q}$ (Eq.~\eqref{eq:fisher_uniform_error}) gives $\Phi_m^{\mathrm F}(\tau)=\tau^\top\widehat F_{\mathcal Q}\tau=R_{\mathcal Q}(\tau)+\tau^\top E_{\mathcal Q}\tau$ for either mode $m$, so that
\begin{equation}
F_m^{\mathrm F}=F_m^{\mathrm J}+\frac{\lambda}{2}\,\tau^\top E_{\mathcal Q}\tau .
\label{eq:app_gap_fisher_objective_perturbation}
\end{equation}
Crucially, the Fisher perturbation $\frac{\lambda}{2}\tau^\top E_{\mathcal Q}\tau$ is the \emph{same} fixed quadratic form in both modes, whereas the output-drift perturbation $\Delta_{\mathrm O}$ is present only in standard mode. This structural asymmetry is ultimately responsible for the proxy-independent gap of \textup{(F)} and the proxy-amplified gap of \textup{(O)}.

We use throughout the signed mode change $\delta_\Phi(t)$ and the regularizer-specific gap $E_\Phi(t)$ defined in Eqs.~\eqref{eq:signed_mode_change} and~\eqref{eq:excess_gap}. For the analysis below it is convenient to name two auxiliary quantities that do not appear in the main text, the \emph{raw mode gap} and the \emph{task-mode baseline},
\begin{equation}
G_\Phi(t):=|\delta_\Phi(t)|,
\qquad
B_t:=G_{\mathrm J}(t),
\label{eq:app_gap_raw_and_baseline}
\end{equation}
which by the triangle inequality satisfy
\begin{equation}
G_\Phi(t)\le B_t+E_\Phi(t).
\label{eq:app_gap_decomposition}
\end{equation}
Note that each regularization is compared against its \emph{own} linearized solution $\tau_{\mathrm{lin}}^{\Phi}$; consequently, $E_{\mathrm J}(t)=0$ does not imply $G_{\mathrm J}(t)=0$, since the raw mode gap still contains the baseline $B_t$.

\subsubsection{Regularity conditions}\label{app:gap_assumptions}

Assumptions \textup{(A1)--(A3)} below spell out the \emph{regularity conditions} collected in Assumption~\ref{assumption:local_regularity} of the main text. All these are standard in the local analysis of fine-tuning and neural tangent kernel theory~\cite{ntk2018,tta2023}.

\paragraph{(A1) Locality and bounded solutions}
All six solutions $\tau_m^{\Phi}$ and every line segment joining the pairs compared below lie in
\begin{equation}
\mathbb B_R:=\{\tau\in\mathbb R^p:\|\tau\|\le R\}.
\label{eq:app_gap_solution_ball}
\end{equation}
This reflects the empirical observation that fine-tuning from a pre-trained model moves the weights only within a small neighborhood of $\theta_0$~\cite{tta2023,Neyshabur2020}, which is further enforced here by the ridge term $\frac{\mu}{2}\|\tau\|^2$ in Eq.~\eqref{eq:method_specific_solution}.

\paragraph{(A2) Smoothness}
For almost every $x$ under $\mathcal D_s$ and $\mathcal Q$, the map $\theta\mapsto f(x;\theta)$ is three times continuously differentiable on $\theta_0+\mathbb B_R$.  There is a finite $M_J$ such that
\begin{equation}
\|J_x\|\le M_J,
\label{eq:app_gap_jacobian_bound}
\end{equation}
and the second- and third-order derivative bounds are integrable. In particular, Taylor's theorem gives
\begin{equation}
f(x;\theta_0+\tau)-f(x;\theta_0)=J_x^\top\tau+\frac12 H_x[\tau,\tau]+\rho_x(\tau),
\label{eq:app_gap_model_taylor}
\end{equation}
with
\begin{equation}
\|\rho_x(\tau)\|\le C_{3,x}\|\tau\|^3,\qquad \|D\rho_x(\tau)\|\le C_{3,x}'\|\tau\|^2,
\label{eq:app_gap_taylor_remainder_bounds}
\end{equation}
for integrable $C_{3,x},C_{3,x}'$.
These conditions are standard in the local analysis of neural networks with smooth activations~\cite{ntk2018,Lee2019,Zhu2019}.

\paragraph{(A3) Bounded residual.}
Let
\begin{equation}
r(x):=y-f(x;\theta_0),\qquad R_{\max}:=\sup_{(x,y)\sim\mathcal D_s}\|r(x)\|<\infty.
\label{eq:app_gap_residual_bound}
\end{equation}
This is automatic for bounded labels and a pre-trained model with bounded outputs, as assumed in standard fine-tuning analyses~\cite{tta2023,Kumar2022}.

In addition, define the first moments of the model Hessian norm under the source and proxy distributions,
\begin{equation}
\overline H_s
:=\mathbb E_{(x,y)\sim\mathcal D_s}\!\left[\|H_x\|\right],
\qquad
\overline H_{\mathcal Q}
:=\mathbb E_{x\sim\mathcal Q}\!\left[\|H_x\|\right],
\label{eq:app_gap_hessian_moments}
\end{equation}
which are finite by the integrability of the second-order derivative bounds in \textup{(A2)}.

\subsubsection{Two auxiliary lemmas}\label{app:gap_auxiliary_lemmas}

We collect two lemmas that will be used throughout the proof.

\begin{lemma}[Interference-difference identity]
\label{lem:app_interference_difference_identity}
For any $u,v\in\mathbb R^p$,
\begin{equation}
\mathcal I(u\mid\mathcal P_t)-\mathcal I(v\mid\mathcal P_t)=(u-v)^\top A_t(u+v) .
\label{eq:app_gap_interference_identity_exact}
\end{equation}
In particular, if $\|u\|,\|v\|\le R$, then
\begin{equation}
\bigl|\mathcal I(u\mid\mathcal P_t)-\mathcal I(v\mid\mathcal P_t)\bigr|\le2R\sqrt{\Lambda_t}\,\|A_t^{1/2}(u-v)\| .
\label{eq:app_gap_interference_identity_radius}
\end{equation}
\end{lemma}

\begin{proof}
Since $A_t$ is symmetric,
\begin{equation}
u^\top A_tu-v^\top A_tv
=u^\top A_tu-u^\top A_tv+u^\top A_tv-v^\top A_tv
=(u-v)^\top A_t(u+v),
\end{equation}
which proves Eq.~\eqref{eq:app_gap_interference_identity_exact}. From this identity two useful bounds follow.

\emph{First bound.}  Applying the Cauchy-Schwarz inequality and the operator norm bound $\|A_t(u+v)\|\le\Lambda_t\|u+v\|$ gives
\begin{equation}
\bigl|(u-v)^\top A_t(u+v)\bigr|
\le\|u-v\|\,\|A_t(u+v)\|
\le\Lambda_t\|u-v\|\,\|u+v\| .
\end{equation}
The triangle inequality $\|u+v\|\le\|u\|+\|v\|$ then yields
\begin{equation}
\bigl|\mathcal I(u\mid\mathcal P_t)-\mathcal I(v\mid\mathcal P_t)\bigr|
\le\Lambda_t\|u-v\|\,(\|u\|+\|v\|).
\label{eq:app_gap_interference_identity_norm}
\end{equation}

\emph{Second bound.}  Since $A_t\succeq0$, we can factor
$A_t=A_t^{1/2}A_t^{1/2}$ and write
\begin{equation}
(u-v)^\top A_t(u+v)
=\bigl\langle A_t^{1/2}(u-v),\,A_t^{1/2}(u+v)\bigr\rangle .
\end{equation}
Cauchy--Schwarz then gives
\begin{equation}
\bigl|\mathcal I(u\mid\mathcal P_t)-\mathcal I(v\mid\mathcal P_t)\bigr|
\le\|A_t^{1/2}(u-v)\|\,\|A_t^{1/2}(u+v)\| .
\end{equation}
To control the second factor, observe that
$\|A_t^{1/2}w\|^2=w^\top A_tw\le\Lambda_t\|w\|^2$ for any $w$, so
$\|A_t^{1/2}(u+v)\|\le\sqrt{\Lambda_t}\,\|u+v\|
\le\sqrt{\Lambda_t}\,(\|u\|+\|v\|)$.  When $\|u\|,\|v\|\le R$, this
gives $\|A_t^{1/2}(u+v)\|\le2R\sqrt{\Lambda_t}$, and hence
\begin{equation}
\bigl|\mathcal I(u\mid\mathcal P_t)-\mathcal I(v\mid\mathcal P_t)\bigr|
\le2R\sqrt{\Lambda_t}\,\|A_t^{1/2}(u-v)\|.
\end{equation}
\end{proof}

The next lemma is a standard tool from perturbation analysis~\cite{bonnans2013,rajeswaran2019}: it turns an additive perturbation $h$ of the objective into an explicit bound on the resulting displacement of the minimizer.  We use a path-averaged Hessian along the segment joining the two solutions, so that only pathwise (rather than global) strong convexity, that is a mild condition consistent with the local, non-convex setting of standard fine-tuning~\cite{tta2023,Neyshabur2020,Kumar2022}, is needed.

\begin{lemma}[Pathwise solution perturbation]
\label{lem:app_pathwise_solution_perturbation}
Let $F$ and $h$ be differentiable, let $u$ satisfy $\nabla F(u)=0$, and let $v$ satisfy $\nabla F(v)+\nabla h(v)=0$.  Define
\begin{equation}
\mathcal G_{u,v}:=\int_0^1\nabla^2F\bigl(u+s(v-u)\bigr)\,ds .
\label{eq:app_gap_path_hessian_definition}
\end{equation}
Then
\begin{equation}
\mathcal G_{u,v}(v-u)=-\nabla h(v).
\label{eq:app_gap_path_perturbation_identity}
\end{equation}
If $\mathcal G_{u,v}\succeq\nu I$ for some $\nu>0$, then
\begin{equation}
\|v-u\|\le\frac{1}{\nu}\|\nabla h(v)\| .
\label{eq:app_gap_path_perturbation_norm_bound}
\end{equation}
\end{lemma}

\begin{proof}
The fundamental theorem of calculus gives
\begin{equation*}
\nabla F(v)-\nabla F(u)
=\int_0^1\nabla^2F\bigl(u+s(v-u)\bigr)(v-u)\,ds
=\mathcal G_{u,v}(v-u).
\end{equation*}
Substituting $\nabla F(u)=0$ and $\nabla F(v)=-\nabla h(v)$ into the
left-hand side yields
\eqref{eq:app_gap_path_perturbation_identity}.  If
$\mathcal G_{u,v}\succeq\nu I$, then $\mathcal G_{u,v}$ is invertible
with $\|\mathcal G_{u,v}^{-1}\|\le 1/\nu$, and taking norms in Eq.~\eqref{eq:app_gap_path_perturbation_identity} yields Eq.~\eqref{eq:app_gap_path_perturbation_norm_bound}.
\end{proof}

\subsubsection{The task-mode baseline $B_t$}
\label{app:gap_jacobian_baseline}
We first isolate the non-linearity of the task loss, which is present even when the regularization mechanism itself is a frozen quadratic form. Recall from Eq.~\eqref{eq:app_gap_psi_definition} that the task-loss mode perturbation $\Psi(\tau)=\mathcal L_{\mathrm{task}}^{\mathrm{std}}(\tau)
-\mathcal L_{\mathrm{task}}^{\mathrm{lin}}(\tau)$ captures the effect of replacing the linearized task loss by the standard one.  The following lemma expands $\Psi$ around $\tau=0$ and provides a gradient bound suitable for the pathwise perturbation analysis to follow.

\begin{lemma}[Task-loss perturbation]
\label{lem:app_task_loss_perturbation}
Under \textup{(A1)--(A3)}, uniformly for $\tau\in\mathbb B_R$,
\begin{equation}
\Psi(\tau)=-\frac12\,\mathbb E_{(x,y)\sim\mathcal D_s}\!\left[\bigl\langle r(x),H_x[\tau,\tau]\bigr\rangle\right]
+O(\|\tau\|^3),
\label{eq:app_gap_task_perturbation_expansion}
\end{equation}
where $r(x)=y-f(x;\theta_0)$ is the pre-training residual
(Eq.~\eqref{eq:app_gap_residual_bound}) and $H_x[\tau,\tau]$ is the
model Hessian bilinear form
(Eq.~\eqref{eq:app_gap_hessian_map}).  Moreover, there exists a finite
constant $C_\Psi$, depending only on $M_J$, $R_{\max}$, $R$, and the
moments under $\mathcal D_s$ of $\|H_x\|$ and of the Taylor remainder
constants in \textup{(A2)}, but not on $\tau$, such that
\begin{equation}
\|\nabla\Psi(\tau)\|
\le R_{\max}\overline H_s\|\tau\|+C_\Psi\|\tau\|^2 .
\label{eq:app_gap_task_perturbation_gradient_bound}
\end{equation}
\end{lemma}

The leading term
$-\frac12\mathbb E\langle r(x),H_x[\tau,\tau]\rangle$ is the coupling
between the pre-training residual and the model Hessian.  It vanishes
when the pre-trained model is a perfect fit ($r(x)\equiv0$), but $\Psi$
itself need not: setting $r(x)\equiv0$ in the exact decomposition
Eq.~\eqref{eq:app_gap_task_perturbation_exact_decomposition} established
in the proof below gives
\begin{equation}
\Psi(\tau)
=\frac12\,\mathbb E_{\mathcal D_s}\!\left[\bigl\|f(x;\theta_0+\tau)-f(x;\theta_0)\bigr\|^2-\bigl\|J_x^\top\tau\bigr\|^2\right]
=\frac12\,\mathbb E_{\mathcal D_s}\!\left[\bigl\langle J_x^\top\tau,H_x[\tau,\tau]\bigr\rangle\right]
+O(\|\tau\|^4).
\label{eq:app_gap_task_perturbation_zero_residual}
\end{equation}
A perfect fit therefore makes $\Psi$ of third rather than second order
in $\tau$, and reduces the gradient bound
Eq.~\eqref{eq:app_gap_task_perturbation_gradient_bound} to
$\|\nabla\Psi(\tau)\|\le C_\Psi\|\tau\|^2$ (as $R_{\max}=0$), without
making $\Psi$ vanish.  In contrast, $\Psi\equiv0$ on $\mathbb B_R$ in the
exact linear regime of Eq.~\eqref{eq:app_gap_exact_linear_model_condition},
where $f$ is affine in $\tau$ on $\theta_0+\mathbb B_R$, so that
$H_x[\tau,\tau]$ and $\rho_x(\tau)$ both vanish.

\begin{proof}
For fixed $(x,y)$, decompose the increment of $f$ in three parts according to the Taylor expansion (Eq.~\eqref{eq:app_gap_model_taylor}):
\begin{equation}
a_x(\tau):=J_x^\top\tau,
\qquad
b_x(\tau):=\frac12H_x[\tau,\tau],
\qquad
c_x(\tau):=\rho_x(\tau),
\label{eq:app_gap_task_expansion_components}
\end{equation}
whose orders in $\tau$ are, respectively, $\|a_x\|=O(\|\tau\|)$, $\|b_x\|=O(\|\tau\|^2)$, and $\|c_x\|=O(\|\tau\|^3)$; the last uses the remainder bound in Eq.~\eqref{eq:app_gap_taylor_remainder_bounds}. Consequently the residual of the standard model is
\begin{equation}
y-f(x;\theta_0+\tau)
=r(x)-a_x(\tau)-b_x(\tau)-c_x(\tau),
\label{eq:app_gap_standard_residual_expansion}
\end{equation}
while the residual of the linearized model is $r(x)-a_x(\tau)$.
 
\emph{Step 1: exact algebraic decomposition of $\Psi$.} Let $u_x(\tau):=r(x)-a_x(\tau)$ denote the linearized residual and $w_x(\tau):=b_x(\tau)+c_x(\tau)$ collect the two non-linear corrections.  Then the standard residual equals $u_x-w_x$.  Expanding $\|u_x-w_x\|^2=\|u_x\|^2-2\langle u_x,w_x\rangle+\|w_x\|^2$ and subtracting $\|u_x\|^2$ gives
\begin{equation}
\Psi(\tau)
=\frac12\,\mathbb E_{\mathcal D_s}
\!\left[\|u_x-w_x\|^2-\|u_x\|^2\right]
=-\mathbb E_{\mathcal D_s}\!\left[\langle u_x,w_x\rangle\right]
+\frac12\,\mathbb E_{\mathcal D_s}\!\left[\|w_x\|^2\right].
\label{eq:app_gap_task_perturbation_exact_decomposition}
\end{equation}
 
\emph{Step 2: identification of the leading term.} Expanding the two expectations in Eq.~\eqref{eq:app_gap_task_perturbation_exact_decomposition} into monomials in $(a_x,b_x,c_x)$ and using the orders recorded above,
\begin{align}
\langle u_x,w_x\rangle
&=\underbrace{\langle r,b_x\rangle}_{O(\|\tau\|^2)}
+\underbrace{\langle r,c_x\rangle}_{O(\|\tau\|^3)}
-\underbrace{\langle a_x,b_x\rangle}_{O(\|\tau\|^3)}
-\underbrace{\langle a_x,c_x\rangle}_{O(\|\tau\|^4)},
\label{eq:app_gap_task_uw_monomial_orders}\\
\|w_x\|^2
&=\underbrace{\|b_x\|^2}_{O(\|\tau\|^4)}
+\underbrace{2\langle b_x,c_x\rangle}_{O(\|\tau\|^5)}
+\underbrace{\|c_x\|^2}_{O(\|\tau\|^6)} .
\label{eq:app_gap_task_ww_monomial_orders}
\end{align}
The only quadratic-in-$\tau$ contribution is $-\mathbb E\langle r,b_x\rangle=-\tfrac12\mathbb E\langle r(x),H_x[\tau,\tau]\rangle$, which is precisely the leading term of Eq.~\eqref{eq:app_gap_task_perturbation_expansion}; every other contribution is at least cubic in $\tau$.  This proves Eq.~\eqref{eq:app_gap_task_perturbation_expansion}.
 
\emph{Step 3: gradient bound.} We now differentiate.  For the leading term, using the symmetry of the bilinear form $H_x$, its directional derivative in the direction $v$ is
\begin{equation}
D\!\left[-\frac12\bigl\langle r(x),H_x[\tau,\tau]\bigr\rangle\right][v]
=-\bigl\langle r(x),H_x[\tau,v]\bigr\rangle .
\label{eq:app_gap_task_leading_differential}
\end{equation}
By Cauchy--Schwarz and the operator norm bound on $H_x$,
\begin{equation}
\bigl|D\!\left[-\tfrac12\langle r(x),H_x[\tau,\tau]\rangle\right][v]\bigr|
\le\|r(x)\|\,\|H_x\|\,\|\tau\|\,\|v\| .
\end{equation}
Taking expectations, using $\|r(x)\|\le R_{\max}$ from~\textup{(A3)} and
$\mathbb E\|H_x\|=\overline H_s$, the gradient of the leading term has
norm at most $R_{\max}\overline H_s\|\tau\|$.
 
For every remaining monomial listed in~\eqref{eq:app_gap_task_uw_monomial_orders}--\eqref{eq:app_gap_task_ww_monomial_orders}, the derivative in $\tau$ is one order lower in $\tau$ than the monomial itself.  Since each such monomial is $O(\|\tau\|^3)$ or higher, its gradient is $O(\|\tau\|^2)$ or higher.  The associated constants take the form of products of:
$M_J$ (from $\|a_x\|$), one or two factors of $\|H_x\|$ (from $b_x$ and
its derivative), $C_{3,x}$ or $C_{3,x}'$ (from $\rho_x=c_x$ and its
derivative, via Eq.~\eqref{eq:app_gap_taylor_remainder_bounds}), and
$\|r(x)\|\le R_{\max}$.  Their expectations under $\mathcal D_s$ are
finite by the integrability assumption in~\textup{(A2)}.  Summing these
finite expectations gives a constant, which we denote $C_\Psi$, such
that on $\mathbb B_R$ the total gradient contribution of the remainders
is at most $C_\Psi\|\tau\|^2$.  Adding the leading contribution
yields Eq.~\eqref{eq:app_gap_task_perturbation_gradient_bound}.
\end{proof}

\begin{proposition}[Bound on the task-mode baseline]
\label{prop:app_jacobian_baseline_bound}
Under \textup{(A1)--(A3)},
\begin{equation}
B_t\le B_t^{\mathrm{ub}}
:=\frac{2\Lambda_tR}{\mu}
\left(R_{\max}\overline H_sR+C_\Psi R^2\right).
\label{eq:app_gap_baseline_explicit_bound}
\end{equation}
\end{proposition}

\begin{proof}
Let
$d_{\mathrm J}:=\tau_{\mathrm{std}}^{\mathrm J}-\tau_{\mathrm{lin}}^{\mathrm J}$.
From Eq.~\eqref{eq:app_gap_jacobian_mode_relation} and first-order
optimality,
\begin{equation}
\nabla F_{\mathrm{lin}}^{\mathrm J}(\tau_{\mathrm{lin}}^{\mathrm J})=0,
\qquad
\nabla F_{\mathrm{lin}}^{\mathrm J}(\tau_{\mathrm{std}}^{\mathrm J})
=-\nabla\Psi(\tau_{\mathrm{std}}^{\mathrm J}).
\label{eq:app_gap_baseline_stationarity}
\end{equation}
The linearized reference objective is $\mu$-strongly convex because
$A_{\mathcal P_s},A_{\mathcal Q}\succeq0$.
Lemma~\ref{lem:app_pathwise_solution_perturbation} therefore gives
\begin{equation}
\|d_{\mathrm J}\|
\le\frac{1}{\mu}\|\nabla\Psi(\tau_{\mathrm{std}}^{\mathrm J})\|
\le\frac{1}{\mu}\left(R_{\max}\overline H_sR+C_\Psi R^2\right).
\label{eq:app_gap_baseline_solution_distance}
\end{equation}
Applying Eq.~\eqref{eq:app_gap_interference_identity_norm} with
$\|\tau_{\mathrm{std}}^{\mathrm J}\|,\|\tau_{\mathrm{lin}}^{\mathrm J}\|\le R$
yields \eqref{eq:app_gap_baseline_explicit_bound}.
\end{proof}

Two features of Eq.~\eqref{eq:app_gap_baseline_explicit_bound} are worth noting.  First, the bound contains no explicit $(\alpha_t,\beta_t)$ factor: the displacement $d_{\mathrm J}$ is controlled in ordinary norm via the ridge coefficient $\mu$ alone, without invoking the
proxy--target spectral relation.  Second, the bound involves only the
source-distribution Hessian moment $\overline H_s$ and no
proxy-evaluated moment such as $\overline H_{\mathcal Q}$, because
$\Psi$ is defined as a difference of task losses over $\mathcal D_s$
alone and never sees the proxy $\mathcal Q$.  These two features
distinguish the baseline from the regularizer-specific gaps analyzed in the next two
subsections.

\subsubsection{Regularizer-specific gap of instantiation (F)}\label{app:gap_fisher_excess}
 
We compare \textup{(F)} with the Jacobian reference \emph{separately in
each mode}.  This is the additional step required because
$\tau_{\mathrm{lin}}^{\mathrm F}$ is the instantiation's own linearized
solution rather than a shared Jacobian solution.

Before proceeding, we record two structural ingredients that the
argument relies on.

\paragraph{Local curvature of the reference objective}
On $\mathbb B_R$, we assume that the standard task loss has (possibly
non-strict) local curvature
\begin{equation}
\nabla^2\mathcal L_{\mathrm{task}}^{\mathrm{std}}(\tau)\succeq\kappa I,
\qquad \kappa\ge 0,
\label{eq:app_gap_local_task_convexity}
\end{equation}
which is standard in the local analysis of fine-tuning near a
pre-trained solution~\cite{tta2023,Kumar2022}. Set
\begin{equation}
\nu_{\mathrm{lin}}:=\mu, \qquad \nu_{\mathrm{std}}:=\kappa+\mu .
\label{eq:app_gap_nu_definitions}
\end{equation}
Combining Eq.~\eqref{eq:app_gap_local_task_convexity} with
$A_{\mathcal P_s},A_{\mathcal Q}\succeq 0$, the Jacobian-reference
objectives obey
\begin{align}
\nabla^2 F_{\mathrm{lin}}^{\mathrm J}
&=A_{\mathcal P_s}+\lambda A_{\mathcal Q}+\mu I
\succeq\lambda A_{\mathcal Q}+\nu_{\mathrm{lin}}I,
\label{eq:app_gap_linear_reference_curvature}\\
\nabla^2 F_{\mathrm{std}}^{\mathrm J}(\tau)
&\succeq\lambda A_{\mathcal Q}+\nu_{\mathrm{std}}I .
\label{eq:app_gap_standard_reference_curvature}
\end{align}
Only pathwise lower bounds along the segments joining the compared
solutions are needed here; global convexity is not required. More
generally, if the standard task loss has negative curvature bounded
below by $-\zeta I$ with $\mu>\zeta$, all subsequent results remain
valid after replacing $\nu_{\mathrm{std}}$ by $\mu-\zeta$.

\paragraph{The frozen curvature error as a shared perturbation}
The frozen curvature decomposition
$\widehat F_{\mathcal Q}=A_{\mathcal Q}+E_{\mathcal Q}$ of
Eq.~\eqref{eq:fisher_uniform_error} enters both modes as the
\emph{same} fixed positive-semidefinite matrix, so that
Eq.~\eqref{eq:app_gap_fisher_objective_perturbation} can be rewritten as
\begin{equation}
F_m^{\mathrm F}(\tau)
=F_m^{\mathrm J}(\tau)+\frac{\lambda}{2}\,\tau^\top E_{\mathcal Q}\tau,
\qquad m\in\{\mathrm{lin},\mathrm{std}\}.
\label{eq:app_gap_fisher_shared_perturbation}
\end{equation}
When the analysis is applied to a family of proxies, we write
$\varepsilon_{\mathrm F}(\mathcal Q):=\|E_{\mathcal Q}\|$ and adopt
the uniform bound
$\sup_{\mathcal Q}\varepsilon_{\mathrm F}(\mathcal Q)\le
\overline\varepsilon_{\mathrm F}$ from Eq.~\eqref{eq:fisher_uniform_error}.

\paragraph{Modewise displacement and pathwise perturbation}
For $m\in\{\mathrm{lin},\mathrm{std}\}$, define the displacement of the
\textup{(F)} solution from its Jacobian counterpart in the same mode,
\begin{equation}
d_m^{\mathrm F}:=\tau_m^{\mathrm F}-\tau_m^{\mathrm J}.
\label{eq:app_gap_fisher_modewise_displacement}
\end{equation}
The first-order conditions of $F_m^{\mathrm J}$ and $F_m^{\mathrm F}$,
combined with Eq.~\eqref{eq:app_gap_fisher_shared_perturbation}, give
\begin{equation}
\nabla F_m^{\mathrm J}(\tau_m^{\mathrm J})=0, \qquad
\nabla F_m^{\mathrm J}(\tau_m^{\mathrm F})
+\lambda E_{\mathcal Q}\tau_m^{\mathrm F}=0.
\label{eq:app_gap_fisher_modewise_stationarity}
\end{equation}
Define the path-averaged Hessian of the reference objective along the
segment from $\tau_m^{\mathrm J}$ to $\tau_m^{\mathrm F}$,
\begin{equation}
\mathcal G_m^{\mathrm F}:=\int_0^1\nabla^2 F_m^{\mathrm J}
\bigl(\tau_m^{\mathrm J}+s\,d_m^{\mathrm F}\bigr)\,ds .
\label{eq:app_gap_fisher_path_hessian}
\end{equation}
By
Eqs.~\eqref{eq:app_gap_linear_reference_curvature}--\eqref{eq:app_gap_standard_reference_curvature},
\begin{equation}
\mathcal G_m^{\mathrm F}
\succeq\lambda A_{\mathcal Q}+\nu_m I
\succeq\nu_m I ,
\label{eq:app_gap_fisher_path_lower_bound}
\end{equation}
so that $\mathcal G_m^{\mathrm F}$ is $\nu_m$-strongly positive.
Applying Lemma~\ref{lem:app_pathwise_solution_perturbation} with
$F=F_m^{\mathrm J}$, $h(\tau)=\tfrac{\lambda}{2}\tau^\top E_{\mathcal Q}\tau$,
and using $\|\tau_m^{\mathrm F}\|\le R$ from \textup{(A1)},
\begin{equation}
\|d_m^{\mathrm F}\|
\le\frac{1}{\nu_m}\|\lambda E_{\mathcal Q}\tau_m^{\mathrm F}\|
\le\frac{\lambda\,\varepsilon_{\mathrm F}(\mathcal Q)\,R}{\nu_m}.
\label{eq:app_gap_fisher_solution_distance_bound}
\end{equation}
Notice that Eq.~\eqref{eq:app_gap_fisher_solution_distance_bound}
controls $d_m^{\mathrm F}$ in ordinary Euclidean norm; no
target--proxy spectral relation is invoked at this stage.

\paragraph{From displacement to interference difference}
Combining Eq.~\eqref{eq:app_gap_fisher_solution_distance_bound} with the
first inequality of
Lemma~\ref{lem:app_interference_difference_identity}
(Eq.~\eqref{eq:app_gap_interference_identity_norm}) and using
$\|\tau_m^{\mathrm F}\|,\|\tau_m^{\mathrm J}\|\le R$ yields
\begin{equation}
\bigl|\mathcal I(\tau_m^{\mathrm F}\mid\mathcal P_t)
-\mathcal I(\tau_m^{\mathrm J}\mid\mathcal P_t)\bigr|
\le\Lambda_t\|d_m^{\mathrm F}\|
\bigl(\|\tau_m^{\mathrm F}\|+\|\tau_m^{\mathrm J}\|\bigr)
\le\frac{2\lambda\Lambda_t\,\varepsilon_{\mathrm F}(\mathcal Q)\,R^2}{\nu_m}.
\label{eq:app_gap_fisher_modewise_interference_difference}
\end{equation}
Expanding the regularizer-specific gap according to its definition
(Eq.~\eqref{eq:excess_gap}) and applying the triangle inequality,
\begin{align}
E_{\mathrm F}(t)
&=\Bigl|
\bigl[\mathcal I(\tau_{\mathrm{std}}^{\mathrm F}\mid\mathcal P_t)
-\mathcal I(\tau_{\mathrm{lin}}^{\mathrm F}\mid\mathcal P_t)\bigr]
-\bigl[\mathcal I(\tau_{\mathrm{std}}^{\mathrm J}\mid\mathcal P_t)
-\mathcal I(\tau_{\mathrm{lin}}^{\mathrm J}\mid\mathcal P_t)\bigr]
\Bigr| \nonumber\\
&\le
\bigl|\mathcal I(\tau_{\mathrm{std}}^{\mathrm F}\mid\mathcal P_t)
-\mathcal I(\tau_{\mathrm{std}}^{\mathrm J}\mid\mathcal P_t)\bigr|
+\bigl|\mathcal I(\tau_{\mathrm{lin}}^{\mathrm F}\mid\mathcal P_t)
-\mathcal I(\tau_{\mathrm{lin}}^{\mathrm J}\mid\mathcal P_t)\bigr| .
\label{eq:app_gap_fisher_excess_triangle}
\end{align}
Applying Eq.~\eqref{eq:app_gap_fisher_modewise_interference_difference}
to both modes gives
\begin{equation}
E_{\mathrm F}(t)
\le C_{\mathrm F}(t,\mathcal Q)
:=2\lambda\Lambda_t\,\varepsilon_{\mathrm F}(\mathcal Q)\,R^2
\left(\frac{1}{\nu_{\mathrm{std}}}+\frac{1}{\nu_{\mathrm{lin}}}\right).
\label{eq:app_gap_fisher_excess_explicit}
\end{equation}
Under the uniform bound $\varepsilon_{\mathrm F}(\mathcal Q)\le
\overline\varepsilon_{\mathrm F}$ from
Eq.~\eqref{eq:fisher_uniform_error},
\begin{equation}
E_{\mathrm F}(t)\le C_{\mathrm F}(t)
:=2\lambda\Lambda_t\,\overline\varepsilon_{\mathrm F}\,R^2
\left(\frac{1}{\kappa+\mu}+\frac{1}{\mu}\right),
\label{eq:app_gap_fisher_uniform_constant}
\end{equation}
which contains no explicit dependence on the proxy--target spectral
parameters $(\alpha_t,\beta_t)$. This proves
Theorem~\ref{theorem:gap_fisher}, with
$C_{\mathrm F}:=C_{\mathrm F}(t)$.

The mechanism underlying
Eq.~\eqref{eq:app_gap_fisher_uniform_constant} is worth emphasizing.
Because the frozen curvature perturbation
$\tfrac{\lambda}{2}\tau^\top E_{\mathcal Q}\tau$ enters both modes as
the \emph{same} quadratic form
(Eq.~\eqref{eq:app_gap_fisher_shared_perturbation}), the modewise
displacements $d_{\mathrm{lin}}^{\mathrm F}$ and
$d_{\mathrm{std}}^{\mathrm F}$ are each controlled in ordinary Euclidean
norm via the ridge parameters $\nu_m$, without ever invoking the
target--proxy spectral relation. Consequently, the gap envelope
scales with the quality of the curvature estimate,
$\overline\varepsilon_{\mathrm F}$, and is independent in $(\alpha_t,\beta_t)$.
This structural asymmetry is precisely what distinguishes the
proxy-independent behavior of \textup{(F)} from the proxy-amplified behavior
of \textup{(O)} analyzed in the next subsection.

\paragraph{Criterion weighting and KFAC estimators}
The bound Eq.~\eqref{eq:app_gap_fisher_uniform_constant} treats
$\overline\varepsilon_{\mathrm F}$ as an abstract error budget for the
frozen curvature. In practice, FIM-based methods such as
TAK~\cite{tak2026} estimate $\widehat F_{\mathcal Q}$ via a
Kronecker-factored (KFAC) approximation of the Fisher/Gauss--Newton
matrix. Suppose the estimator takes the form
\begin{equation}
\widehat F_{\mathcal Q}^{\mathrm{KFAC}}
=\mathbb E_{x\sim\mathcal Q}\!\left[
J_x\,W(x;\theta_0)\,J_x^\top\right]
+E_{\mathrm{KFAC}},
\label{eq:app_gap_kfac_matrix}
\end{equation}
where $W(x;\theta_0)\succeq 0$ is the output-space criterion curvature
(\textit{e.g.}, $W=I$ for squared loss and the softmax Hessian for
cross-entropy), and $E_{\mathrm{KFAC}}$ collects the Kronecker-factored
approximation error. For any scalar $c>0$, the triangle inequality gives
\begin{equation}
\bigl\|\widehat F_{\mathcal Q}^{\mathrm{KFAC}}-c\,A_{\mathcal Q}\bigr\|
\le\mathbb E_{x\sim\mathcal Q}\!\left[
\|J_x\|^2\,\|W(x;\theta_0)-cI\|\right]
+\|E_{\mathrm{KFAC}}\| .
\label{eq:app_gap_kfac_error_sources}
\end{equation}
After absorbing $c$ into $\lambda$, the right-hand side is a valid
choice of $\varepsilon_{\mathrm F}(\mathcal Q)$; no strictly positive
lower bound on the criterion Hessian $W$ is needed. In the exact
squared-loss Gauss--Newton case, $W=I$ and $E_{\mathrm{KFAC}}=0$, so
$\varepsilon_{\mathrm F}=0$ and
\begin{equation}
\tau_m^{\mathrm F}=\tau_m^{\mathrm J}
\quad(m\in\{\mathrm{lin},\mathrm{std}\}),
\qquad E_{\mathrm F}(t)=0 .
\label{eq:app_gap_exact_ggn_zero_excess}
\end{equation}
Thus the regularizer-specific gap of instantiation \textup{(F)} vanishes whenever the
frozen curvature is exact, $\widehat F_{\mathcal Q}=A_{\mathcal Q}$:
the constant $C_{\mathrm F}$ measures the quality of the curvature
estimate rather than an error intrinsic to the mechanism itself.

\subsubsection{Regularizer-specific gap of instantiation (O)}\label{app:gap_odr_excess}

We now turn to the output-drift regularization. Recall from
Eq.~\eqref{eq:app_gap_linear_odr_equals_jacobian} that its linearized
solution coincides with the Jacobian reference exactly,
$\tau_{\mathrm{lin}}^{\mathrm O}=\tau_{\mathrm{lin}}^{\mathrm J}$, so
unlike \textup{(F)}, no modewise comparison in the linearized regime is
needed: the regularizer-specific gap of \textup{(O)} is driven entirely by the
standard-mode perturbation $\Delta_{\mathrm O}$ of
Eq.~\eqref{eq:app_gap_delta_o_definition}. The analysis proceeds in
three steps. We first derive the Taylor structure of
$\Delta_{\mathrm O}$, which in particular proves the main-text
expansion Eq.~\eqref{eq:odr_mode_perturbation}. We then bound the
displacement between the standard-mode solutions of \textup{(O)} and
\textup{(J)}, exploiting the fact that this perturbation is
\emph{supported on the proxy}. Finally, we propagate the displacement
to the target direction through the relative spectral bound, which is
where the factor $q_t(\alpha_t,\beta_t)$ emerges.

\paragraph{The output-drift mode perturbation and its gradient}
The following lemma expands $\Delta_{\mathrm O}$ around $\tau=0$ and
provides the gradient bound required by the pathwise perturbation
analysis, in direct analogy with
Lemma~\ref{lem:app_task_loss_perturbation} for the task loss.

\begin{lemma}[Output-drift Taylor perturbation]
\label{lem:app_odr_taylor_perturbation}
Under \textup{(A1)--(A3)}, uniformly for $\tau\in\mathbb B_R$,
\begin{equation}
\Delta_{\mathrm O}(\tau)
=\mathbb E_{x\sim\mathcal Q}
\!\left[\bigl\langle J_x^\top\tau,\,H_x[\tau,\tau]\bigr\rangle\right]
+R_4(\tau),
\label{eq:app_gap_odr_expansion}
\end{equation}
where $R_4(\tau)=O(\|\tau\|^4)$, and there exists a finite constant
$C_4$, depending only on the moments of $\|H_x\|$ under $\mathcal Q$
and on the Taylor remainder constants in \textup{(A2)} but not on
$\tau$, such that
\begin{equation}
\|\nabla R_4(\tau)\|\le C_4\|\tau\|^3 .
\label{eq:app_gap_odr_remainder_gradient_bound}
\end{equation}
Moreover,
\begin{equation}
\|\nabla\Delta_{\mathrm O}(\tau)\|
\le3M_J\overline H_{\mathcal Q}\|\tau\|^2+C_4\|\tau\|^3 .
\label{eq:app_gap_odr_gradient_bound}
\end{equation}
\end{lemma}

Two features of Lemma~\ref{lem:app_odr_taylor_perturbation} deserve
comment before the proof. First, the leading term of
Eq.~\eqref{eq:app_gap_odr_expansion} is \emph{cubic} in $\tau$, one
order higher than the leading (quadratic) term of the task-loss
perturbation in Eq.~\eqref{eq:app_gap_task_perturbation_expansion};
this is because the linear contribution $\|J_x^\top\tau\|^2$ is common
to both modes and cancels exactly in the difference. Second, all
moments in Eqs.~\eqref{eq:app_gap_odr_expansion}--\eqref{eq:app_gap_odr_gradient_bound}
are taken under the \emph{proxy} distribution $\mathcal Q$ rather than
the source distribution $\mathcal D_s$---the mirror image of the
baseline analysis in Appendix~\ref{app:gap_jacobian_baseline}, where
only source-distribution moments appeared.

\begin{proof}
For fixed $x\sim\mathcal Q$, decompose the increment of $f$ using the
components introduced in
Eq.~\eqref{eq:app_gap_task_expansion_components}:
\begin{equation}
f(x;\theta_0+\tau)-f(x;\theta_0)=a_x+b_x+c_x .
\label{eq:app_gap_odr_components_sum}
\end{equation}
Expanding the squared norm of this sum,
\begin{equation}
\|a_x+b_x+c_x\|^2
=\|a_x\|^2
+2\langle a_x,b_x\rangle
+\|b_x\|^2
+2\langle a_x,c_x\rangle
+2\langle b_x,c_x\rangle
+\|c_x\|^2 .
\label{eq:app_gap_odr_square_full_expansion}
\end{equation}
Taking expectations over $x\sim\mathcal Q$, the first term is exactly
the linearized penalty,
$\mathbb E_{\mathcal Q}\|a_x\|^2
=\tau^\top A_{\mathcal Q}\tau
=\Phi_{\mathrm{lin}}^{\mathrm O}(\tau)$, and therefore cancels in the
difference $\Delta_{\mathrm O}$. Since $b_x=\tfrac12H_x[\tau,\tau]$,
the unique cubic-in-$\tau$ contribution is
\begin{equation}
2\langle a_x,b_x\rangle
=\bigl\langle J_x^\top\tau,\,H_x[\tau,\tau]\bigr\rangle ,
\label{eq:app_gap_odr_unique_cubic_term}
\end{equation}
which is precisely the leading term of
Eq.~\eqref{eq:app_gap_odr_expansion}. Every remaining term in
Eq.~\eqref{eq:app_gap_odr_square_full_expansion} is of order
$\|\tau\|^4$ or higher, by the orders recorded below
Eq.~\eqref{eq:app_gap_task_expansion_components}. Collecting them into
$R_4$, the remainder bounds in
Eq.~\eqref{eq:app_gap_taylor_remainder_bounds}, together with the
Jacobian bound Eq.~\eqref{eq:app_gap_jacobian_bound} and the
integrability of the derivative moments in \textup{(A2)}, imply
$R_4(\tau)=O(\|\tau\|^4)$ and
Eq.~\eqref{eq:app_gap_odr_remainder_gradient_bound}: as in the proof of
Lemma~\ref{lem:app_task_loss_perturbation}, each monomial's derivative
is one order lower in $\tau$ than the monomial itself, and the
associated constants are products of $M_J$, factors of $\|H_x\|$, and
$C_{3,x},C_{3,x}'$, whose expectations under $\mathcal Q$ are finite.
This proves Eq.~\eqref{eq:app_gap_odr_expansion} and, in particular,
the main-text expansion Eq.~\eqref{eq:odr_mode_perturbation}.

It remains to bound the gradient of the cubic term. Let
\begin{equation}
\Delta_3(\tau)
:=\mathbb E_{x\sim\mathcal Q}
\!\left[\bigl\langle J_x^\top\tau,\,H_x[\tau,\tau]\bigr\rangle\right].
\label{eq:app_gap_odr_cubic_definition}
\end{equation}
For an arbitrary direction $v$, using the symmetry of the bilinear map
$H_x$,
\begin{equation}
D\Delta_3(\tau)[v]
=\mathbb E_{\mathcal Q}
\!\left[\bigl\langle J_x^\top v,\,H_x[\tau,\tau]\bigr\rangle\right]
+2\,\mathbb E_{\mathcal Q}
\!\left[\bigl\langle J_x^\top\tau,\,H_x[\tau,v]\bigr\rangle\right],
\label{eq:app_gap_odr_cubic_directional_derivative}
\end{equation}
or equivalently, in gradient form,
\begin{equation}
\nabla\Delta_3(\tau)
=\mathbb E_{\mathcal Q}
\!\left[J_xH_x[\tau,\tau]
+2\,H_x(\tau,\cdot)^{*}\bigl(J_x^\top\tau\bigr)\right],
\label{eq:app_gap_odr_cubic_gradient_exact}
\end{equation}
where $H_x(\tau,\cdot)^{*}$ denotes the adjoint of the linear map
$v\mapsto H_x[\tau,v]$. By the Cauchy--Schwarz inequality and the
operator norm bounds in
Eqs.~\eqref{eq:app_gap_hessian_map} and
\eqref{eq:app_gap_jacobian_bound}, the two terms obey
\begin{equation}
\|J_xH_x[\tau,\tau]\|\le M_J\|H_x\|\|\tau\|^2,
\qquad
2\,\|H_x(\tau,\cdot)^{*}(J_x^\top\tau)\|\le2M_J\|H_x\|\|\tau\|^2 .
\label{eq:app_gap_odr_cubic_gradient_component_bounds}
\end{equation}
Taking expectations and using
$\mathbb E_{\mathcal Q}\|H_x\|=\overline H_{\mathcal Q}$ from
Eq.~\eqref{eq:app_gap_hessian_moments} yields
$\|\nabla\Delta_3(\tau)\|\le3M_J\overline H_{\mathcal Q}\|\tau\|^2$;
adding Eq.~\eqref{eq:app_gap_odr_remainder_gradient_bound} proves
Eq.~\eqref{eq:app_gap_odr_gradient_bound}.
\end{proof}

\paragraph{Displacement between the standard-mode solutions}
We now compare the standard-mode solutions of \textup{(O)} and
\textup{(J)}. Define
\begin{equation}
d_{\mathrm O}
:=\tau_{\mathrm{std}}^{\mathrm O}-\tau_{\mathrm{std}}^{\mathrm J}.
\label{eq:app_gap_odr_standard_displacement}
\end{equation}
By the objective decomposition
Eq.~\eqref{eq:app_gap_odr_standard_relation}, the two first-order
conditions read
\begin{equation}
\nabla F_{\mathrm{std}}^{\mathrm J}(\tau_{\mathrm{std}}^{\mathrm J})=0,
\qquad
\nabla F_{\mathrm{std}}^{\mathrm J}(\tau_{\mathrm{std}}^{\mathrm O})
+\frac{\lambda}{2}\,
\nabla\Delta_{\mathrm O}(\tau_{\mathrm{std}}^{\mathrm O})=0 ,
\label{eq:app_gap_odr_standard_stationarity}
\end{equation}
which places us exactly in the setting of
Lemma~\ref{lem:app_pathwise_solution_perturbation} with
$F=F_{\mathrm{std}}^{\mathrm J}$ and
$h=\tfrac{\lambda}{2}\Delta_{\mathrm O}$. Define the path-averaged
Hessian of the reference objective along the segment joining the two
solutions,
\begin{equation}
\mathcal G_{\mathrm O}
:=\int_0^1
\nabla^2F_{\mathrm{std}}^{\mathrm J}
\bigl(\tau_{\mathrm{std}}^{\mathrm J}+s\,d_{\mathrm O}\bigr)\,ds .
\label{eq:app_gap_odr_path_hessian}
\end{equation}
By the curvature bound
Eq.~\eqref{eq:app_gap_standard_reference_curvature},
\begin{equation}
\mathcal G_{\mathrm O}
\succeq\lambda A_{\mathcal Q}+\nu_{\mathrm{std}}I .
\label{eq:app_gap_odr_path_curvature}
\end{equation}
Lemma~\ref{lem:app_pathwise_solution_perturbation}
(Eq.~\eqref{eq:app_gap_path_perturbation_identity}) then gives the exact
resolvent representation
\begin{equation}
d_{\mathrm O}=-\mathcal G_{\mathrm O}^{-1}e_{\mathrm O},
\qquad
e_{\mathrm O}
:=\frac{\lambda}{2}\,
\nabla\Delta_{\mathrm O}(\tau_{\mathrm{std}}^{\mathrm O}).
\label{eq:app_gap_odr_displacement_resolvent}
\end{equation}
The key structural observation is that the perturbation
$e_{\mathrm O}$ is \emph{supported on the proxy}: by
Eq.~\eqref{eq:app_gap_odr_cubic_gradient_exact}, up to the higher-order
remainder, it is an expectation of vectors lying in the range of
$J_x$ for $x\sim\mathcal Q$---the very directions along which
$\mathcal G_{\mathrm O}$ carries the strong curvature
$\lambda A_{\mathcal Q}$ in
Eq.~\eqref{eq:app_gap_odr_path_curvature}. Rather than bounding
$\|d_{\mathrm O}\|$ in ordinary norm via
$\mathcal G_{\mathrm O}\succeq\nu_{\mathrm{std}}I$ alone, as was
sufficient for \textup{(F)}, we bound the \emph{target-directional}
displacement
$\|A_t^{1/2}d_{\mathrm O}\|
=\|A_t^{1/2}\mathcal G_{\mathrm O}^{-1}e_{\mathrm O}\|$
directly, exploiting both the proxy--target spectral relation and the
proxy-aligned curvature of $\mathcal G_{\mathrm O}$. This refinement is
what produces the factor $q_t(\alpha_t,\beta_t)$.

\paragraph{Spectral amplification through the resolvent}
We first control the proxy quadratic form of the resolvent. Fix any
$z\in\mathbb R^p$ and set $y:=\mathcal G_{\mathrm O}^{-1}z$, so that
$z=\mathcal G_{\mathrm O}y$. The Cauchy--Schwarz inequality combined
with the curvature lower bound
Eq.~\eqref{eq:app_gap_odr_path_curvature} gives
\begin{equation}
\|z\|\,\|y\|
\ge y^\top z
=y^\top\mathcal G_{\mathrm O}y
\ge\lambda\,y^\top A_{\mathcal Q}y+\nu_{\mathrm{std}}\|y\|^2
\ge2\sqrt{\lambda\nu_{\mathrm{std}}\,
\bigl(y^\top A_{\mathcal Q}y\bigr)\|y\|^2},
\label{eq:app_gap_odr_amgm_step}
\end{equation}
where the last step is the AM--GM inequality $a+b\ge2\sqrt{ab}$.
Dividing by $\|y\|$ (the case $y=0$ being trivial) and squaring,
\begin{equation}
z^\top\mathcal G_{\mathrm O}^{-1}A_{\mathcal Q}\mathcal G_{\mathrm O}^{-1}z
=y^\top A_{\mathcal Q}y
\le\frac{\|z\|^2}{4\lambda\nu_{\mathrm{std}}},
\qquad\text{i.e.}\qquad
\mathcal G_{\mathrm O}^{-1}A_{\mathcal Q}\mathcal G_{\mathrm O}^{-1}
\preceq\frac{1}{4\lambda\nu_{\mathrm{std}}}I .
\label{eq:app_gap_odr_proxy_resolvent_bound}
\end{equation}
Note the mechanism at work in
Eq.~\eqref{eq:app_gap_odr_proxy_resolvent_bound}: the stronger the
regularization $\lambda$, the more the resolvent
$\mathcal G_{\mathrm O}^{-1}$ attenuates perturbations along the proxy
directions---exactly the directions on which $e_{\mathrm O}$ is
supported.

We now transfer this attenuation to the target direction. By the
hypothesis of Theorem~\ref{theorem:gap_output}, the target
matrix admits an $(\alpha_t,\beta_t)$-relative spectral bound with
respect to the proxy
(Definition~\ref{definition:relative_spectral_bound}),
\begin{equation}
A_t\preceq\alpha_t A_{\mathcal Q}+\beta_t I .
\label{eq:app_gap_relative_spectral_bound}
\end{equation}
Together with
$\mathcal G_{\mathrm O}^{-2}\preceq\nu_{\mathrm{std}}^{-2}I$, which
follows from Eq.~\eqref{eq:app_gap_odr_path_curvature}, congruence by
$\mathcal G_{\mathrm O}^{-1}$ yields
\begin{equation}
\mathcal G_{\mathrm O}^{-1}A_t\mathcal G_{\mathrm O}^{-1}
\preceq\alpha_t\,
\mathcal G_{\mathrm O}^{-1}A_{\mathcal Q}\mathcal G_{\mathrm O}^{-1}
+\beta_t\,\mathcal G_{\mathrm O}^{-2}
\preceq\left(
\frac{\alpha_t}{4\lambda\nu_{\mathrm{std}}}
+\frac{\beta_t}{\nu_{\mathrm{std}}^2}
\right)I .
\label{eq:app_gap_odr_target_resolvent_bound}
\end{equation}
Setting $z=e_{\mathrm O}$ and taking square roots gives the
target-directional displacement bound
\begin{equation}
\|A_t^{1/2}d_{\mathrm O}\|
=\|A_t^{1/2}\mathcal G_{\mathrm O}^{-1}e_{\mathrm O}\|
\le q_t^{\sharp}(\alpha_t,\beta_t)\,\|e_{\mathrm O}\|,
\label{eq:app_gap_odr_target_directional_displacement}
\end{equation}
where
\begin{equation}
q_t^{\sharp}(\alpha_t,\beta_t)
:=\left(
\frac{\alpha_t}{4\lambda\nu_{\mathrm{std}}}
+\frac{\beta_t}{\nu_{\mathrm{std}}^2}
\right)^{1/2}.
\label{eq:app_gap_tight_proxy_factor}
\end{equation}
Because $\nu_{\mathrm{std}}=\kappa+\mu\ge\mu$ by
Eq.~\eqref{eq:app_gap_nu_definitions}, this tight factor is dominated
by the factor Eq.~\eqref{eq:proxy_spectral_factor} of the main text,
\begin{equation}
q_t^{\sharp}(\alpha_t,\beta_t)\le q_t(\alpha_t,\beta_t).
\label{eq:app_gap_tight_to_main_proxy_factor}
\end{equation}
In other words, the main-text factor $q_t$ corresponds to the
conservative choice $\kappa=0$; any strictly positive local curvature
of the standard task loss only tightens the envelope.

\paragraph{From displacement to regularizer-specific gap}
Since the linearized solutions of \textup{(O)} and \textup{(J)}
coincide (Eq.~\eqref{eq:app_gap_linear_odr_equals_jacobian}), the
linearized interference terms cancel exactly in the regularizer-specific gap:
\begin{equation}
E_{\mathrm O}(t)
=|\delta_{\mathrm O}(t)-\delta_{\mathrm J}(t)|
=\bigl|
\mathcal I(\tau_{\mathrm{std}}^{\mathrm O}\mid\mathcal P_t)
-\mathcal I(\tau_{\mathrm{std}}^{\mathrm J}\mid\mathcal P_t)
\bigr| .
\label{eq:app_gap_odr_excess_linear_cancellation}
\end{equation}
Applying the seminorm bound
Eq.~\eqref{eq:app_gap_interference_identity_radius} of
Lemma~\ref{lem:app_interference_difference_identity}, followed by the
target-directional displacement bound
Eq.~\eqref{eq:app_gap_odr_target_directional_displacement} and the
gradient bound Eq.~\eqref{eq:app_gap_odr_gradient_bound} with
$\|\tau_{\mathrm{std}}^{\mathrm O}\|\le R$ from \textup{(A1)},
\begin{align}
E_{\mathrm O}(t)
&\le2R\sqrt{\Lambda_t}\,\|A_t^{1/2}d_{\mathrm O}\|
\le2R\sqrt{\Lambda_t}\,
q_t^{\sharp}(\alpha_t,\beta_t)\,\|e_{\mathrm O}\|
\nonumber\\
&=\lambda R\sqrt{\Lambda_t}\,
q_t^{\sharp}(\alpha_t,\beta_t)\,
\bigl\|\nabla\Delta_{\mathrm O}(\tau_{\mathrm{std}}^{\mathrm O})\bigr\|
\le\lambda R\sqrt{\Lambda_t}\,
q_t^{\sharp}(\alpha_t,\beta_t)
\left(3M_J\overline H_{\mathcal Q}R^2+C_4R^3\right).
\label{eq:app_gap_odr_excess_explicit_tight}
\end{align}
Thus, with
\begin{equation}
C_{\mathrm O}(t,\mathcal Q)
:=\lambda R\sqrt{\Lambda_t}
\left(3M_J\overline H_{\mathcal Q}R^2+C_4R^3\right),
\label{eq:app_gap_Co_definition}
\end{equation}
we obtain
\begin{equation}
E_{\mathrm O}(t)
\le C_{\mathrm O}(t,\mathcal Q)\,q_t^{\sharp}(\alpha_t,\beta_t)
\le C_{\mathrm O}(t,\mathcal Q)\,q_t(\alpha_t,\beta_t).
\label{eq:app_gap_odr_excess_main_form}
\end{equation}
When the analysis is applied to a family of proxies with uniformly
bounded curvature moment $\overline H_{\mathcal Q}$, remainder constant
$C_4$, and radius $R$, we have
$C_{\mathrm O}(t,\mathcal Q)\le C_{\mathrm O}(t)$ for a finite
$C_{\mathrm O}(t)$ carrying no $(\alpha_t,\beta_t)$ dependence, and the
only explicit proxy--target spectral factor is
$q_t(\alpha_t,\beta_t)$. This proves
Theorem~\ref{theorem:gap_output}, with
$C_{\mathrm O}:=C_{\mathrm O}(t)$.

The contrast with instantiation \textup{(F)} is now fully explicit.
For \textup{(F)}, the perturbation is the same frozen quadratic form in
both modes, so the displacement is controlled in ordinary Euclidean
norm and the gap envelope is independent in $(\alpha_t,\beta_t)$
(Eq.~\eqref{eq:app_gap_fisher_uniform_constant}). For \textup{(O)},
the perturbation exists only in standard mode, is cubic rather than
quadratic, and is supported on the proxy; propagating it through the
resolvent toward the target direction necessarily passes through the
proxy--target spectral relation, and the resulting envelope grows with
the spectral mismatch through $q_t(\alpha_t,\beta_t)$. In exchange,
when the proxy is of high quality (small $\alpha_t,\beta_t$), the same
mechanism guarantees that the regularizer-specific gap of \textup{(O)} is small---the
theoretical basis for the practical prediction stated after
Corollary~\ref{cor:three_way_hierarchy} in the main text.

\paragraph{Fully proxy-dependent envelope}
For completeness, we record a variant of
Eq.~\eqref{eq:app_gap_odr_excess_explicit_tight} in which the target
operator norm $\Lambda_t$ is itself expressed through the proxy. If, in
addition, the proxy operator norm is bounded,
\begin{equation}
\|A_{\mathcal Q}\|\le M ,
\label{eq:app_gap_proxy_operator_bound}
\end{equation}
then the relative spectral bound
Eq.~\eqref{eq:app_gap_relative_spectral_bound} implies
\begin{equation}
\Lambda_t=\|A_t\|
\le\alpha_t\|A_{\mathcal Q}\|+\beta_t
\le\alpha_tM+\beta_t .
\label{eq:app_gap_target_operator_proxy_envelope}
\end{equation}
Substitution into
Eq.~\eqref{eq:app_gap_odr_excess_explicit_tight} yields
\begin{equation}
E_{\mathrm O}(t)
\le\lambda R
\sqrt{(\alpha_tM+\beta_t)
\left(
\frac{\alpha_t}{4\lambda\nu_{\mathrm{std}}}
+\frac{\beta_t}{\nu_{\mathrm{std}}^2}
\right)}
\left(3M_J\overline H_{\mathcal Q}R^2+C_4R^3\right).
\label{eq:app_gap_odr_fully_proxy_dependent_envelope}
\end{equation}
For fixed uniform curvature and radius bounds, the spectral factor in
Eq.~\eqref{eq:app_gap_odr_fully_proxy_dependent_envelope} is
coordinatewise non-decreasing in $\alpha_t$ and $\beta_t$. We
emphasize that this is an upper-envelope statement: the realized gap
can be smaller, because the cubic and higher-order terms may cancel.

\subsubsection{Instantiation (J) and additional remarks}
\label{app:gap_completion}

We now assemble the three ingredients derived in
Appendices~\ref{app:gap_jacobian_baseline}--\ref{app:gap_odr_excess}
into the three-way hierarchy of
Corollary~\ref{cor:three_way_hierarchy}, and close
with a few remarks that delimit the scope of the statements and check
their boundary behavior.

\paragraph{Regularizer-specific gap of instantiation (J)}
For the Jacobian reference, the regularizer-specific gap vanishes tautologically: by the
definition Eq.~\eqref{eq:excess_gap},
\begin{equation}
E_{\mathrm J}(t)=|\delta_{\mathrm J}(t)-\delta_{\mathrm J}(t)|=0,
\qquad G_{\mathrm J}(t)=B_t ,
\label{eq:app_gap_jacobian_excess_zero}
\end{equation}
which is the content of Proposition~\ref{prop:gap_jacobian}.
Combining Eq.~\eqref{eq:app_gap_jacobian_excess_zero} with the
\textup{(F)} bound Eq.~\eqref{eq:app_gap_fisher_uniform_constant} and the
\textup{(O)} bound Eq.~\eqref{eq:app_gap_odr_excess_main_form} yields the
three-way hierarchy Eq.~\eqref{eq:three_way_excess_summary} of
Corollary~\ref{cor:three_way_hierarchy}; the corresponding raw mode gaps
follow from the decomposition Eq.~\eqref{eq:app_gap_decomposition}.
By construction, $C_{\mathrm F}(t)$ and $C_{\mathrm O}(t)$ depend on
$(t,\lambda,\mu,\kappa,R)$ and the uniform moment bounds established in
Appendices~\ref{app:gap_assumptions} and~\ref{app:gap_fisher_excess},
but carry no explicit $(\alpha_t,\beta_t)$ factor, as claimed.
\hfill$\square$

\paragraph{Raw mode gaps of the frozen penalties}
Theorems~\ref{theorem:gap_fisher} and~\ref{theorem:gap_output} are stated in terms of the
regularizer-specific gap $E_\Phi$ rather than the raw mode gap $G_\Phi$. To keep the
statement about (J) precise, we record here what the raw mode gap
looks like for the two frozen-penalty instantiations. Since both
\textup{(J)} and \textup{(F)} satisfy
$\Phi_{\mathrm{std}}=\Phi_{\mathrm{lin}}$, their own raw mode changes
are driven solely by the task-loss perturbation $\Psi$. Repeating the
argument of Proposition~\ref{prop:app_jacobian_baseline_bound} with the
appropriate frozen positive-semidefinite matrix ($A_{\mathcal Q}$ for
\textup{(J)}, $\widehat F_{\mathcal Q}$ for \textup{(F)}) gives, in both
cases,
\begin{equation}
G_{\mathrm J}(t),\;G_{\mathrm F}(t)
\le\frac{2\Lambda_tR}{\mu}
\left(R_{\max}\overline H_sR+C_\Psi R^2\right).
\label{eq:app_gap_frozen_raw_mode_gap_envelope}
\end{equation}
Two observations follow from
Eq.~\eqref{eq:app_gap_frozen_raw_mode_gap_envelope}. First, both raw
mode gaps are small in a near-linear regime (where
$R_{\max}\overline H_s$ and $C_\Psi$ are moderate) and are independent in the
proxy--target spectral parameters $(\alpha_t,\beta_t)$. Second, they
are not exactly zero in a non-linear model: the vanishing statement
$E_{\mathrm J}(t)=0$ is about the \emph{regularizer-specific gap}, not the raw mode gap,
which is precisely the point of the regularizer-specific-gap formulation.

\paragraph{Sanity check: exact linear regime}
As a consistency check, consider the degenerate case where the model
is exactly linear in $\tau$ throughout the relevant region,
\begin{equation}
f(x;\theta_0+\tau)=f(x;\theta_0)+J_x^\top\tau .
\label{eq:app_gap_exact_linear_model_condition}
\end{equation}
Then the two task losses coincide,
$\mathcal L_{\mathrm{task}}^{\mathrm{std}}
=\mathcal L_{\mathrm{task}}^{\mathrm{lin}}$, and the two forms of the
output-drift penalty coincide,
$\Phi_{\mathrm{std}}^{\mathrm O}
=\Phi_{\mathrm{lin}}^{\mathrm O}=R_{\mathcal Q}$. Consequently, for
every $\Phi\in\{\mathrm J,\mathrm F,\mathrm O\}$,
\begin{equation}
F_{\mathrm{std}}^{\Phi}=F_{\mathrm{lin}}^{\Phi},
\qquad
\tau_{\mathrm{std}}^{\Phi}=\tau_{\mathrm{lin}}^{\Phi},
\qquad
G_\Phi(t)=E_\Phi(t)=0 .
\label{eq:app_gap_all_raw_gaps_zero_exact_linear}
\end{equation}
Note that the \textup{(F)} solution may still differ from the
\textup{(J)} solution when the frozen curvature is inexact,
$\widehat F_{\mathcal Q}\neq A_{\mathcal Q}$; nevertheless, its
\emph{own} standard and linearized solutions coincide, so its raw mode
gap is zero. The regularizer-specific-gap formulation is therefore consistent with the
naive expectation that all three instantiations behave identically in
the linear regime.

\paragraph{Scope of the (F) statement: frozen matrix in both modes}
The proxy-independent conclusion for \textup{(F)} in
Theorem~\ref{theorem:gap_fisher} relies crucially on the
\emph{same} matrix $\widehat F_{\mathcal Q}$, frozen at $\theta_0$,
being used in both modes---this is what makes the perturbation
$\tfrac{\lambda}{2}\tau^\top E_{\mathcal Q}\tau$ a shared quadratic
form and permits the modewise reduction in
Appendix~\ref{app:gap_fisher_excess}. If, instead, the standard-mode
curvature matrix is recomputed along the trajectory, then
$\Phi_{\mathrm{std}}^{\mathrm F}\neq\Phi_{\mathrm{lin}}^{\mathrm F}$
and instantiation \textup{(F)} becomes a trajectory regularizer akin
to \textup{(O)}. In that setting an additional curvature perturbation
must be analyzed, and the proxy-independent constant bound does not follow
from the present theorem. The static, frozen-at-$\theta_0$ form is
also the standard practical setup for FIM/KFAC regularizers such as
TAK~\cite{tak2026}, and is the setting to which
Theorem~\ref{theorem:gap_fisher} applies.

\paragraph{Interpretation of the hierarchy}
With method-specific linearized references, the mathematically correct
summary is the hierarchy of regularizer-specific gaps in
Eq.~\eqref{eq:three_way_excess_summary}---zero for \textup{(J)},
bounded and proxy-independent for \textup{(F)}, and proxy-amplified through
$q_t(\alpha_t,\beta_t)$ for \textup{(O)}---while every raw mode gap
additionally contains the task-mode baseline through the decomposition
$G_\Phi(t)\le B_t+E_\Phi(t)$ of Eq.~\eqref{eq:app_gap_decomposition}.
Hence the shorthand ``zero / constant / increasing'' refers to the
\emph{regularizer-specific gap} envelope, not to the raw
standard-versus-linearized gap of each instantiation. In particular,
the practical prediction made after
Corollary~\ref{cor:three_way_hierarchy} in the main text---that with
a high-quality proxy the inexpensive instantiation \textup{(O)} tracks
Jacobian regularization even under standard fine-tuning, and that
among the three regularizer-specific gaps only that of \textup{(O)} carries an explicit
proxy-quality dependence---is a statement about the hierarchy of Eq.~\eqref{eq:three_way_excess_summary}. The corresponding empirical
verification is reported in Section~\ref{sec:experiments}.
\section{Additional Experimental Details For CLIP Vision Tasks}
\subsection{Detailed Basic Experimental Setup}\label{subsec:basic_experimenta_setup}
\textbf{Vision tasks.} We evaluate on eight standard image-classification benchmarks: Cars~\cite{DBLP:conf/iccvw/Krause0DF13}, DTD~\cite{Cimpoi_2014_CVPR}, EuroSAT~\cite{Helber2017EuroSATAN}, GTSRB~\cite{Stallkamp2011TheGT}, MNIST~\cite{726791}, RESISC45~\cite{Cheng2017RemoteSI}, SUN397~\cite{5539970}, and SVHN~\cite{37648}. These datasets were chosen to cover diverse visual domains and to match prior work on task arithmetic~\cite{tta2023,taujp2025,Adamerging2024,tsv2025}.

\textbf{Evaluation protocol.} We follow the task-addition protocol commonly used in the task-arithmetic literature: each expert model is fine-tuned independently on its task, converted to a task vector, and multiple task vectors are summed to produce the merged multi-task model. The merged model is then evaluated on each task’s test set. For fairness, the scalar scaling factor is tuned by grid search on a held-out validation split.

\textbf{Metrics.} Consistent with~\cite{ta2023, linearatt2025, tta2023}, we report two complementary metrics: (1) Absolute accuracy, the standard top-1 classification accuracy of the merged model on each task; and (2) Normalized accuracy, which measures the performance retention relative to single-task specialists. Formally, for a set of tasks $\mathcal{T}$, the normalized accuracy is defined as the average ratio of the merged model's performance to the specialist's performance:

\begin{equation}
    \setlength\abovedisplayskip{4pt}    %
    \setlength\belowdisplayskip{4pt}
    \text{Norm. Acc.} = \frac{1}{|\mathcal{T}|} \sum_{t \in \mathcal{T}} \frac{\text{Acc}(\theta_{\text{merged}}, \mathcal{D}_t^{\text{test}})}{\text{Acc}(\theta_t^*, \mathcal{D}_t^{\text{test}})}
\end{equation}

where $\text{Acc}(\cdot, \mathcal{D}_t^{\text{test}})$ denotes the evaluation accuracy on the test set of task $t$, $\theta_{\text{merged}}$ represents the parameters of the combined multi-task model, and $\theta_t^*$ refers to the fine-tuned expert specific to task $t$.

\textbf{Architectures.}
Experiments are run with three CLIP-based Vision Transformer scales: ViT-B/32, ViT-B/16 and ViT-L/14. These choices permit evaluation across computational scales and align with prior model-merging studies~\cite{tta2023,taujp2025,linearatt2025}. During fine-tuning the image encoder is updated while the CLIP text encoder remains frozen.

\textbf{Hyperparameter configuration for scaling coefficient $\gamma$.}
To enable an equitable comparison across methods, a common scaling factor $ \gamma $ is applied uniformly to the aggregated task vectors, \textit{i.e.} $\theta_{MT} = \theta_0 + \gamma \sum \tau_t$. This single parameter is selected by grid search over $\{0.0, 0.05, \ldots, 1.0\}$ for each method. Notably, we do not adopt more flexible per-task coefficient strategies (\textit{e.g.}, assigning distinct $\gamma_t$ to each task). This uniform-scaling protocol aligns with earlier studies~\cite{linearatt2025, tta2023} and ensures that comparisons directly reflect the intrinsic properties of the task vectors generated by each method.

\subsection{Proxy Distribution Configurations}
\label{app:proxy_config}
We now provide complete details on the distributions used in our empirical validation (Section~\ref{subsec:exp_spectral}) and subsequent experiments.

\textbf{Oracle peer tasks (Oracle Union).} This distribution represents the theoretical upper bound of the disentanglement capability within our framework. For a specific task $t$ being fine-tuned, the Oracle proxy $\mathcal{Q}_{\text{oracle}}$ is constructed by sampling uniformly from the union of the datasets of all other tasks in the merge collection $\mathcal{T} \setminus {t}$.

Specifically, in our 8-task experimental setup, if the current task is Cars, the Oracle proxy consists of a balanced mixture of data from DTD, EuroSAT, GTSRB, MNIST, RESISC45, SUN397, and SVHN. By minimizing the gradient projection onto this distribution, the model explicitly orthogonalizes its weight updates against the actual subspaces of future merge targets. Although this approach is impractical for scalable model merging (as it requires access to private or unknown peer datasets), it provides a critical reference point: the closer a task-agnostic proxy (like ImageNet) is to this Oracle point in Figure~\ref{fig:theory_validation} (\textit{i.e.}, low $\alpha, \beta$), the more effective it is as a universal proxy.

\textbf{ImageNet (img).} We use the ILSVRC-2012 training split~\cite{DBLP:conf/cvpr/DengDSLL009, DBLP:journals/ijcv/RussakovskyDSKS15}, containing 1.28M images across 1000 classes. 

\textbf{COCO (coco).} We use the 5,000 images from the MS COCO 2017 validation split~\cite{coco2014} as a compact, task-agnostic proxy distribution. COCO contains diverse everyday scenes, typically involving multiple objects and rich contextual relationships, and therefore provides a natural-image distribution that emphasizes scene-level and multi-object structure rather than the predominantly object-centric categories of ImageNet. We discard all category, bounding-box, and instance annotations and use only the raw images as unlabeled proxy samples.

\textbf{ImageNet low-similarity subset (img-sub-u).} To examine whether an effective proxy needs to be semantically related to the target tasks, we construct a subset of ImageNet whose classes have low CLIP text similarity to all eight target tasks. The procedure has three steps.
\begin{enumerate}
\item Task-ImageNet similarity computation. For each target task $t \in \{\text{Cars}, \text{DTD}, \ldots\}$, we first aggregate all class labels within that task into a set $\mathcal{L}_t = \{l_1^{(t)}, l_2^{(t)}, \ldots\}$. For each ImageNet class $c$, we compute its maximum similarity to any class in task $t$,
\begin{equation}\label{eq:semantic_sim}
    \text{sim}(c, t) = \max_{l \in \mathcal{L}_t} \frac{\phi_{\text{text}}(\text{``a photo of } c\text{''})^\top \phi_{\text{text}}(\text{``a photo of } l\text{''})}{\|\phi_{\text{text}}(\cdot)\|_2^2},
\end{equation}
where $\phi_{\text{text}}$ denotes the frozen CLIP text encoder. This captures the strongest semantic connection between ImageNet class $c$ and task $t$.

\item Per-task filtering. For each task $t$, we rank all 1000 ImageNet classes by similarity and retain only the bottom 20\% (200 classes), yielding a task-specific low-similarity set $\mathcal{C}_t^{\text{orth}}$.
    
\item Universal intersection. To obtain a low-similarity subset to all tasks simultaneously, we compute the intersection,
\begin{equation}
    \mathcal{C}^{\text{universal}} = \bigcap_{t=1}^{8} \mathcal{C}_t^{\text{orth}}.
\end{equation}
This yields 150 ImageNet classes that are semantically distant from all target tasks. 
\end{enumerate}

\begin{figure}[t]
  \centering
   \includegraphics[width=0.8\linewidth]{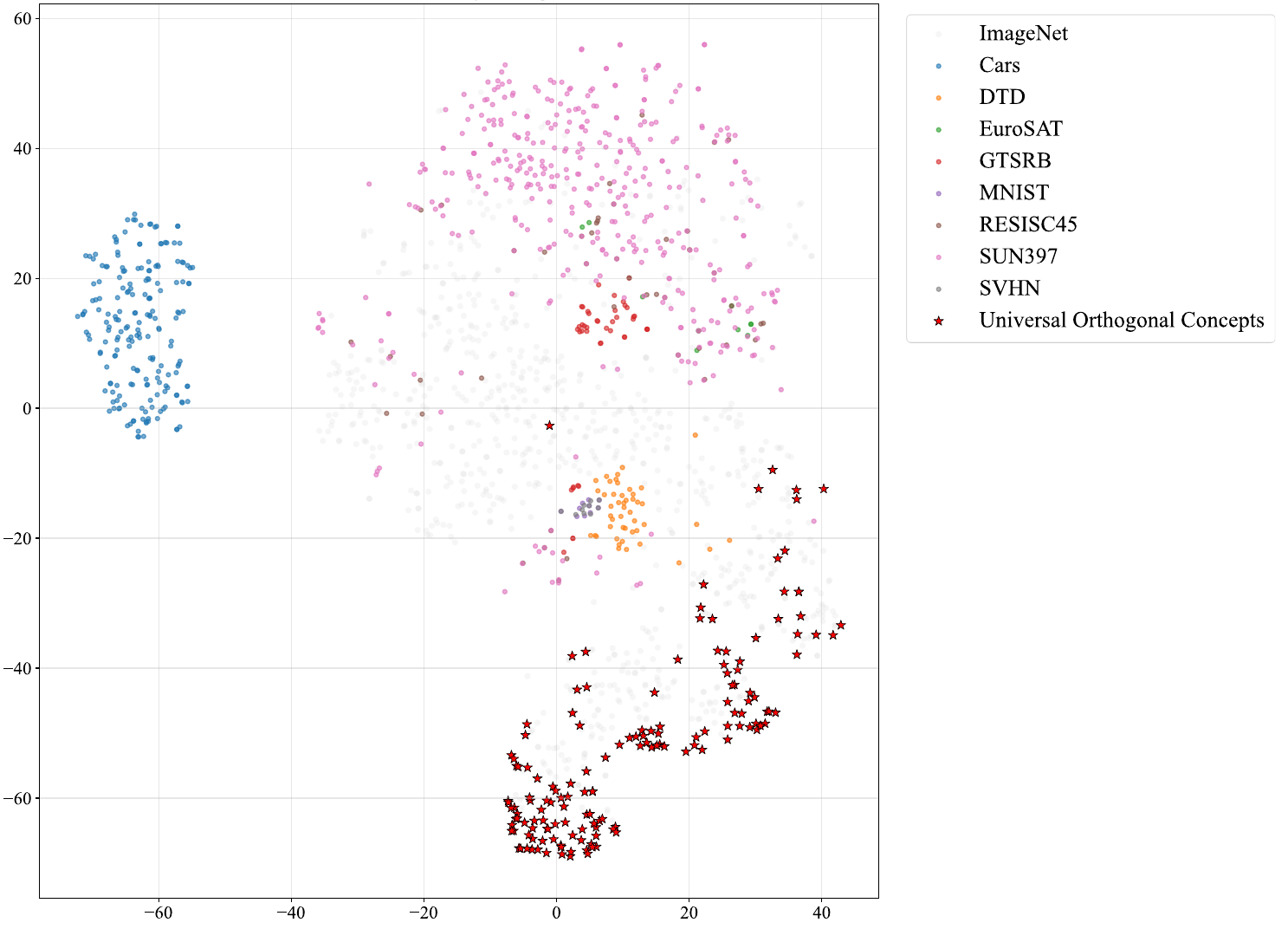}
   \caption{t-SNE projection of CLIP image embeddings of samples from the eight target benchmarks (colored points), full ImageNet (gray), and the 150 low-similarity ImageNet classes used for \textit{img-sub-u} (stars), shown for qualitative illustration. The classes are selected by their low CLIP text similarity to the labels of every target task; per-task similarity statistics are reported in Table~\ref{tab:img_sub_u_semantics}.}
   \label{fig:tsne_orthogonal}
\end{figure}

Table~\ref{tab:img_sub_u_semantics} reports the resulting similarities. Averaged over the eight tasks, the mean similarity of img-sub-u is 0.20 lower than that of all ImageNet classes and 0.33 lower than that of the 150 most similar classes, and for every task even the most similar img-sub-u class falls below the ImageNet average. None of these closest pairs shares an object category; where an association exists, it is at the level of texture or habitat (\textit{e.g.}, American coot $\leftrightarrow$ marsh). We therefore describe img-sub-u as having low, rather than zero, semantic similarity to the target tasks. Fig.~\ref{fig:tsne_orthogonal} provides a qualitative t-SNE view of the corresponding CLIP image embeddings.

\begin{table*}[t]
\centering
\caption{CLIP text similarity $\mathrm{sim}(c,t)$ (Eq.~\eqref{eq:semantic_sim}) between ImageNet classes and each target task, computed with CLIP ViT-B/32 and the prompt ``a photo of a [$\cdot$]''. For each task, we report the mean similarity over all 1{,}000 ImageNet classes, over the 150 classes most similar to the task (a reference of the same size as img-sub-u), and over the 150 classes of img-sub-u, together with the most similar img-sub-u class and its matched target label. The last row is an equal-weight average over the eight tasks.}
\label{tab:img_sub_u_semantics}
\small
\begin{tabular}{lccccl}
\toprule
 & \multicolumn{3}{c}{Mean $\mathrm{sim}(c,t)$ over ImageNet classes} & \multicolumn{2}{c}{Most similar \textit{img-sub-u} class} \\
\cmidrule(lr){2-4} \cmidrule(lr){5-6}
Target & All (1000) & Most similar (150) & img-sub-u (150) & $\mathrm{sim}$ & img-sub-u class $\leftrightarrow$ target label \\
\midrule
Cars     & 0.618 & 0.713 & 0.496 & 0.560 & European polecat $\leftrightarrow$ Volvo XC90 SUV 2007 \\
DTD      & 0.720 & 0.828 & 0.520 & 0.645 & Gila monster $\leftrightarrow$ scaly \\
EuroSAT  & 0.698 & 0.815 & 0.491 & 0.624 & bustard $\leftrightarrow$ pasture land \\
GTSRB    & 0.692 & 0.817 & 0.473 & 0.592 & green mamba $\leftrightarrow$ go straight (mandatory) \\
MNIST    & 0.696 & 0.822 & 0.478 & 0.596 & military hat (bearskin or shako) $\leftrightarrow$ 1 \\
RESISC45 & 0.720 & 0.857 & 0.507 & 0.627 & American coot $\leftrightarrow$ wetland \\
SUN397   & 0.738 & 0.898 & 0.528 & 0.638 & American coot $\leftrightarrow$ marsh \\
SVHN     & 0.696 & 0.822 & 0.478 & 0.596 & military hat (bearskin or shako) $\leftrightarrow$ 1 \\
\midrule
Average  & 0.697 & 0.822 & 0.496 & 0.610 & -- \\
\bottomrule
\end{tabular}
\end{table*}

\textbf{Fixed noise (fixed-noise).} We generate a single fixed batch of $B = 128$ images by sampling from a standard Gaussian distribution in pixel space: $X_{\text{fix}} \sim \mathcal{N}(0, I_{B \times 3 \times 224 \times 224})$ using a fixed random seed (42). During training, every iteration uses this identical batch. This serves as a naive baseline to demonstrate that meaningful gradient structure is necessary for effective regularization.

\textbf{Stable diffusion generated proxies.} We generate three variants using Stable Diffusion v1.5~\cite{DBLP:conf/cvpr/RombachBLEO22}, all with fixed random seed (42) for reproducibility.

\begin{itemize}
\item \textbf{sd-img:} Images generated from prompts ``a photo of a [class]'' using the 1000 ImageNet class labels~\cite{DBLP:conf/cvpr/DengDSLL009}. We generate 10 images per class sequentially, yielding a dataset of 10000 images whose semantic categories mirror ImageNet but whose pixels come entirely from the diffusion model's generative distribution. This variant isolates the effect of the underlying generative process from that of natural image collection: comparing \textit{sd-img} with \textit{ImageNet} tests whether a synthetic dataset that shares ImageNet's label distribution can replicate its regularization efficacy, while comparing it with \textit{sd-noise} and \textit{sd-wordnet} isolates the contribution of ImageNet-aligned semantic priors.

\item \textbf{sd-noise:} 1000 images generated by directly sampling random embedding vectors in the latent text space. Specifically, we sample $\mathbf{e} \sim \mathcal{N}(0, I_{77 \times 768})$ where 77 is the sequence length and 768 is the embedding dimension of CLIP. This serves as an unconditional generation baseline that produces visually coherent images (unlike pixel-space noise) but lacks semantic grounding. We use CFG scale 7.5 and 20 DDIM steps, with automatic retry for NSFW-filtered outputs to ensure exactly 1000 valid images.

\item \textbf{sd-wordnet:} Images generated from prompts ``a photo of a [noun]'' using a list of approximately 40,000 nouns extracted from WordNet noun synsets~\cite{DBLP:journals/cacm/Miller95}. We generate one image per noun sequentially using CFG scale 7.5 and 20 DDIM steps yielding a final dataset of $\sim$40K images with broad semantic coverage spanning everyday objects, animals, natural phenomena, and abstract concepts.
\end{itemize}

All generated images are resized to $224 \times 224$ and normalized using the standard CLIP preprocessing pipeline. 
These generated proxies allow us to progressively add semantic structure on top of the diffusion prior, while holding the underlying generative model fixed. \textit{sd-noise} tests whether visual manifold structure alone (without any semantic content) suffices for regularization; unlike \textit{fixed-noise} (pure pixel-space Gaussian noise), SD-generated images lie on the natural image manifold learned during diffusion training. \textit{sd-wordnet} tests whether broad semantic diversity through unconstrained noun coverage improves upon this manifold-only baseline. \textit{sd-img} further tightens the semantic scope to the 1000 ImageNet categories, so that a direct comparison against the real \textit{ImageNet} proxy isolates the effect of synthetic vs.\ real pixels while holding the label distribution fixed.

As shown in Section~\ref{subsec:exp_spectral}, all proxies significantly outperform \textit{fixed-noise}, validating that proxy distributions must possess meaningful gradient structure to enable effective regularization.

\subsection{Spectral Parameter Estimation Algorithm}\label{app:estimation}
We now describe how we estimate the regularized coverage coefficient $\alpha_t(\rho)$ of Remark~\ref{rem:coverage} for a target task $\mathcal{P}_t$ and a proxy distribution $\mathcal{Q}$. It yields the admissible pair $(\alpha_t(\rho),\rho\,\alpha_t(\rho))$ of the $(\alpha_t, \beta_t)$-relative spectral bound (Definition~\ref{definition:relative_spectral_bound}) used in Section~\ref{subsec:exp_spectral}. Forming $A_{\mathcal{P}_t}, A_{\mathcal{Q}} \in \mathbb{R}^{p \times p}$ explicitly is infeasible for modern neural networks (\textit{e.g.}, ViT-B/32 has $p \approx 88$M parameters), so we work with matrix-free operators throughout.

\textbf{Operators.} Let $J(x)=\nabla_\theta f(x;\theta_0)\in\mathbb{R}^{p\times c}$, where $f$ is the raw (unnormalized) image embedding of the CLIP visual encoder, \textit{i.e.}, the same output that instantiation (J) regularizes, and $\theta$ ranges over all $87.8$M visual-encoder parameters, including the patch embedding (\texttt{conv1}) and the positional embeddings. For a fixed sample set $\mathcal{S}=\{x_i\}_{i=1}^{n}$, the empirical Jacobian covariance acts on $v\in\mathbb{R}^p$ as
\begin{equation}
\widehat{A}_{\mathcal{S}}v=\frac{1}{n}\sum_{i=1}^{n}J(x_i)\bigl[J(x_i)^\top v\bigr],
\end{equation}
where $J(x_i)^\top v$ is a Jacobian--vector product (forward mode) and $J(x_i)z$ is a vector--Jacobian product (reverse mode). We write $\widehat{A}_{\mathcal{P}_t}$ and $\widehat{A}_{\mathcal{Q}}$ for this operator on the target and proxy sample sets. The model is kept in evaluation mode in float32, and all inputs are preprocessed once with the deterministic CLIP evaluation transform, so both operators are fixed throughout the iteration.

\textbf{Generalized eigensolver.} With $A=\widehat{A}_{\mathcal{P}_t}$ and $B=\widehat{A}_{\mathcal{Q}}+\rho I\succ 0$, the estimate $\widehat{\alpha}_t(\rho)$ is the largest eigenvalue of the generalized problem $Av=\alpha Bv$. We compute it with a single-vector locally optimal Rayleigh--Ritz iteration (Algorithm~\ref{alg:alpha_rho}). Each iteration requires one new product with $A$ and one with $B$; products for linear combinations of stored vectors are obtained by linearity, and no inverse of $B$ is formed.

\begin{algorithm}[t]
\caption{Matrix-free estimation of $\widehat{\alpha}_t(\rho)$.}
\label{alg:alpha_rho}
\begin{algorithmic}[1]
\REQUIRE Operators $v\mapsto Av$ and $v\mapsto Bv$ with $A=\widehat{A}_{\mathcal{P}_t}$, $B=\widehat{A}_{\mathcal{Q}}+\rho I$; maximum iterations $K$; tolerances $\eta_{\mathrm{tol}}=10^{-4}$, $\delta_{\mathrm{tol}}=10^{-5}$
\ENSURE Estimates $\widehat{\alpha}_t(\rho)$ and $\widehat{\beta}_t(\rho)=\rho\,\widehat{\alpha}_t(\rho)$

\STATE Draw $v_0 \sim \mathcal{N}(0, I_p)$ and normalize it so that $v_0^\top Bv_0=1$; set $p_0 \leftarrow \emptyset$
\FOR{$k = 0, 1, \ldots, K-1$}
    \STATE $\alpha_k \leftarrow v_k^\top Av_k$, \quad $r_k \leftarrow Av_k-\alpha_k Bv_k$
    \STATE $\eta_k \leftarrow \|r_k\|/(\|Av_k\|+|\alpha_k|\,\|Bv_k\|)$, \quad $\delta_k \leftarrow |\alpha_k-\alpha_{k-1}|/\max(|\alpha_k|,|\alpha_{k-1}|)$ (with $\delta_0:=0$)
    \IF{$\eta_k \le \eta_{\mathrm{tol}}$ \textbf{and} $\delta_k \le \delta_{\mathrm{tol}}$}
        \STATE \textbf{break}
    \ENDIF
    \STATE $S_k \leftarrow$ $B$-orthonormal basis $(s_1,s_2,\ldots)$ of $\mathrm{span}\{v_k, r_k, p_k\}$, with $s_1 = v_k$
    \STATE Solve $(S_k^\top AS_k)c=\lambda(S_k^\top BS_k)c$ and take the eigenvector $c$ of the largest $\lambda$
    \STATE $v_{k+1} \leftarrow S_k c$, $B$-normalized
    \STATE $p_{k+1} \leftarrow \sum_{j\ge 2}c_j s_j$, $B$-orthogonalized against $v_{k+1}$ and $B$-normalized
    \STATE Every $10$ iterations, recompute $Av_{k+1}$ and $Bv_{k+1}$ from scratch and reset $p_{k+1} \leftarrow \emptyset$
\ENDFOR
\STATE Recompute $Av$ and $Bv$ at the final vector $v$
\STATE \textbf{return} $\widehat{\alpha}_t(\rho)=v^\top Av$, $\widehat{\beta}_t(\rho)=\rho\,\widehat{\alpha}_t(\rho)$
\end{algorithmic}
\end{algorithm}

\textbf{Choice of $\rho$.} By Corollary~\ref{cor:coverage}(i), the tightest bound of Theorem~\ref{theorem:unified_bound} is obtained by minimizing $\alpha_t(\rho)\bigl(\frac{1}{4\lambda\mu}+\frac{\rho}{\mu^2}\bigr)$ over $\rho$, and the minimizer depends on $(\lambda,\mu)$ through the ratio $4\lambda\rho/\mu$ between the leakage and coverage contributions. We fix a single ridge level $\rho=10^5$, shared by all target tasks and proxies, so that every proxy is evaluated along the same direction of the admissible boundary. This level corresponds to roughly $10$--$20\%$ of $\lambda_{\max}(\widehat{A}_{\mathcal{P}_t})$ across tasks, so the factor $\lambda_{\max}(A_{\mathcal{P}_t})/\rho$ by which the consistency bound below amplifies sampling error is only about $5$--$10$.

\textbf{Setting.} For each target task, the target set contains $n=1{,}024$ images drawn without replacement, with a fixed seed, from the union of the official training and test splits. Labels are discarded, and the same target set is shared by all proxies of a task. Each universal proxy set is a fixed sample drawn in the same way from the corresponding proxy distribution, except fixed-noise, for which we use its entire fixed batch. For the Oracle Union, the proxy set for target $t$ contains an equal number of images from each of the seven tasks other than $t$. We run Algorithm~\ref{alg:alpha_rho} from two independent random starts with $K=150$ and keep the larger final Rayleigh quotient. All (task, proxy) pairs, including the Oracle Union, met both stopping criteria from both starts within $6$--$22$ iterations, and the two starts agreed to a relative difference below $10^{-6}$. On a small nonlinear network where the dense generalized eigenproblem can be solved exactly, the matrix-free operator matches the explicit Jacobian product to within $6\times 10^{-16}$, and the solver reproduces the largest generalized eigenvalue to $13$ significant digits.

\textbf{Consistency of the plug-in estimator.} Let $\varepsilon_P=\|\widehat{A}_{\mathcal{P}_t}-A_{\mathcal{P}_t}\|$ and $\varepsilon_Q=\|\widehat{A}_{\mathcal{Q}}-A_{\mathcal{Q}}\|$ in operator norm. From $\widehat{A}_{\mathcal{P}_t}\preceq A_{\mathcal{P}_t}+\varepsilon_P I$, $A_{\mathcal{Q}}\preceq\widehat{A}_{\mathcal{Q}}+\varepsilon_Q I$, and $I\preceq\rho^{-1}(\widehat{A}_{\mathcal{Q}}+\rho I)$, together with the symmetric relations obtained by exchanging the empirical and population operators,
\begin{equation}
\bigl|\widehat{\alpha}_t(\rho)-\alpha_t(\rho)\bigr| \leq \frac{\varepsilon_P+\max\{\alpha_t(\rho),\widehat{\alpha}_t(\rho)\}\,\varepsilon_Q}{\rho}.
\end{equation}
Since the empirical covariances converge in operator norm as $n\to\infty$, $\widehat{\alpha}_t(\rho)$ is a consistent estimator of $\alpha_t(\rho)$ for every fixed $\rho>0$.

\subsection{Relation between instantiation (F) and TAK.}
\label{app:f_tak_protocol}
Instantiation (F) adopts the frozen Fisher/KFAC quadratic penalty introduced by TAK~\cite{tak2026}, but the (F) entries in Table~IV were produced by integrating this penalty into our unified codebase. They should therefore be read as a controlled evaluation of the Fisher/KFAC regularizer inside our framework rather than as a reproduction of the complete TAK method. This design lets (J), (F), and (O) share the same data processing, task heads, fine-tuning protocol, and task-addition evaluation, so that any performance difference is attributable to the regularizer itself rather than to a method-specific training recipe.

Our reported vision results follow the widely used 8-Vision task-arithmetic pipeline inherited from prior work. We use AdamW with learning rate $10^{-5}$ and weight decay $0.1$, a dataset-specific epoch schedule, batch size $128$ for ViT-B/32 and ViT-B/16, and gradient-norm clipping at $1.0$. The CLIP text encoder and the task-specific classification head are frozen, while all parameters of the visual encoder are optimized. As a result, the patch-embedding convolution (\texttt{conv1}) and the positional embeddings remain trainable.

The original TAK implementation adopts a TAK-specific optimization recipe. It optimizes zero-initialized task-vector parameters with learning rate $3\times10^{-4}$, applies no gradient clipping, and freezes \texttt{conv1} and the positional embeddings by default (\texttt{ft\_conv=ft\_pos\_embed=0}). It also uses its own curvature estimation and aggregation procedure together with parameter-wise penalty scaling. These differences alter both the optimization trajectory and the set of directions represented in the resulting task vector.

The parameter-subset difference is especially consequential for (F). In our shared implementation, the KFAC penalty covers the Transformer's linear projections and includes explicit Fisher terms for the normalization parameters, but it does not supply curvature blocks for \texttt{conv1} or the positional embeddings. When those parameters are trainable, their updates enter the task vector without being directly constrained by (F), and these unconstrained components accumulate once task vectors are summed. Freezing them instead sets the corresponding task-vector components to zero and can substantially change the merged accuracy. We observed this sensitivity in our implementation checks. Because several protocol choices differ at the same time, we do not attribute the entire numerical gap between our (F) entries and the TAK paper to any single factor. Reproducing TAK itself would require its original codebase and training recipe, whereas Table~IV deliberately retains the shared protocol so that (J), (F), and (O) can be compared in a controlled manner within our framework.

\subsection{Direct Measurement of the Regularizer-Specific Gaps}
\label{app:direct_gap}
We now describe how the regularizer-specific gaps $\widehat E_\Phi$ in Fig.~\ref{fig:gap_direct} are computed.

\textbf{Estimator.} For a source task $s$ and a target task $t\neq s$, we estimate the interference of Definition~\ref{definition:distribution_interference} on a fixed target sample set $\{x_i\}_{i=1}^{n}$ as
\begin{equation}\label{eq:I_hat}
\widehat{\mathcal I}(\tau\mid\mathcal P_t)=\frac{1}{n}\sum_{i=1}^{n}\bigl\|J(x_i;\theta_0)^\top\tau\bigr\|^2 ,
\end{equation}
where $J(x_i;\theta_0)^\top\tau$ is a Jacobian--vector product at $\theta_0$, as in Appendix~\ref{app:estimation}. Substituting $\widehat{\mathcal I}$ into Eq.~\eqref{eq:signed_mode_change} and Eq.~\eqref{eq:excess_gap} gives
\begin{equation}
\hat\delta_\Phi=\widehat{\mathcal I}(\tau^{\Phi}_{\mathrm{std},s}\mid\mathcal P_t)-\widehat{\mathcal I}(\tau^{\Phi}_{\mathrm{lin},s}\mid\mathcal P_t),
\qquad
\widehat E_\Phi=\bigl|\hat\delta_\Phi-\hat\delta_{\mathrm J}\bigr|,
\end{equation}
for $\Phi\in\{\mathrm J,\mathrm F,\mathrm O\}$. As in Eq.~(25), the Jacobian is taken at $\theta_0$ for both the linearized and the standard solutions. The six task vectors $\tau^{\Phi}_{m,s}$, $m\in\{\mathrm{lin},\mathrm{std}\}$, are evaluated on the same images, so $\widehat E_\Phi$ is the absolute value of the sample mean of the per-image combination
\begin{equation}
h_{\Phi,i}=\|J_i^\top\tau^{\Phi}_{\mathrm{std},s}\|^2-\|J_i^\top\tau^{\Phi}_{\mathrm{lin},s}\|^2-\|J_i^\top\tau^{\mathrm J}_{\mathrm{std},s}\|^2+\|J_i^\top\tau^{\mathrm J}_{\mathrm{lin},s}\|^2,
\end{equation}
with $J_i:=J(x_i;\theta_0)$. The absolute value is taken after averaging over images, and $\widehat E_{\mathrm J}=0$ holds identically.

\textbf{Setting.} We evaluate the task vectors of Table~\ref{tab:method_specific_proxy_results}, each taken as the full parameter difference $\tau=\theta-\theta_0$ of the fine-tuned checkpoint, without the merging coefficient $\gamma$. As in Appendix~\ref{app:estimation}, $f$ is the raw image embedding of the CLIP visual encoder and $J$ is taken over all visual-encoder parameters. For each target task, $n=1{,}024$ images are drawn without replacement, with a fixed seed, from the union of the official training and test splits, preprocessed once with the deterministic CLIP evaluation transform, and shared by all proxies, instantiations, and modes.
The model is kept in evaluation mode in float32, and $J(x_i;\theta_0)^\top\tau$ is computed by forward-mode automatic differentiation.

\textbf{Ordering by $\widehat\alpha_t(\rho)$.} Since $\beta_t=\rho\,\alpha_t(\rho)$ on the boundary of Remark~\ref{rem:coverage}, the factor of Theorem~\ref{theorem:gap_output} is
\begin{equation}
q_t(\alpha_t,\beta_t)=\sqrt{\alpha_t(\rho)}\,\Bigl(\frac{1}{4\lambda\mu}+\frac{\rho}{\mu^2}\Bigr)^{1/2}.
\end{equation}
At fixed $\rho$, $\lambda$, and $\mu$, Theorem~\ref{theorem:gap_output} therefore orders the proxies by $\widehat\alpha_t(\rho)$, which is the horizontal axis of Fig.~\ref{fig:gap_direct}, with $\rho=10^5$ as in Appendix~\ref{app:estimation}.
\section{Additional Experimental Details For LLM Experiments}
\subsection{LLM Experimental Setup for TRACE}\label{sec:app_llm_trace}

This appendix details the full setup used in Section~\ref{sec:llm_trace}.

\textbf{Backbones and tasks.} We consider two Llama-3.2 instruction variants~\cite{llama3}, Llama-3.2-1B-Instruct and Llama-3.2-3B-Instruct, and fine-tune one specialist per task on the seven tasks of TRACE~\cite{trace2023}: C-STANCE, FOMC, MeetingBank, ScienceQA, NumGLUE-cm, NumGLUE-ds, and 20Minuten. Task inputs are truncated to $512$ tokens. Following the reporting convention of~\cite{trace2023,mergopt2026}, we report accuracy for C-STANCE, FOMC, ScienceQA, NumGLUE-cm, and NumGLUE-ds, ROUGE-L for MeetingBank, and SARI for 20Minuten, and average the seven scores.

\textbf{Fine-tuning.} All experts are trained in bfloat16, at a learning rate of $2\times10^{-5}$ with linear decay. Per-task epoch counts follow the TRACE recommendation~\cite{trace2023}. MergOPT uses the same hyperparameters in all configurations, and instantiations~$(O)$ and~$(J)$ add their penalty to the objective that the optimizer minimizes.

\textbf{Universal proxy.} The proxy $\mathcal Q$ consists of $256$ non-empty WikiText-103~\cite{wikitext103-2017} sequences, each truncated or padded to $128$ tokens; it contains no TRACE labels or samples. Both instantiations use this proxy with a proxy mini-batch of size eight. Instantiation $(O)$ uses $\lambda_{O}=0.1$, and its reference cache $h(x;\theta_0)$ is computed once per backbone in bfloat16 and held on CPU. Instantiation $(J)$ uses $\lambda_{J}=10^{-5}$, selected with AdamW on Llama-3.2-1B from $\{10^{-6}, 3{\times}10^{-6}, 10^{-5}, 3{\times}10^{-5}, 10^{-4}, 3{\times}10^{-4}\}$ and used in all runs.

\textbf{Merging.} The seven task vectors are merged by Task Arithmetic with a shared coefficient $\gamma$, and each method is reported at the $\gamma$ that maximizes its seven-task average.

\textbf{Instantiation $(J)$ at LLM scale.} For instantiation~$(J)$, we compute the directional derivative $J_h(x;\theta_0)^{\top}\tau_s$ on proxy samples via forward-mode automatic differentiation. Following the standard practice of representing a Jacobian-vector product through a scalar tangent, we introduce an auxiliary scalar $\rho\in\mathbb{R}$ and define $\theta(\rho):=\theta_0+\rho\,\tau_s$, so that the JVP reduces to a one-dimensional derivative,
\begin{equation}
    \left.\frac{\mathrm d}{\mathrm d\rho}
        h\!\bigl(x;\theta(\rho)\bigr)\right|_{\rho=0}
    \;=\; J_h(x;\theta_0)^{\top}\tau_s .
    \label{eq:llm_taujp_full}
\end{equation}
This propagates the full parameter tangent $\tau_s$ through the network in a single forward-mode pass, without materializing the Jacobian $J_h\in\mathbb R^{p\times d}$. Because the SDPA attention kernel used at LLM scale in our software stack does not implement forward-mode automatic differentiation, we temporarily switch to eager attention and disable gradient checkpointing inside the forward-mode transform; gradient checkpointing is restored before the reverse-mode task-loss backward pass. The $(J)$-augmented objective is
\begin{equation}
    \mathcal{L}^{(J)}_s
    \;=\; \mathcal L_s(\theta_0+\tau_s)
        \;+\; \lambda_{J}\,
        \mathbb E_{x\sim\mathcal Q}
        \bigl\|J_h(x;\theta_0)^{\top}\tau_s\bigr\|_2^2 ,
    \label{eq:llm_taujp_loss}
\end{equation}
where the expectation is estimated on each proxy mini-batch. The penalty in Eq.~\eqref{eq:llm_taujp_loss} is exactly the quadratic form $R_{\mathcal Q}(\tau_s)=\tau_s^{\top}A_{\mathcal Q}\tau_s$ of the main theory (Section~\ref{sec:proxy_regularization}).

\textbf{Efficiency benchmark.} The comparison between instantiation $(O)$ and this implementation of $(J)$ reported at the end of Section~\ref{sec:llm_trace} is measured with AdamW on Llama-3.2-1B in a $100$-step controlled benchmark: task micro-batch of size eight and proxy micro-batch of size one. Peak GPU memory is measured by \texttt{torch.cuda.max\_memory\_allocated}, and per-step time is averaged over the $100$ steps after a $10$-step warm-up.

\textbf{On the omission of $(F)$.} At LLM scale, instantiation $(F)$ requires estimating and storing the frozen curvature $\widehat A_{\mathcal Q}$ at $\theta_0$ before fine-tuning and keeping it available throughout. Even in Kronecker-factored form~\cite{tak2026}, the per-layer factors of an 8B transformer occupy tens of gigabytes of persistent GPU memory, on top of the resident model, optimizer state, and activations. This is incompatible with our single-accelerator training budget and motivates the choice to evaluate only $(J)$ and $(O)$ at LLM scale. The theoretical characterization of $(F)$ in Theorem~2 (proxy-independent gap bounded by the curvature error $\varepsilon_F$) remains a natural target for future work with a Kronecker-factored curvature estimator adapted to distributed LLM training.

\begin{table*}[t]
\centering
\caption{Per-task performance of TRACE merging with Task Arithmetic on Llama-3.2-1B-Instruct. All variants of our framework use WikiText-103 as the universal proxy, which contains no TRACE labels or samples.  Within each optimizer module, \textbf{bold} denotes the best result per column.}
\label{tab:trace_llama1b}
\resizebox{\textwidth}{!}{%
\begin{tabular}{lccccccc|c}
\toprule
Method & C-STANCE & FOMC & MeetingBank & ScienceQA & NumGLUE-cm & NumGLUE-ds & 20Minuten & Avg.\ ($\uparrow$) \\
\midrule
Pre-Trained            & 0.3386 & 0.2581 & 0.2036 & 0.6780 & 0.1220 & 0.1646 & 0.3802 & 0.3064 \\
AdamW Fine-Tuned       & 0.4980 & 0.5988 & 0.3707 & 0.8524 & 0.3902 & 0.5793 & 0.3880 & 0.5254 \\
MergOPT Fine-Tuned     & 0.4957 & 0.6331 & 0.3158 & 0.8780 & 0.4390 & 0.5305 & 0.3829 & 0.5250 \\
\midrule
AdamW               & 0.4120 & 0.4899 & 0.2270 & 0.7245 & 0.2439 & 0.3963 & 0.3812 & 0.4107 \\
AdamW + (J)         & 0.4210 & 0.5242 & 0.2213 & \textbf{0.7440} & \textbf{0.2927} & 0.4024 & \textbf{0.3827} & 0.4269 \\
AdamW + (O)         & \textbf{0.4260} & \textbf{0.5464} & \textbf{0.2278} & 0.7405 & 0.2683 & \textbf{0.4207} & 0.3815 & \textbf{0.4302} \\
\midrule
MergOPT             & 0.4184 & \textbf{0.5403} & 0.2137 & 0.7495 & 0.2927 & 0.3963 & 0.3818 & 0.4275 \\
MergOPT + (J)       & 0.4185 & \textbf{0.5403} & 0.2155 & 0.7570 & 0.2927 & 0.3902 & \textbf{0.3863} & 0.4286 \\
MergOPT + (O)       & \textbf{0.4280} & 0.5383 & \textbf{0.2203} & \textbf{0.7655} & \textbf{0.3659} & \textbf{0.4268} & 0.3854 & \textbf{0.4472} \\
\bottomrule
\end{tabular}%
}
\end{table*}

\begin{table*}[t]
\centering
\caption{Per-task performance of TRACE merging with Task Arithmetic on Llama-3.2-3B-Instruct. All variants of our framework use WikiText-103 as the universal proxy. Formatting follows Table~\ref{tab:trace_llama1b}.}
\label{tab:trace_llama3b}
\resizebox{\textwidth}{!}{%
\begin{tabular}{lccccccc|c}
\toprule
Method & C-STANCE & FOMC & MeetingBank & ScienceQA & NumGLUE-cm & NumGLUE-ds & 20Minuten & Avg.\ ($\uparrow$) \\
\midrule
Pre-Trained            & 0.4082 & 0.3528 & 0.2054 & 0.8962 & 0.1707 & 0.2195 & 0.3857 & 0.3770 \\
\midrule
AdamW               & 0.4680 & 0.5565 & 0.2168 & 0.9110 & 0.4878 & 0.4268 & \textbf{0.3904} & 0.4939 \\
AdamW + (J)    & 0.4870 & \textbf{0.5726} & \textbf{0.2303} & 0.9130 & \textbf{0.5610} & 0.4695 & 0.3846 & \textbf{0.5169} \\
AdamW + (O)    & \textbf{0.4880} & 0.5565 & 0.2285 & \textbf{0.9160} & 0.5122 & \textbf{0.5000} & 0.3829 & 0.5120 \\
\midrule
MergOPT             & 0.4755 & 0.6028 & 0.2172 & 0.9155 & 0.4878 & 0.4268 & \textbf{0.3923} & 0.5026 \\
MergOPT + (J)       & 0.4890 & 0.6169 & 0.2188 & 0.9155 & \textbf{0.5122} & 0.5061 & 0.3902 & \textbf{0.5213} \\
MergOPT + (O)       & \textbf{0.4985} & \textbf{0.6452} & \textbf{0.2191} & \textbf{0.9185} & 0.4146 & \textbf{0.5183} & 0.3882 & 0.5146 \\
\bottomrule
\end{tabular}%
}
\end{table*}

\subsection{LLM Experimental Setup and Additional Results for GLUE} \label{sec:app_llm_glue} 
This appendix details the full setup used in Section~\ref{sec:llm_glue} and reports the per-task results. 

\textbf{Backbone and tasks.} We use Llama-3-8B~\cite{llama3} and fine-tune one specialist per task on the eight tasks of GLUE~\cite{glue2019}: CoLA, SST-2, MRPC, STS-B, QQP, MNLI, QNLI, and RTE.

\textbf{LoRA configuration.} We use rank-$8$ LoRA~\cite{lora2022} adapters with scaling factor $16$ and no LoRA dropout, applied to the query and value projections of every self-attention block, following the OSRM protocol~\cite{osrm2025}. Both LoRA factors $A_{\ell}$ and $B_{\ell}$ are trainable, yielding approximately $3.42$M trainable parameters ($\approx 0.046\%$ of the model).

\textbf{Fine-tuning.} Training uses bfloat16, a learning rate of $4\times10^{-4}$ with linear decay, a $6\%$ warmup ratio, and gradient checkpointing. SST-2, QQP, and QNLI are trained for two epochs and the remaining five tasks for five epochs. The task micro-batch is of size eight with eight-step gradient accumulation, giving an effective task batch of $64$ for every method. The best checkpoint is selected by the primary validation metric of each task.

\textbf{OSRM baseline.} We re-run OSRM~\cite{osrm2025} in the same pipeline: backbone, LoRA configuration, fine-tuning schedule, seed, merging, and evaluation are identical to those of the other rows. OSRM differs from standard LoRA only in the initialization of the adapters, which it computes analytically from training examples of other task so that the subspace of each adapter is orthogonal to the features of the other tasks to be merged. All OSRM results in Table~\ref{tab:llm_glue} and in this appendix come from this reproduction.

\textbf{Universal proxy and reference cache.} The proxy $\mathcal Q$ consists of $2{,}048$ non-empty WikiText-103~\cite{wikitext103-2017} sequences, obtained by removing empty records, shuffling the training split with seed $42$, and taking the first $2{,}048$ examples. It contains no GLUE labels or samples.
For instantiation $(O)$, we compute the reference representations $h(x;\theta_0)$ once with the pre-trained backbone in evaluation mode and cache them in bfloat16 on CPU. All eight tasks share the same proxy inputs and the same cached targets. During fine-tuning, one proxy mini-batch is paired with each task mini-batch. The output-drift penalty is averaged over both proxy examples and hidden dimensions, giving the objective
\begin{equation}
    \mathcal{L}^{(O)}_s= \mathcal{L}_{s}(\theta_0+\tau_s)+ \frac{\lambda_O}{2}\mathbb{E}_{x\sim\mathcal{Q}}\left[\frac{1}{d}
      \bigl\|h(x;\theta_0+\tau_s)-h(x;\theta_0)\bigr\|_2^2\right],
    \label{eq:app_llm_output_drift}
\end{equation}
with $\lambda_{O}=8$, selected from $\{1,2,4,8,16\}$, and a proxy mini-batch of size eight. Because the reference representations are pre-computed, training requires only one resident LLM and one additional forward-backward path through the proxy batch. It never keeps a second 8B-parameter model in GPU memory.

\textbf{LoRA-aware JVP for instantiation $(J)$.} Under LoRA, the task vector $\tau_s$ acts on each adapted layer through the effective update $\Delta W_{\ell}=B_{\ell}A_{\ell}$, which is bilinear in the two trainable factors. A forward-mode JVP that assigns independent tangents to $A_{\ell}$ and $B_{\ell}$ therefore propagates the first-order term $B_{\ell}\dot A_{\ell}+\dot B_{\ell}A_{\ell}$, which is not the directional derivative associated with the merged task vector. To recover the correct quantity, we introduce a single differentiable scalar $\rho$ that multiplies the adapter scaling of every active LoRA branch simultaneously, define $\theta(\rho):=\theta_0+\rho\,\tau_s$, and evaluate a JVP at $\rho=0$ with tangent one:
\begin{equation}
    g_s(x)
    \;:=\;
    \left.\frac{\mathrm{d}}{\mathrm{d}\rho}
      h\bigl(x;\theta(\rho)\bigr)\right|_{\rho=0}
    \;=\;
    J_h(x;\theta_0)^{\top}\tau_s .
    \label{eq:app_llm_taujp}
\end{equation}
This returns the derivative of the final hidden representation along the full effective update $\{B_{\ell}A_{\ell}\}_{\ell}$ in a single forward-mode pass, without materializing the Jacobian $J_h\in\mathbb{R}^{p\times d}$. The augmented objective is
\begin{equation}
    \mathcal{L}^{(J)}_s
    = \mathcal{L}_{s}(\theta_0+\tau_s)
    + \lambda_J\,\mathbb{E}_{x\sim\mathcal{Q}}\bigl\|g_s(x)\bigr\|_2^2 , %
    \label{eq:app_llm_taujp_loss}
\end{equation}
with $\lambda_{J}=3\times10^{-4}$, selected from $\{3{\times}10^{-5}$, $10^{-4}$, $3{\times}10^{-4}$, $5{\times}10^{-4}$, $10^{-3}\}$. Since $g_s(x)=J_h(x;\theta_0)^{\top}\tau_s$, the penalty in Eq.~\eqref{eq:app_llm_taujp_loss} is exactly the quadratic form $R_{\mathcal Q}(\tau_s)=\mathbb{E}_{x\sim\mathcal{Q}}\|J_h(x;\theta_0)^{\top}\tau_s\|_2^2=\tau_s^{\top}A_{\mathcal Q}\tau_s$ of the main theory (Section~\ref{sec:proxy_regularization}), with the last-token hidden representation $h$ as the model output.

\textbf{Merging and evaluation.} After fine-tuning, each adapter is merged into its backbone weights and converted to a task vector relative to the common pre-trained model, excluding parameters belonging to the task-specific score head. For Task Arithmetic, the eight task vectors are summed with a shared coefficient $\gamma\in\{0.1,0.2,\dots,0.6\}$. For TIES-Merging, each task vector is trimmed to the $20\%$ of its entries with the largest magnitude (reset rate $0.8$), a sign is elected for every entry, and the entries that agree with it are averaged before scaling by $\gamma\in\{0.1,0.2,\dots,1.3\}$. The trimming threshold is computed over the entries that LoRA updates, i.e., the query and value projection weights ($0.67$B of the $7.50$B parameters in each task vector). Under both operators, each method is reported at the $\gamma$ that maximizes its eight-task average (Table~\ref{tab:llm_glue_per_task}). At evaluation, the score head saved by the corresponding specialist is re-attached to the merged backbone. We report Matthews correlation for CoLA, the mean of Pearson and Spearman correlations for STS-B, and accuracy for the remaining tasks. 
Abs.\ in Table~\ref{tab:llm_glue} is the mean of the eight primary scores, and Norm.\ divides each merged score by the corresponding specialist score and averages the eight retention ratios.

\textbf{Per-task results.} Table~\ref{tab:llm_glue_per_task} lists the per-task scores behind Table~\ref{tab:llm_glue}. Under Task Arithmetic, the gains of $(O)$ and $(J)$ over standard LoRA come mainly from CoLA, SST-2, MNLI, and QNLI, which improve by $15.8$ to $28.3$ points. 
Together with MRPC, these four tasks are the ones on which merged standard LoRA falls furthest below its specialists (23 to 39 points), compared with 7 to 15 points on QQP, RTE, and STS-B; MRPC is discussed separately below.

The scores across $\gamma$ show where these gains come from. Standard LoRA has no coefficient that suits all tasks: MNLI and SST-2 need a large $\gamma$ (MNLI rises from $51.4\%$ at $\gamma=0.2$ to $81.0\%$ at $\gamma=0.4$), whereas CoLA and MRPC collapse once $\gamma$ exceeds $0.2$ (from $24.3$ to $6.0$ and from $55.1\%$ to $30.1\%$ at $\gamma=0.3$), so its best average is reached at $\gamma=0.2$. The proxy penalty largely removes this conflict. At $\gamma=0.4$, standard LoRA drops to $-0.3$ on CoLA and $64.8\%$ on QNLI, whereas $(O)$ retains $49.2$ and $87.3\%$, and $(J)$ retains $49.2$ and $86.9\%$. The best coefficient of both instantiations therefore moves to $\gamma=0.4$. The gains on CoLA and QNLI thus come directly from reduced interference, and those on MNLI and SST-2 from the larger coefficient that this reduction permits.

RTE and STS-B change little with $\gamma$ in this range: standard LoRA stays at $79$--$80\%$ on RTE and $76$--$79$ on STS-B for $\gamma\in\{0.2,0.3,0.4\}$, and $(O)$ and $(J)$ reach similar values. Interference on these two tasks does not grow with $\gamma$ in this range, so the penalty has little to remove there. 
The proxy penalty also lowers the specialists on the three smallest tasks, MRPC, STS-B, and RTE, by 2.5 to 6.5 points; on RTE, merged (J) stays within 1.1 points of its own specialist (79.42\% vs.\ 80.51\%). Under Task Arithmetic, (J) is also 3.4 points lower than standard LoRA on QQP.

MRPC is not retained by any merged model; apart from OSRM under Task Arithmetic, all merged accuracies lie below the majority-class rate of $68.38\%$. Its failure is a shift of the decision boundary rather than a gradual loss: $279$ of the $408$ pairs are paraphrases, yet the number of pairs labelled as paraphrases ranges from $88$ to $333$ across the merged rows of Table~\ref{tab:llm_glue_per_task}. Excluding MRPC, the Task Arithmetic averages are $63.91\%$ for standard LoRA, $75.08\%$ for OSRM, $76.61\%$ for $(O)$, and $76.83\%$ for $(J)$. Compared with OSRM under Task Arithmetic, $(J)$ is higher on CoLA, SST-2, and MNLI and lower on MRPC and RTE. Under TIES-Merging, both instantiations again improve CoLA, SST-2, and QNLI, with further gains on QQP for $(O)$ and on MNLI for $(J)$.

\begin{table*}[t]
\centering
\caption{Per-task GLUE results for the eight-task LoRA merging experiment of Section~\ref{sec:llm_glue}. All values are percentages. Each merged row uses the merging coefficient $\gamma$ that maximizes its eight-task average (Task Arithmetic: $0.2$, $0.3$, $0.4$, $0.4$; TIES-Merging: $0.6$, $0.8$, $1.0$, $1.0$; in row order). Within each merging operator (Task Arithmetic and TIES-Merging), the best result per task column is \textbf{bolded}, excluding the oracle OSRM (shaded); specialist rows are provided as reference and are not marked. OSRM accesses peer-task data during fine-tuning; the other three configurations do not.}
\label{tab:llm_glue_per_task}
\resizebox{\textwidth}{!}{%
\setlength{\tabcolsep}{5pt}
\begin{tabular}{llccccccccc}
\toprule
Stage & Method & CoLA & SST-2 & MRPC & STS-B & QQP & MNLI & QNLI & RTE & Avg. \\
\midrule
 & Standard LoRA~\cite{lora2022} & 63.45 & 95.76 & 88.24 & 90.83 & 90.25 & 88.56 & 94.29 & 87.00 & 87.30 \\
\rowcolor{lightgray}
\cellcolor{white} & OSRM~\cite{osrm2025} & 66.93 & 96.56 & 90.44 & 90.72 & 90.20 & 90.27 & 95.73 & 87.36 & 88.53 \\
 & Standard LoRA $+(O)$ & 65.02 & 96.33 & 85.78 & 86.59 & 90.05 & 89.56 & 95.22 & 81.59 & 86.27 \\
\multirow{-4}{*}{Specialist} & Standard LoRA $+(J)$ & 64.98 & 96.10 & 84.56 & 87.40 & 89.75 & 89.90 & 95.41 & 80.51 & 86.08 \\
\midrule
 & Standard LoRA & 24.34 & 62.84 & 55.15 & 76.02 & 81.52 & 51.38 & 71.10 & \textbf{80.14} & 62.81 \\
\rowcolor{lightgray}
\cellcolor{white} & OSRM & 40.78 & 80.39 & 74.51 & 77.09 & 77.37 & 76.53 & 87.13 & 86.28 & 75.01 \\
 & Standard LoRA $+(O)$ & \textbf{49.21} & 84.06 & \textbf{59.56} & 76.67 & \textbf{85.63} & 75.74 & \textbf{87.33} & 77.62 & \textbf{74.48} \\
\multirow{-4}{*}{Task Arithmetic} & Standard LoRA $+(J)$ & 49.18 & \textbf{86.47} & 55.64 & \textbf{78.06} & 78.14 & \textbf{79.67} & 86.89 & 79.42 & 74.18 \\
\midrule
 & Standard LoRA & 18.57 & 69.04 & 34.56 & \textbf{72.76} & 76.93 & 60.15 & 66.41 & \textbf{77.26} & 59.46 \\
\rowcolor{lightgray}
\cellcolor{white} & OSRM & 35.91 & 88.42 & 51.72 & 60.75 & 77.29 & 73.16 & 82.54 & 71.48 & 67.66 \\
 & Standard LoRA $+(O)$ & \textbf{43.55} & 76.26 & \textbf{56.37} & 65.61 & \textbf{84.75} & 59.93 & \textbf{85.06} & 71.48 & 67.88 \\
\multirow{-4}{*}{TIES-Merging} & Standard LoRA $+(J)$ & 40.28 & \textbf{82.68} & 56.13 & 71.44 & 73.79 & \textbf{80.64} & 83.32 & 73.65 & \textbf{70.24} \\
\bottomrule
\end{tabular}}
\end{table*}
\section{Comparison with Related Model Merging Methods}
\label{app:related}

Tables~\ref{tab:related-during} and~\ref{tab:related-pre} summarize every model merging method referenced in this paper except the survey~\cite{Yang2024}, grouped by the stage of the merging pipeline at which each method operates, and position our framework relative to each of them. 

\emph{During-merging methods (Table~\ref{tab:related-during}).} These methods resolve conflicts when aggregating task vectors that are already fixed, whereas our framework changes how each task vector is produced during fine-tuning. The two act on different stages and therefore compose. 
Replacing standard task vectors with ones fine-tuned under our framework improves every aggregation rule we tested: by $4.65$ to $12.03$ points on ViT-B/32 (Tables~\ref{tab:clip_scaling_complete} and~\ref{tab:during_merging}), by $8.42$ to $11.67$ points under Task Arithmetic and TIES-Merging on Llama-3-8B (Table~\ref{tab:llm_glue}), and by $1.20$ to $1.97$ points under Task Arithmetic on Llama-3.2 with~(O) (Table~\ref{tab:trace_1b}).
Moreover, AdaMerging~\cite{Adamerging2024}, Representation Surgery~\cite{Representation_surgery}, and E-WEMoE~\cite{li2026} require unlabeled inputs of the target tasks at merging time, while our framework requires no data of the target tasks at any stage. RI~\cite{ri2026} replaces these inputs with unlabeled auxiliary data such as ImageNet, but adapts each expert only after the set of tasks to be merged is fixed, since it uses the heads of all these tasks.

\emph{Pre-merging methods without an explicit cross-task constraint (Table~\ref{tab:related-pre}, top).} These methods change the parameterization~\cite{tta2023,llora2024}, the set of trainable parameters~\cite{linearatt2025}, the weight-space structure~\cite{orthoreg2026}, or the optimizer~\cite{mergopt2026,liu2026muon}. None of them explicitly measures how a task vector perturbs the outputs on inputs outside its own task. Our framework adds exactly such a function-space penalty and quantifies, through $(\alpha,\beta)$, how well it transfers to unseen tasks. Being a plug-in objective, it is combined with TTA, ATT-FT, and MergOPT in Tables~\ref{tab:clip_scaling_complete} and~\ref{tab:trace_1b}.

\emph{Pre-merging methods with an explicit cross-task constraint (Table~\ref{tab:related-pre}, middle).} These are the closest to ours in mechanism: $\tau_{Jp}$~\cite{taujp2025} and TAK~\cite{tak2026} correspond to instantiations~(J) and~(F) evaluated on the distributions of the other tasks, and OSRM~\cite{osrm2025} enforces a related orthogonality structurally through LoRA initialization. All three derive their constraint from the tasks to be merged, either from raw samples ($\tau_{Jp}$, OSRM) or from curvature statistics computed on them (TAK). The appendix of~\cite{tak2026} additionally reports ImageNet-TAK, whose KFAC factors are estimated once on ImageNet. It is the closest to our setting, but remains an empirical observation restricted to~(F). Our framework replaces task-specific information with a single task-agnostic proxy $\mathcal{Q}$ for all three instantiations, and Theorem~\ref{theorem:unified_bound} states the spectral condition under which this replacement retains control over the interference. In addition, existing analyses of such regularizers stop at the NTK linearization, whereas Theorems~\ref{theorem:gap_fisher} and~\ref{theorem:gap_output} characterize their behavior under standard non-linear fine-tuning.

\begin{table}[t]
\centering
\caption{During-merging methods. These methods act on fixed task vectors at the aggregation stage, whereas ours shapes the task vectors during fine-tuning, so the two can be combined. \emph{Alone}: the method applied to standard task vectors; \emph{+\,Ours}: the same method applied to task vectors fine-tuned with instantiation~(O). Unmarked numbers are absolute accuracies (\%) on the eight-task ViT-B/32 benchmark with the ImageNet proxy. $^\dagger$E-WEMoE-90\%, reported in~\cite{li2026} under its own pipeline. $^\S$Reported in~\cite{knots2025} under LoRA fine-tuning. AdaMerging refers to its task-wise variant. \emph{Merging rule}: used to merge task vectors in our main experiments; \emph{Combined}: applied on top of our task vectors in the compatibility study (Table~\ref{tab:during_merging}); \emph{Related work}: discussed only. ``--'': not evaluated in this paper.}
\label{tab:related-during}
\footnotesize
\setlength{\tabcolsep}{4pt}
\begin{tabular}{@{}llcc@{}}
\toprule
Method & Role in this paper & Alone & +\,Ours \\
\midrule
Task Arithmetic~\cite{ta2023}         & Merging rule, all experiments & 70.32 & 82.24 \\
TIES-Merging~\cite{Tiesmerging2023}   & Merging rule, Tabs.~\ref{tab:during_merging}, \ref{tab:osrm} & 72.73 & 81.20 \\
DARE~\cite{dare2024} & Related work & -- & -- \\
TSV~\cite{tsv2025}                    & Combined, Tab.~\ref{tab:during_merging} & 85.64 & 90.33 \\
KnOTS~\cite{knots2025}                & Related work & $68.00^\S$ & -- \\
AdaMerging~\cite{Adamerging2024}      & Combined, Tab.~\ref{tab:during_merging} & 70.71 & 82.74 \\
Representation Surgery~\cite{Representation_surgery} & Combined, Tab.~\ref{tab:during_merging} & 80.70 & 85.35 \\
E-WEMoE~\cite{li2026}                 & Related work & $89.1^\dagger$ & -- \\
RI~\cite{ri2026} & Related work & 76.5 & -- \\
\bottomrule
\end{tabular}
\end{table}

\begin{table*}[t]
\centering
\providecommand{\cmark}{\ensuremath{\checkmark}}
\providecommand{\xmark}{\ensuremath{\times}}
\caption{Pre-merging methods and how they differ from ours. \emph{No peer-task info}: \cmark{} if fine-tuning a task vector needs neither data nor statistics from the other tasks to be merged; the text in parentheses names the extra information used during fine-tuning. \emph{ViT-B/32}: absolute accuracy (\%) of Task Arithmetic over 8 tasks with grid-searched $\gamma$. Unmarked numbers come from Table~\ref{tab:clip_scaling_complete} (same pipeline); $^\dagger$reported in the original paper under its own pipeline and base fine-tuning, hence not directly comparable; $^\ddagger$reported by~\cite{gu2026smat} under its own pipeline. TAK, ImageNet-TAK, and DELTA are trained under setups that differ from ours (Appendix~\ref{app:f_tak_protocol}). \emph{Oracle baseline}: a method that uses peer-task information, which our setting does not assume. ``--'': no directly comparable absolute accuracy on this benchmark.}
\label{tab:related-pre}
\footnotesize
\setlength{\tabcolsep}{4pt}
\renewcommand{\arraystretch}{1.25}
\begin{tabular}{@{}>{\raggedright\arraybackslash}p{2.45cm}
>{\centering\arraybackslash}p{2.1cm}
>{\raggedright\arraybackslash}p{3.25cm}
>{\centering\arraybackslash}p{1.1cm}
>{\raggedright\arraybackslash}p{\dimexpr\textwidth-8.9cm-8\tabcolsep\relax}@{}}
\toprule
Method & No peer-task info & Role in this paper & ViT-B/32 & Difference from ours \\
\midrule
\multicolumn{5}{@{}l}{\textit{No explicit cross-task constraint: shape task vectors during fine-tuning, but do not penalize their effect on other tasks' inputs.}} \\
\addlinespace[1pt]
Nonlin.\ FT~\cite{ta2023} & \cmark & Baseline, Tab.~\ref{tab:clip_scaling_complete} & 70.32
& Standard fine-tuning, the starting point of every row below. Ours adds the proxy penalty to this step. \\
TTA~\cite{tta2023} & \cmark & Baseline, Tab.~\ref{tab:clip_scaling_complete} & 74.86
& Tangent-space fine-tuning improves disentanglement only implicitly; ours explicitly constrains \emph{where} $\tau_s$ acts and is analyzed beyond linearization. TTA also serves as the linearized mode (Sec.~\ref{subsec:exp_method_gap}). \\
L-LoRA~\cite{llora2024} & \cmark & Related work & --
& Linearizes only the LoRA adapters (PEFT counterpart of TTA). \\
ATT-FT~\cite{linearatt2025} & \cmark & Baseline, Tab.~\ref{tab:clip_scaling_complete} & 78.07
& Fine-tunes only the attention layers, which induces kernel-like behavior and improves disentanglement implicitly. Ours adds an explicit penalty and can be stacked on it (last row). \\
OrthoReg~\cite{orthoreg2026} & \cmark & Related work & $73.41^\dagger$
& Orthogonality prior on each layer's update in weight space, without reference to inputs. \\
MergOPT~\cite{mergopt2026} & \cmark{} & Baseline, Tab.~\ref{tab:trace_1b} & $73.15^\ddagger$
& Robust to adversarial weight offsets (flat joint landscape). Ours bounds functional drift on inputs. \\
Muon-FT~\cite{liu2026muon} & \cmark & Related work & $75.34^\dagger$
& Fine-tunes with Muon instead of AdamW. Ours changes the objective, not the optimizer. \\
\midrule
\multicolumn{5}{@{}l}{\textit{Explicit cross-task constraint: penalize the effect of $\tau_s$ on other tasks' inputs, using peer-task information.}} \\
\addlinespace[1pt]
$\tau_{Jp}$~\cite{taujp2025} & \xmark{} (raw samples) & Oracle baseline, Tab.~\ref{tab:clip_scaling_complete} & 86.74
& Same JVP penalty as our (J), but on peer-task samples: $O(N^2)$ cost, retraining whenever the task set changes, analyzed only under linearization. \\
TAK~\cite{tak2026} & \xmark{} (KFAC stats) & Oracle baseline, App.~XI-D & $86.0^\dagger$
& Same frozen-curvature penalty as our (F), but with KFAC pre-computed on every task to be merged and analyzed only under linearization; its ImageNet variant is listed below. Trained with the TAK recipe, which differs from ours (App.~XI-D). We bound the gap of (F) between standard and linearized fine-tuning by a proxy-independent constant. \\
OSRM~\cite{osrm2025} & \xmark{} (raw samples) & Oracle baseline, Tab.~\ref{tab:osrm} & --
& LoRA-specific: initializes the adapter subspace orthogonally to peer-task features, a structural constraint fixed before fine-tuning. Ours is a penalty in the training objective, for both full and LoRA fine-tuning. \\
\midrule
\multicolumn{5}{@{}l}{\textit{Task-agnostic data: constrain $\tau_s$ with a task-agnostic dataset in place of peer-task information.}} \\
\addlinespace[1pt]
ImageNet-TAK~\cite{tak2026} & \cmark\ (ImageNet KFAC) & Related work & $84.7^\dagger$
& TAK with a single KFAC estimated once on ImageNet and shared by all tasks (appendix of~\cite{tak2026}). It corresponds to our TTA+(F) with $\mathcal{Q}=$ ImageNet, but is trained with the TAK recipe (App.~\ref{app:f_tak_protocol}). We give the condition under which such a proxy suffices. \\
DELTA~\cite{delta2026} & \cmark\ (reference dataset) & Related work & $88.3^\dagger$
& EK-FAC curvature penalty estimated on a task-agnostic reference dataset (ImageNet-21k, C4), combined with distillation from a linearized teacher into the non-linear student; its training procedure and cost differ from ours (App.~\ref{app:f_tak_protocol}). We give the condition under which such a proxy suffices and bound the non-linear gaps. \\
\midrule
\multicolumn{5}{@{}l}{\textit{Ours: penalize the effect of $\tau_s$ on a single task-agnostic proxy $\mathcal{Q}$ (ImageNet); $(\alpha,\beta)$ decides when $\mathcal{Q}$ suffices.}} \\
\addlinespace[1pt]
\textbf{Ours} (J) & \cmark & & 83.73 & \\
\textbf{Ours} (F) & \cmark & & 77.03 & \\
\textbf{Ours} (O) & \cmark & & 82.24 & \\
ATT-FT + \textbf{Ours} (O) & \cmark & & 84.64 & \\
\bottomrule
\end{tabular}
\end{table*}

The ImageNet variant of TAK (ImageNet-TAK in Table~\ref{tab:related-pre}) follows the same TAK recipe, with the KFAC factors estimated once on ImageNet instead of on the other tasks. DELTA~\cite{delta2026} uses AdamW with the same learning rate, weight decay, and batch size as our pipeline, but its training procedure differs. For each task, it trains a linearized teacher and a non-linear student together, and distills the teacher's representation before the projection head, evaluated at a randomly rescaled teacher task vector, into the student. Its EK-FAC curvature regularizer is estimated on a 15\% subset of ImageNet-21k rather than on our ImageNet proxy, and its released configuration, like TAK, trains only the parameters covered by this regularizer, leaving conv1 and the positional embeddings frozen. As reported in~\cite{delta2026}, this takes about three times the training time and twice the memory of standard non-linear fine-tuning.

\textbf{Updated parameters.} Of these differences, the set of parameters that fine-tuning updates matters most for (F). The official implementations of TAK and DELTA~\cite{delta2026} update only the parameters covered by their curvature regularizer, namely the attention and MLP projections, the normalization layers, the class embedding, and the output projection, and keep the patch-embedding convolution (conv1) and the positional embeddings frozen. Every component of their task vectors is therefore penalized. Our shared pipeline instead updates all visual-encoder parameters, while our KFAC penalty supplies no curvature block for conv1 or the positional embeddings. These components of $\tau$ thus enter the task vector without being constrained by (F). On ViT-B/32 they amount to about 2.4M parameters, 2.7\% of the visual encoder. Because they act on every input token before the first Transformer block, their updates shift the representation of inputs from all tasks rather than only from the task being fine-tuned, and these shifts add up when the task vectors are summed. Freezing these parameters instead sets the corresponding components of every task vector to zero, and we observed in our
implementation checks that this choice can substantially change the merged accuracy of (F).

\vfill

\end{document}